\documentclass{article}

\usepackage{microtype}
\usepackage{lipsum,graphicx,multicol}
\usepackage{subcaption}
\usepackage{amsmath}
\usepackage{booktabs} 

\usepackage{hyperref}
\usepackage{xr-hyper}

\usepackage[accepted]{icml2020}

\usepackage{amssymb}
\usepackage{amsthm}
\usepackage{placeins}

\icmltitlerunning{BaCOUn: Bayesian Classifers with Out-of-Distribution Uncertainty}

\begin{document}

\twocolumn[
\icmltitle{BaCOUn: Bayesian Classifers with Out-of-Distribution Uncertainty}

\icmlsetsymbol{equal}{*}
\begin{icmlauthorlist}
\icmlauthor{Th\'eo Gu\'enais}{equal,seas}
\icmlauthor{Dimitris Vamvourellis}{equal,seas}
\icmlauthor{Yaniv Yacoby}{seas}
\icmlauthor{Finale Doshi-Velez}{seas}
\icmlauthor{Weiwei Pan}{seas}
\end{icmlauthorlist}

\icmlaffiliation{seas}{John A. Paulson School of Engineering and Applied Sciences,
Harvard University, Cambridge, MA, USA}

\icmlcorrespondingauthor{Th\'eo Gu\'enais}{tguenais@fas.harvard.edu}
\icmlcorrespondingauthor{Dimitris Vamvourellis}{dvamvourellis@g.harvard.edu}

\icmlkeywords{Machine Learning, ICML}

\vskip 0.3in
]



\printAffiliationsAndNotice{\icmlEqualContribution} 

\begin{abstract}
Traditional training of deep classifiers yields overconfident models that are not reliable under dataset shift. We propose a Bayesian framework to obtain reliable uncertainty estimates for deep classifiers. Our approach consists of a plug-in ``generator" used to augment the data with an additional class of points that lie on the boundary of the training data, followed by Bayesian inference on top of features that are trained to distinguish these ``out-of-distribution" points. 
\end{abstract}

\section{Introduction}
\label{submission}

In high-risk domains, classifiers must provide predictive uncertainty, so that decision making can be deferred to human expertise when the model is in doubt.
For humans to make effective decisions under uncertainty, a classifier must furthermore decompose uncertainty into: epistemic uncertainty, uncertainty that can be reduced with more observations; and aleatoric uncertainty, uncertainty that cannot be reduced due to inherent noise in the system \cite{gal2016uncertainty}.
For a classifier, epistemic uncertainty can be further decomposed into out-of-distribution (OOD) uncertainty arising in regions far away from the high-mass regions, and model uncertainty referring to uncertainty over the classification boundary in the high-mass regions of the data space \cite{malinin2018predictive}.
 As such, OOD uncertainty depends on the expressiveness of the functions in the model class. The higher the capacity of the model class, the more the models in this class will extrapolate differently in data-scarce regions resulting in higher OOD uncertainty.

Gaussian Processes (GPs) have long served as standard method for providing both reliable and accurate estimates of predictive uncertainty in classification. However, due to their inability to scale with the number of observations, Bayesian Neural Networks (BNNs) have been proposed as a scalable alternative \cite{neal2012bayesian,mackay1992practical}. BNNs provide a way to capture model uncertainty by placing a prior distribution over network weights. Although more scalable than GPs, inference for large BNNs remains  challenging. For this reason, Neural Linear Models (NLM) are becoming a popular BNN replacement ~\cite{activelearning_nlm, rl_nlm}. In a NLM, we place priors only on the last layer of network weights and we learn point estimates for the remaining weights. Inference for the last weight layer can then be performed analytically.
 
Unfortunately, both BNNs and NLMs struggle with modeling OOD uncertainty. 
While BNNs are equivalent to GPs in the limit of infinite width~\cite{neal1996priors}, 
recent work shows that, unlike GPs, the epistemic uncertainty of finite-sized BNN classifiers does not increase in data-poor regions \cite{vernekar2019analysis}. In this work, we show that NLM likewise struggles with providing high epistemic uncertainty for OOD data, irrespective of the architecture chosen.

Our contributions in this paper are twofold.
We first explain why, using pedagogical examples NLM classifiers are unable to model OOD uncertainty. Specifically, we show that in order to capture OOD uncertainty, the posteriors of these models must include decision boundaries that properly bound the data. However, the training procedure of NLM does not encourage learning such boundaries. 
Next, we propose a novel scalable  framework for training NLM classifiers that provides reliable model uncertainty, OOD uncertainty and aleatoric uncertainty. 

On synthetic datasets we show that we attain uncertainty estimates comparable with that of GPs while baseline models underestimate OOD uncertainty. On real datasets, we show that our training framework attains higher AUC on observed data and provides reliable epistemic uncertainty estimates allowing for distinguish in-distribution and OOD test data. 

\setlength{\textfloatsep}{8pt plus 1.0pt minus 2.0pt}
\begin{figure*}[!h]    
    \centering
    \begin{subfigure}[t]{0.24\textwidth}    
        \includegraphics[width=1.0\textwidth]{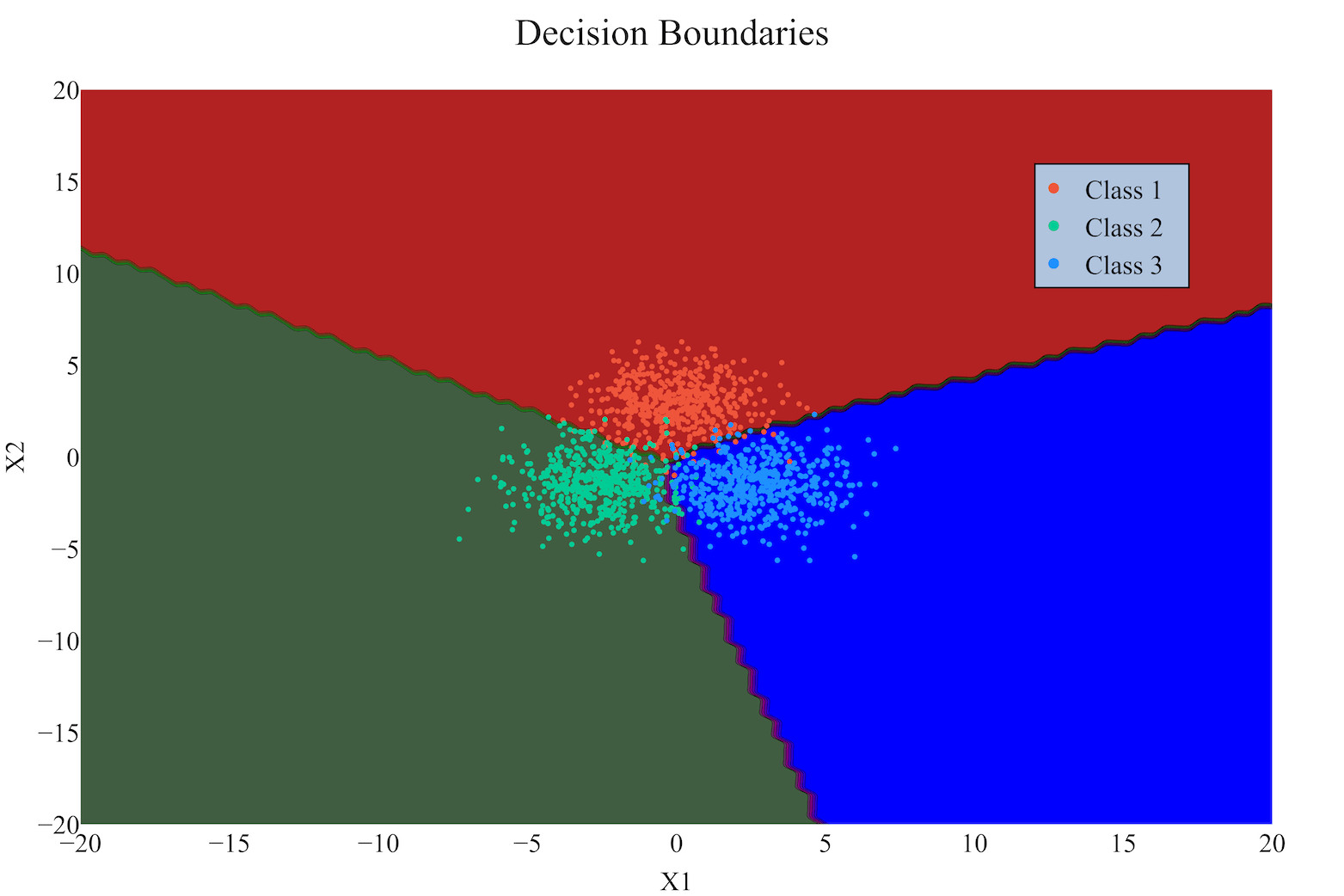}    
        \caption{NN Decision Boundaries}    
        \label{fig:pathology-1}    
    \end{subfigure}        
    \begin{subfigure}[t]{0.24\textwidth}            
        \includegraphics[width=1.0\textwidth]{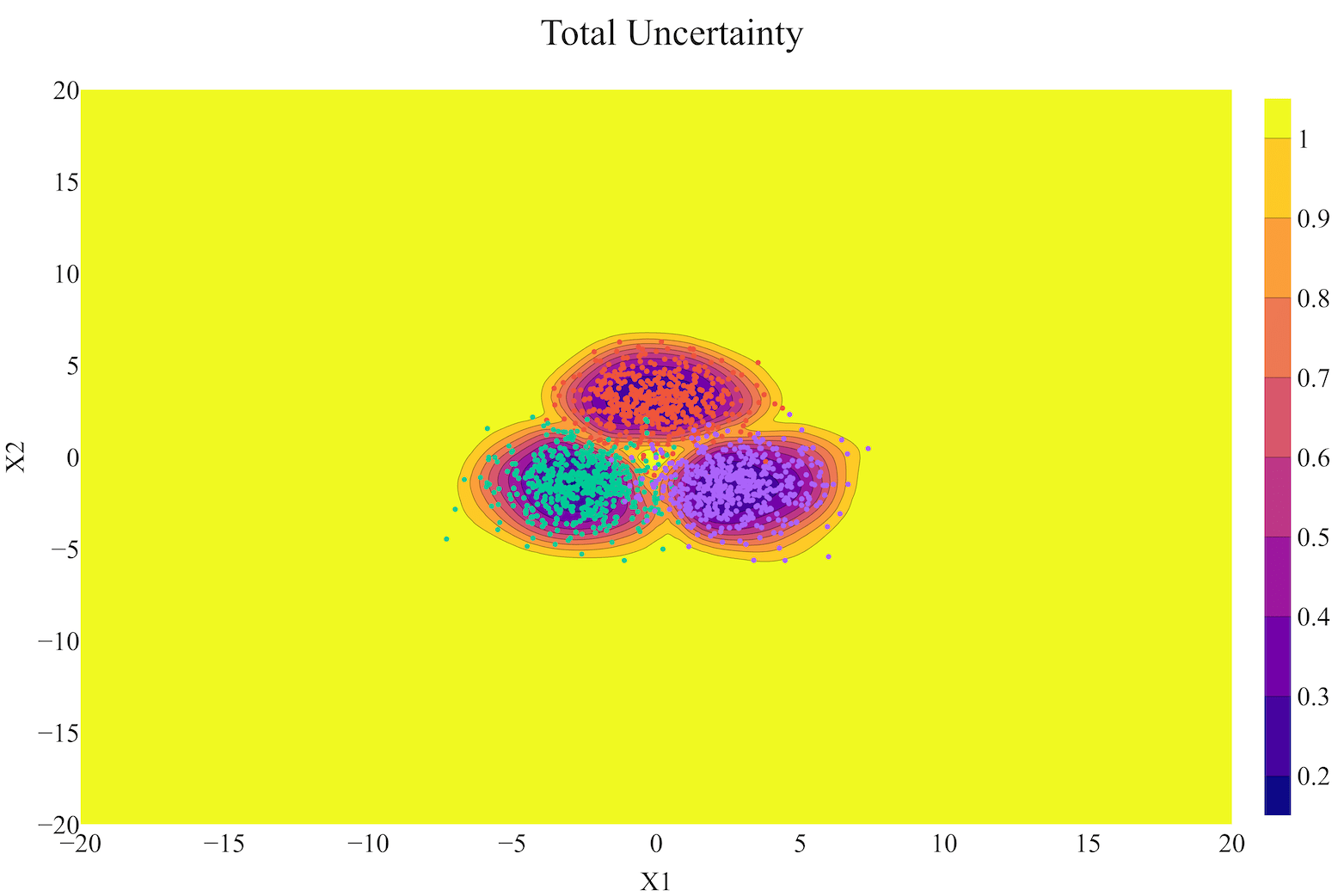}    
        \caption{GP Uncertainty}   
        \label{fig:pathology-2}    
    \end{subfigure}
    \begin{subfigure}[t]{0.24\textwidth}    
        \includegraphics[width=1.0\textwidth]{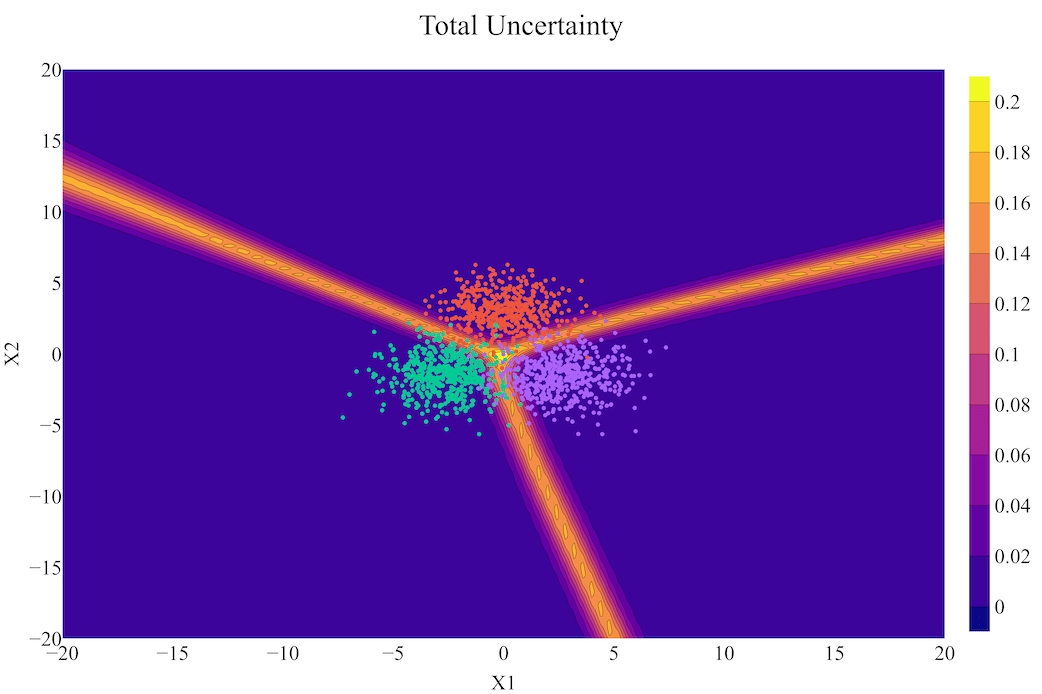}    
        \caption{NLM Uncertainty}
        \label{fig:pathology-3}    
    \end{subfigure}
    \begin{subfigure}[t]{0.24\textwidth}    
        \includegraphics[width=1.0\textwidth]{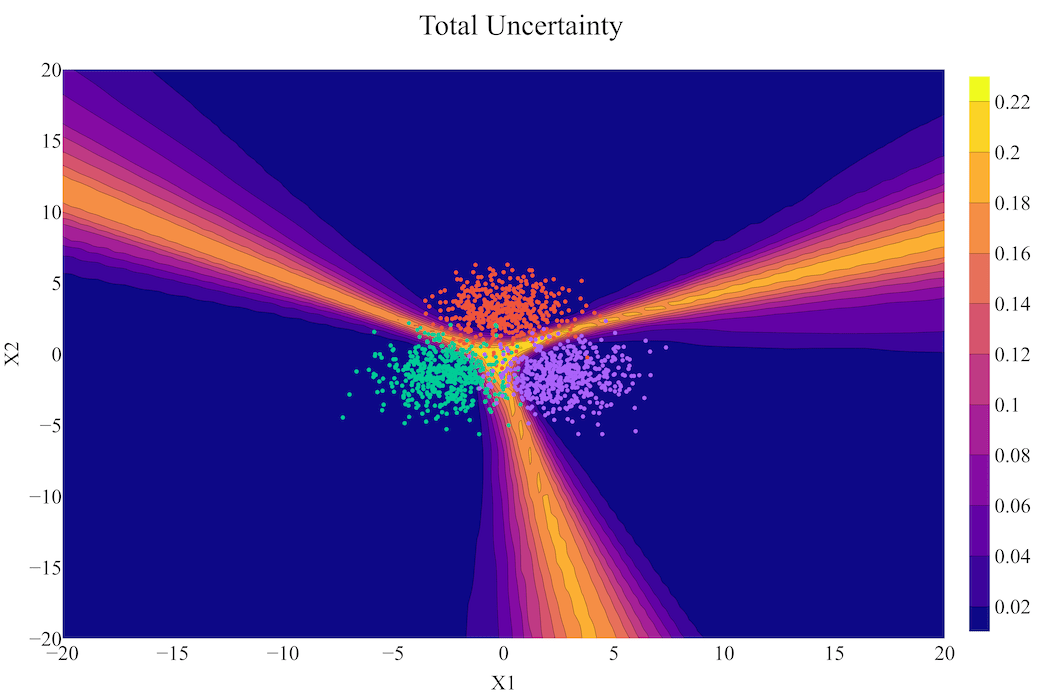}    
        \caption{BNN Uncertainty}
        \label{fig:pathology-4}    
    \end{subfigure}
    
     \caption{\textbf{Neural network models cannot capture OOD uncertainty.}  The above shows the decision boundaries of a Neural Net (a) along with entropy of the categorical distribution predicted by different methods:  GP (b), NLM (c), BNN (d). GPs make high-entropy (uncertain) predictions far from the training data while other models significantly underestimate OOD uncertainty.}
    \label{fig:pathology}
    \end{figure*}

\textbf{Related Work} Several non-Bayesian methods have been proposed to model OOD uncertainty.
They fall into two categories.
The first constrains a classifier to output high-entropy predictions on a priori OOD examples ~\cite{liang2017enhancing,lee2017training,sricharan2018building}, and the second trains a classifier on the original classes, plus an additional class of OOD examples \cite{vernekar2019analysis,vernekar2019out}. However, these methods do not provide uncertainty decomposition. In the first approach, points lying on the boundary between classes (which should have high aleatoric uncertainty) and points lying on the boundary of the data (which should have high epistemic uncertainty) both have high predictive entropy, but the different sources of the uncertainty (aleatoric and OOD) cannot be disentangled. Both approaches do not account for uncertainty over model parameters which may lead a classifier to make overconfident predictions for OOD points. 

\textbf{Background}
Let $\mathcal{D} =\{(x_1, y_1), \dots,(x_N , y_N)\}$ be a dataset of $N$ observations. Each input $x_n \in \mathbb{R}^D$ is a $D$-dimensional vector and each output $y_n$ is a label corresponding to one of $K$ classes.

A Neural Linear Model classifier assumes the following generative process:
\begin{align}
    y | x \sim \mathrm{Cat}\left( \mathrm{softmax}(W^\top \phi_\theta(x)  )\right), \quad W \sim p(W),
    \label{eq:nlm}
\end{align}
where $\phi_\theta(x)$ is the output of a neural network with $L$ number of output nodes and parameter $\theta$, augmented with a 1 for the bias term. We call $\phi_\theta$ the feature map since it extracts meaningful features from the data for the Bayesian linear classifier. First, the feature map $\phi_\theta$ is trained to maximize the observed data log-likelihood:
\begin{align}
\theta^* = \underset{\theta, W}{\mathrm{argmax}} \log p(y_1, \dots, y_N | x_1, \dots, x_N; \theta, W).
\label{eq:learned-basis}
\end{align}
Then, fixing $\phi_{\theta^*}$, we infer the posterior $p(W|\mathcal{D}, \theta^*)$ for the NLM classifier by either Hamiltonian Monte Carlo \cite{neal2011mcmc} or mean-field Gaussian variational inference \cite{ranganath2014black}. At test time, we make predictions by computing the posterior predictive distribution:
\begin{align}
    p(y^* | x^*, \mathcal{D}) &= \int p(y^* | x^*, W) p(W | \mathcal{D}, \theta^*) dW.
    \label{eq:predictive}
\end{align}

\section{Analysis of the OOD Uncertainty of Neural Linear Model Classifiers}

In this section, we show that in order for posterior predictives of NLMs to capture OOD uncertainty, these posteriors must include decision boundaries that bound high likelihood data regions. However, the training objective for NLM classifiers does not encourage for learning these boundaries.

\textbf{Neural Linear Model Classifiers are unable to model OOD uncertainty.} 

In Figure \ref{fig:pathology}, we visualize the entropy of the posterior predictive mean of various classifiers over the input space. While the GP classifier makes low-entropy (i.e. confident) predictions for points close to the observed data and high-entropy predictions for points far away, NLM classifiers make over-confident low-entropy predictions far from the observed data. The same failure can be observed in other models that provide predictive uncertainty such as full BNN trained with Black-Box Variational Inference (BBVI) \cite{ranganath2014black}. The OOD uncertainty of the GP classifier comes from the fact that a significant number of decision boundaries in the GP posterior properly bound each class in the data and hence the predicted labels probabilities for points far from the data have high variance. So why doesn't the posterior for NLM classifiers include such decision boundaries?

\textbf{Neural Linear Model underestimates OOD uncertainty because training does not encourage learning decision boundaries that bound the data.}
The training objective for the NLM model  encourages $\phi_\theta$ to extract features that are predictive of $y$. 
But a $\phi_\theta$ that is good for classification (e.g. one that expresses only linear boundaries) is not necessarily capable of expressing a decision boundary $W^\top\phi_\theta(x)$ that bounds the data. Hence, for these $\phi_\theta$'s, the resulting NLM classifier will underestimate OOD uncertainty.

In a simple experiment, we demonstrate that the features learned by an NLM classifier is good for classification but is not useful for OOD detection. Consider a dataset $\{ (x_n, y_n, b_n) \}_{n=1}^N$, where $x_n$ is the input, $y_n$ is the label, and $b_n$ is a binary indicator signaling whether the data-point is OOD. We train an NLM classifier to predict $y | x$ and ensure that the AUC is high. We then train a Naive Bayes classifier to predict $b$ given the features learned by $\phi_{\theta^*}$. 

In Appendix \ref{appendix:pathologies} Figure \ref{app_fig:nlm_pathology}, we see that the features learned by $\phi_{\theta}$ for classification are unable to predict the label $b$ (i.e. if a point is OOD). Furthermore, the posterior of the NLM model underestimates OOD uncertainty.

In contrast, we train a neural network classifier to jointly predict $y, b$ given $x$. We then extract the learned features, given by output of the last hidden layer $\phi_{\mathrm{latent}}(x)$, to predict $b$. In this case, since the features are explicitly trained to predict the OOD points, we can accurately predict $b$ with a Naive Bayes classifier fitted on these features. Appendix \ref{appendix:pathologies} Figure \ref{app_fig:nlm_pathology} shows that the posterior of this model includes decision boundaries that bound the data and thus the posterior uncertainty is higher for OOD points. The intuition behind this experiment forms the core of our inference framework in Section \ref{sec:bacoun}.

We emphasize that we show a problem with the training objective of NLMs and not an issue with the architecture of the model. In fact, we show later that, given identical NLM architectures, our training framework is able to capture OOD uncertainty while traditional training cannot.

\section{BaCOUn: Bayesian Classifier with OOD Uncertainty}
\label{sec:bacoun}

To ensure that the features of NLMs can be used to build decision boundaries that separate data from OOD points, we propose BaCOUn, a general training framework for NLMs:

\textbf{1.} Generate OOD samples on the boundary of the data.

\textbf{2.} Train a deterministic classifier to distinguish between the original $K$ classes and additionally between the generated OOD samples.

\textbf{3.} Fix the features, $\phi_{\theta^*}(x)$, learned by the deterministic classifier, and fit a Bayesian logistic regression model (in Equation \ref{eq:nlm}) on these features.

By explicitly training a classifier on the boundary samples, we show later that the learned features will be able to express decision boundaries that bound the data, and hence the posteriors of BaCOUn will capture OOD uncertainty.

Our framework is general, one can plug-in any method for each step.
Here, we propose one instantiation of this framework that performs well in practice. For the first step -- generating OOD samples -- we present a novel OOD sampler, described below. For the second step, we train a deep neural network with a $\mathrm{softmax}$ output activation, and for the final step, we use Hamiltonian Monte-Carlo (HMC) as well as mean-field BBVI on more taxing experiments.

\textbf{Novel Method for Generating OOD Boundary Points}
We propose a novel method to generate samples from the boundary of a given data distribution based on Normalizing Flows \cite{rezende,realnvp,huang2018neural}.
In our method, we train a RealNVP~\cite{dinh2016density} to maximize the log likelihood of the data $x$.
We then sample points on the boundary of the latent space of the learned flow, mapping the samples back into the data space. Details in Appendix \ref{appendix:flows}.


\begin{figure*}[!h]    
    \centering
    \begin{subfigure}[t]{0.24\textwidth}    
        \includegraphics[width=1.0\textwidth]{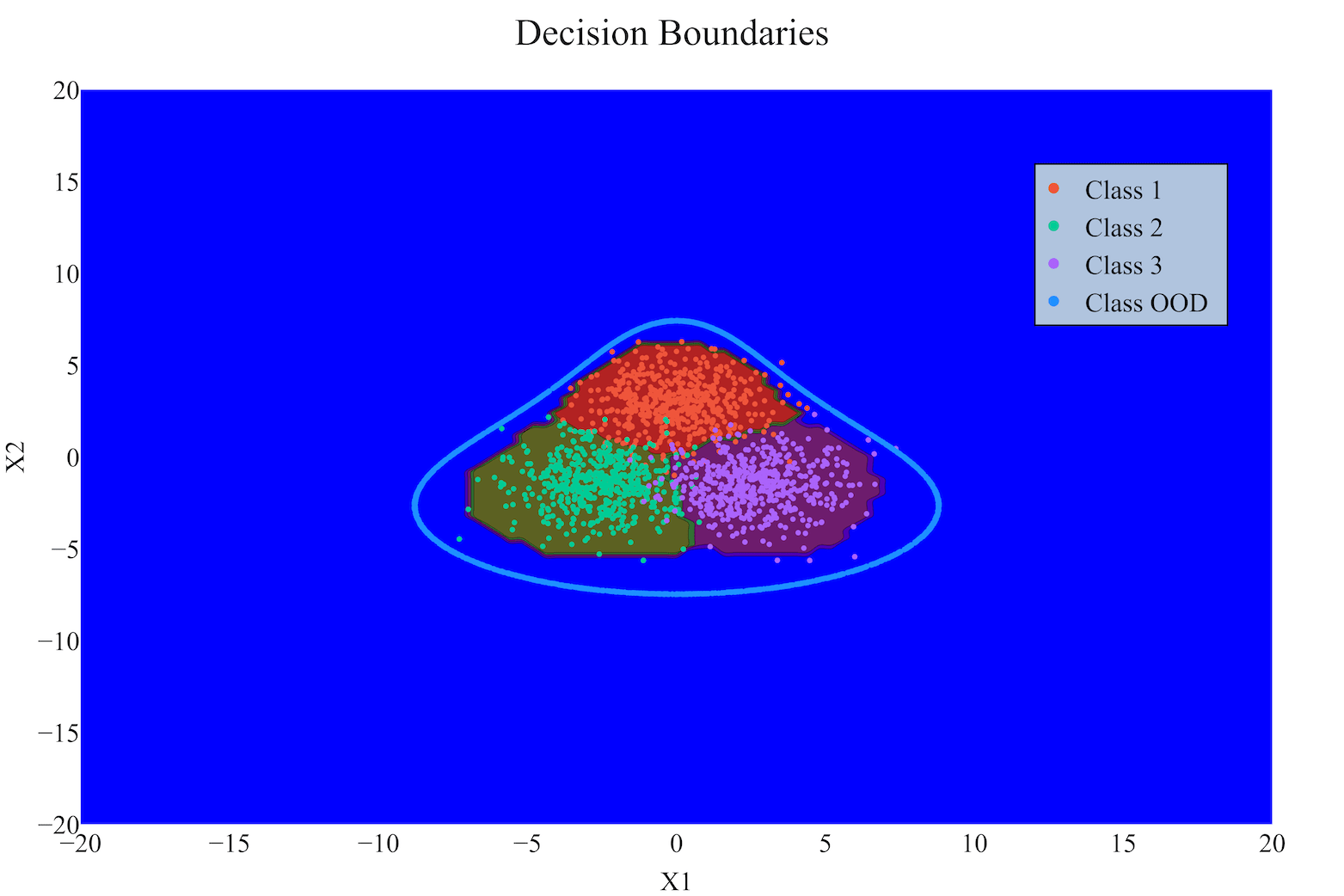}    
        \caption{Decision Boundaries}    
        \label{fig:bacoun-1}    
    \end{subfigure}        
    \begin{subfigure}[t]{0.24\textwidth}            
        \includegraphics[width=1.0\textwidth]{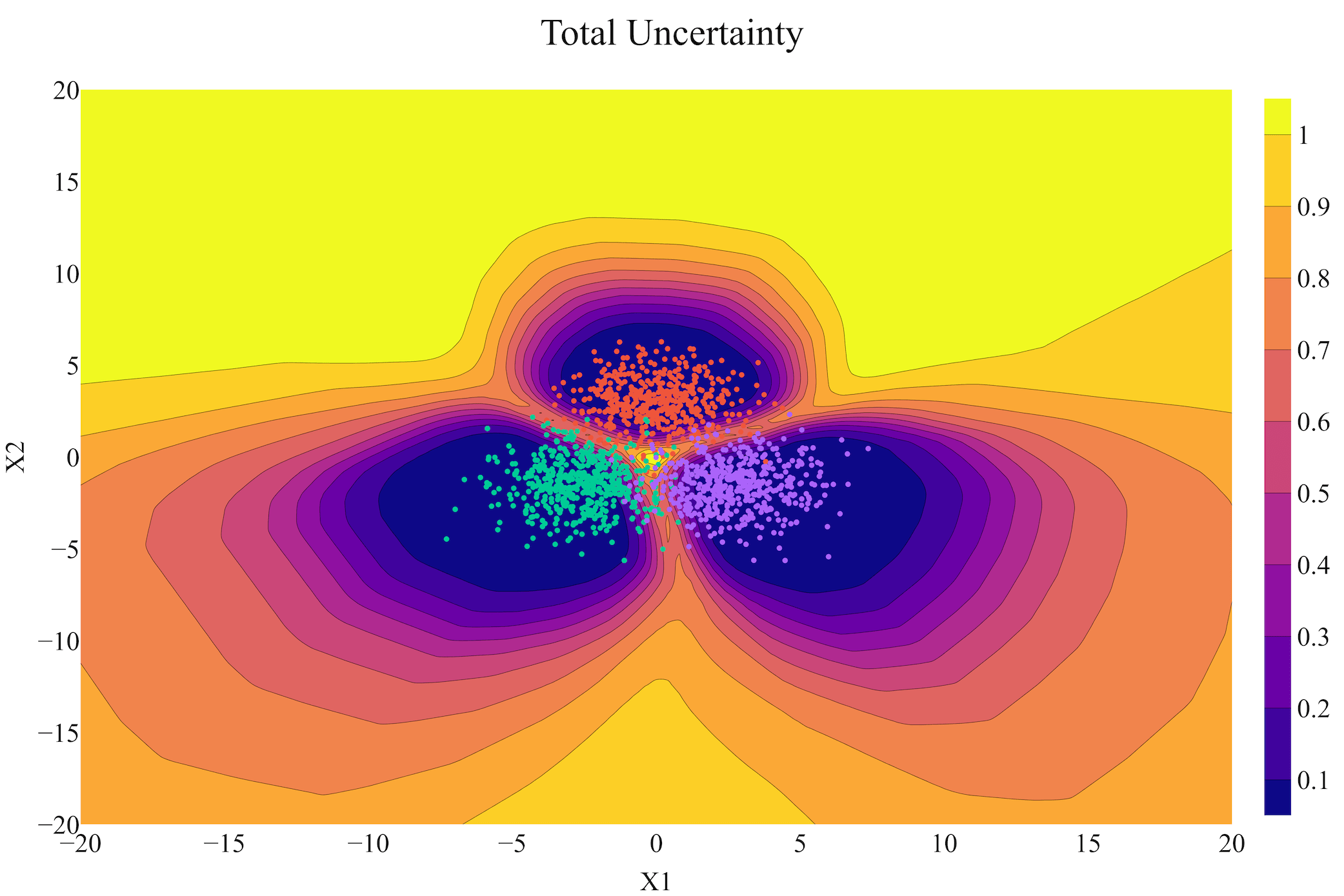}    
        \caption{Total Uncertainty}   
        \label{fig:bacoun-2}    
    \end{subfigure}
    \begin{subfigure}[t]{0.24\textwidth}    
        \includegraphics[width=1.0\textwidth]{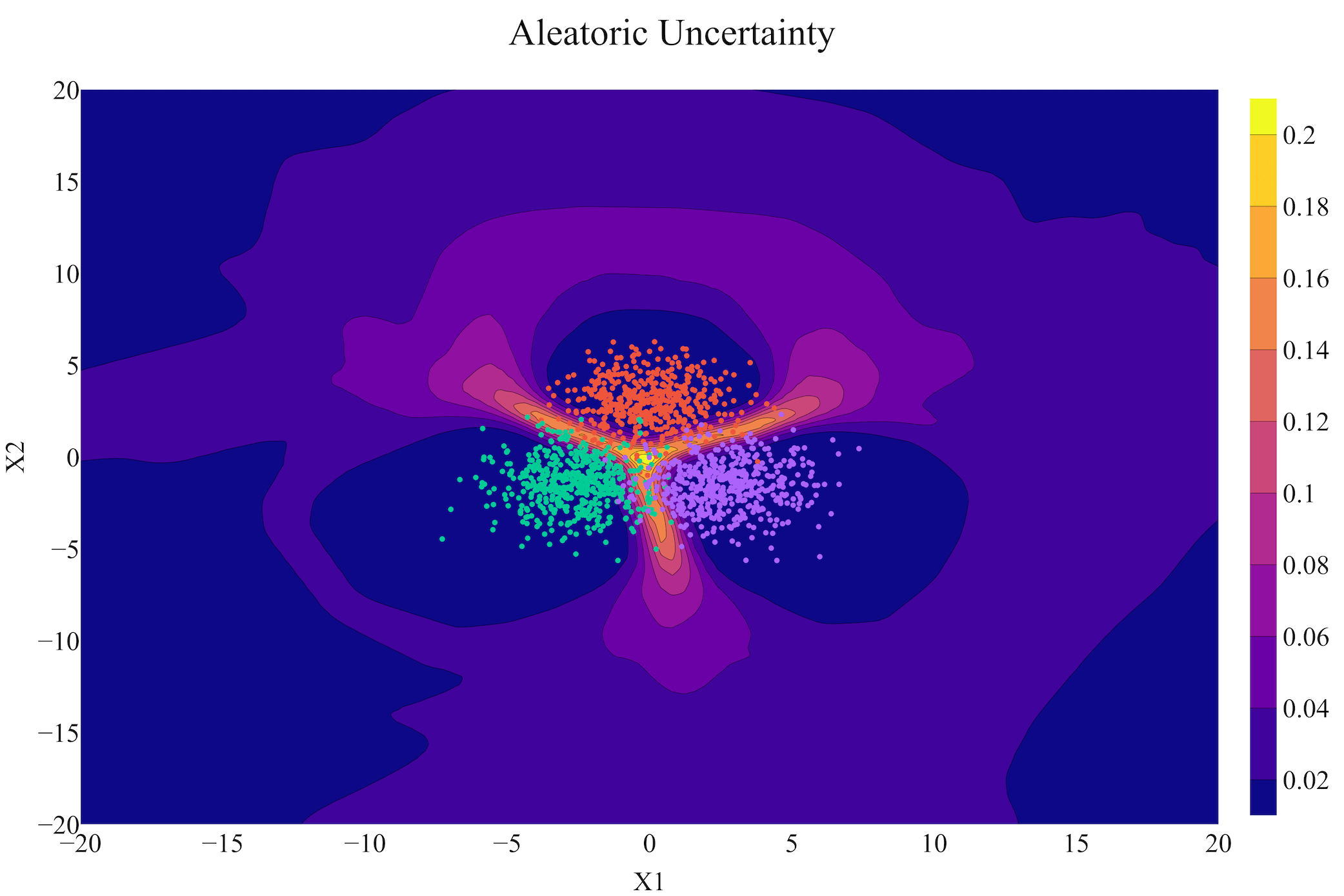}    
        \caption{Aleatoric Uncertainty}
        \label{fig:bacoun-3}    
    \end{subfigure}
    \begin{subfigure}[t]{0.24\textwidth}    
        \includegraphics[width=1.0\textwidth]{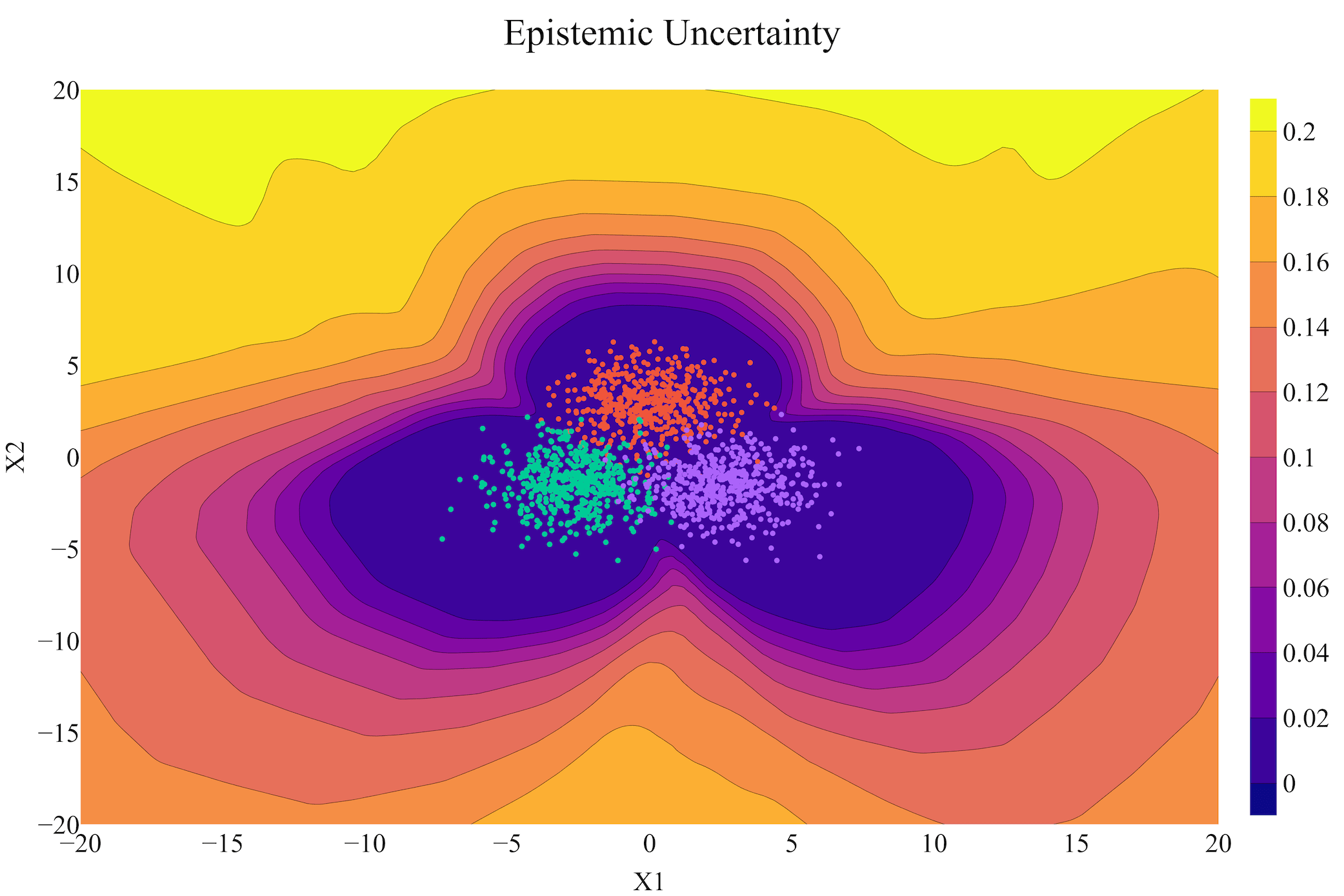}    
        \caption{Epistemic Uncertainty}
        \label{fig:bacoun-4}    
    \end{subfigure}
    
     \caption{Uncertainty decomposition provided by BaCOUn along with the decision boundaries learned in Step 2 of BaCOUn's training procedure (a). BaCOUn captures OOD uncertainty and decomposes uncertainty accurately, giving high aleatoric uncertainty in regions of class overlap and high epistemic uncertainty in regions far from the observed data.}
    \label{fig:bacoun}
    \end{figure*}

\section{Experiments}

\textbf{Baselines.} In our experiments, we compare BaCOUn with NLM, BNN, and MC-Dropout on both synthetic and real datasets. See Appendix \ref{appendix:implementation} for details on each method. 

\textbf{Evaluation.} We evaluate the fit of the learned models using AUC.
We quantitatively evaluate epistemic and aleatoric uncertainties using in-distribution and OOD points. On image data we additionally provide a qualitative evaluation.  
We consider a model successful if it is able to have high AUC while being uncertain about OOD points.
We use the uncertainty decomposition used by \cite{depeweg2017decomposition}, shown in Appendix \ref{appendix:measures} Equation \ref{appendix:uncertainty-eq}.

\textbf{Datasets}
We conduct experiments on two synthetic datasets: the Gaussian Mixture Model dataset and the Moons dataset (Appendix \ref{appendix:data}). 
We also conducted experiments on two real datasets: \textit{Wine Quality} \cite{CorCer09} and MNIST \cite{lecun-mnisthandwrittendigit-2010}. 
On \textit{Wine Quality}, we train our classifier on only two of the $K$ classes and used the remaining classes as OOD examples at test-time together with artificially created OOD points (details in Appendix \ref{sec:realdatasets}).
On MNIST, we used examples from other datasets (CIFAR \cite{cifar}, USPS \cite{usps}, EMNIST \cite{cohen2017emnist}), as well as boundary points created from the normalizing flow) as example OOD points for evaluation. See Appendix \ref{appendix:quantitative} and \ref{appendix:qualitative} for details. 

\setlength{\textfloatsep}{8pt plus 1.0pt minus 2.0pt}
\begin{figure}[h]
\centering
\begin{subfigure}{0.95\linewidth}
\includegraphics[width=\linewidth]{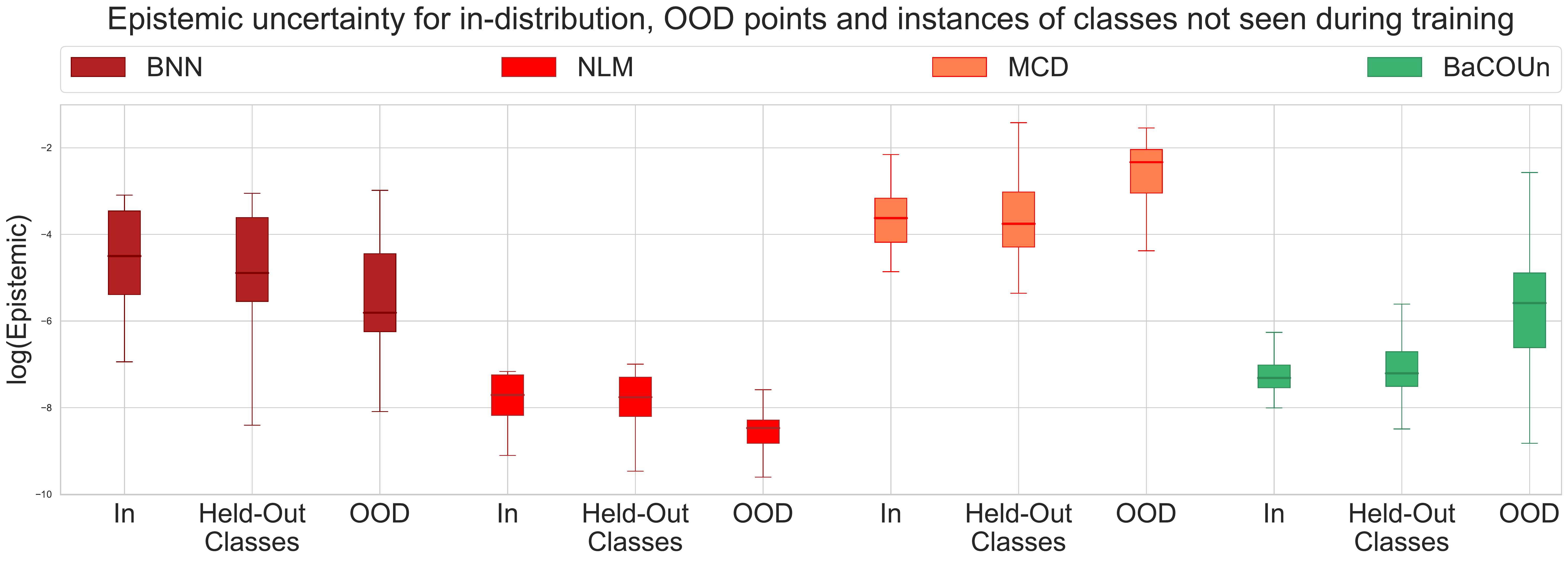}
\end{subfigure}
\begin{subfigure}{0.95\linewidth}
    \includegraphics[width=\linewidth]{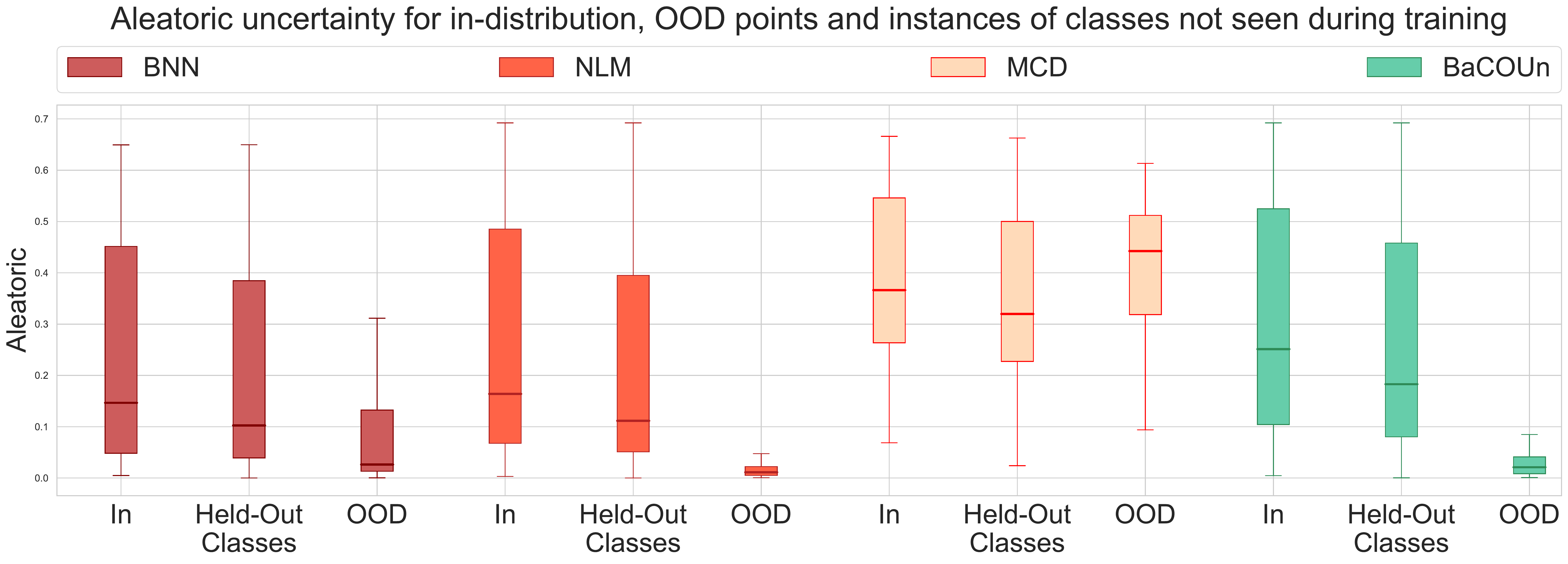}
\end{subfigure}
\caption{Epistemic and aleatoric uncertainty of BNN, NLM, MC-Dropout and BaCOUn for in-distribution points, instances of held-out classes and OOD points on \textit{Wine Quality} dataset. BaCOUn provides reliable OOD uncertainty while maintaining accurate aleatoric uncertainty. Specifically, epistemic uncertainty is maximized while aleatoric uncertainty is minimized for OOD points.}
\label{fig:wine}
\end{figure}

\setlength{\textfloatsep}{6pt plus 1.0pt minus 2.0pt}
\begin{figure}[h]
\includegraphics[width=\linewidth]{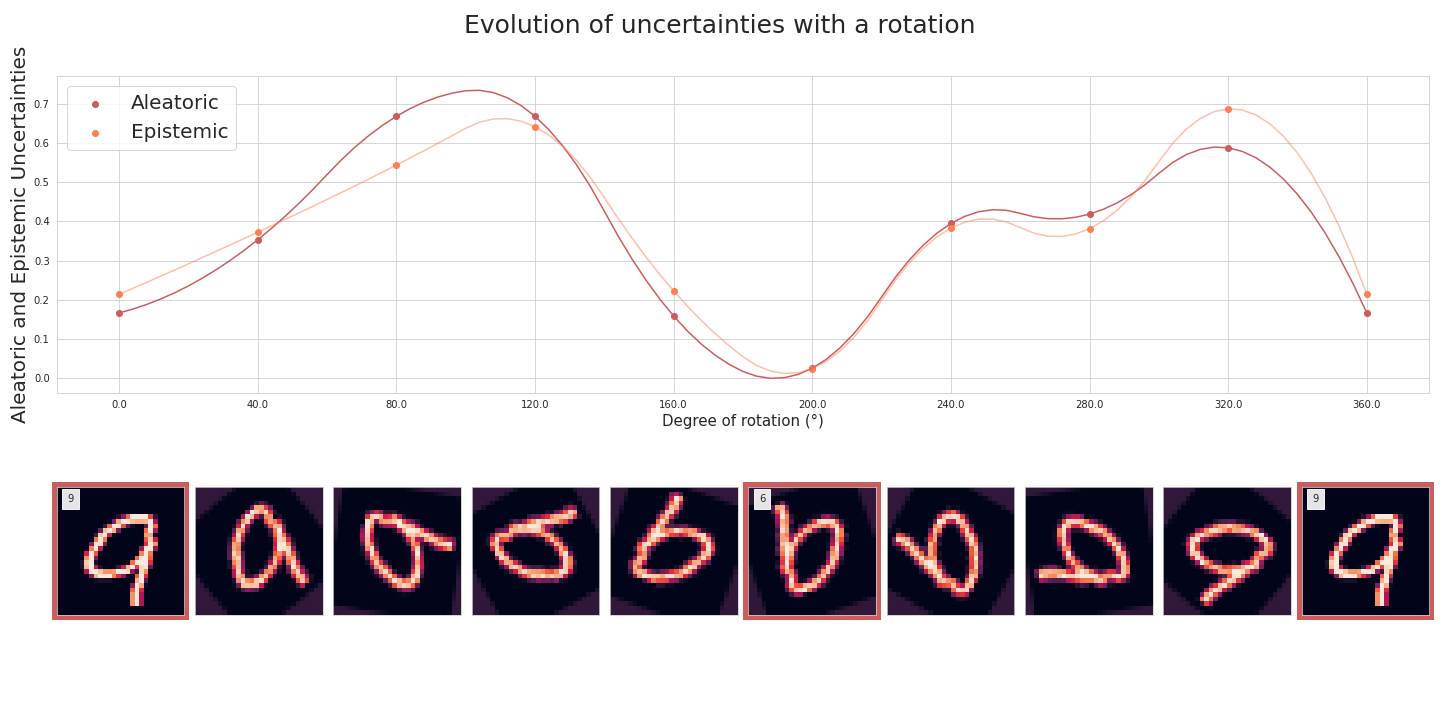}
\caption{Aleatoric and epistemic uncertainty of rotated MNIST images. When the rotated image looks like a known digit, both types of uncertainties decrease; both uncertainties increase when the rotated image does not look like a digit.}
\label{fig:rotated-9}
\end{figure}

\section{Results}
\textbf{BaCOUn can capture OOD uncertainty whereas baselines struggle.} Figure \ref{fig:bacoun-1} shows the decision boundaries learned by the $K+1$-class classifier trained in Step 2 of BaCOUn. The classifier labels points on the data boundary as OOD and this labeling generalizes to regions far from the data. The final hidden layer of this classifier encodes these decision boundaries. Consequently, NLM trained with BaCOUn expresses higher uncertainty far from the training data, recovering the optimal GP behaviour. Appendix \ref{appendix:qualitative} Figure \ref{app_fig:moons_bacoun} shows that BaCOUn captures OOD uncertainties even in more complex data settings like the Moons dataset. Appendix \ref{appendix:qualitative} Figure \ref{app_fig:moons_mcd} and Figure \ref{app_fig:moons_nlm} show that baselines significantly underestimate OOD uncertainty.

\textbf{BaCOUn provides simple decomposition of total uncertainty.} Using BaCOUn, we are able to reliably decompose the total predictive uncertainty into: epistemic uncertainty (Figure \ref{fig:bacoun-4}), arising due to lack of knowledge in a region of the feature space, and aleatoric uncertainty (\ref{fig:bacoun-3}), arising due to class overlap. To our knowledge, there is no easy way to decompose GP predictive uncertainty.

\textbf{BacOUn provides interpretable uncertainty in real data settings.}
Using \textit{Wine Quality}, we show that BaCOUn provides uncertainties on real datasets that align with human intuition. Figure \ref{fig:wine} shows that BaCOUn returns higher epistemic uncertainty for instances of held-out classes (classes 3,4,8,9) compared to the uncertainty for in-distribution points (class 5 or 7). As expected, BaCOUn gives even higher epistemic uncertainty estimates for OOD points. Only MC-Dropout achieves comparable performance in terms of detecting OOD points using epistemic uncertainty; however it fails to provide interpretable aleatoric uncertainty, since it is maximized for OOD points where classes do not overlap. In contrast, BaCOUn provides aleatoric uncertainty that also aligns with human intuition: points from held-out classes have comparable aleatoric uncertainty to in-distribution samples, however aleatoric uncertainty is much lower for OOD points, since they are sampled from regions far from where the observed data overlap. 

In Appendix \ref{appendix:quantitative}, Figure \ref{app_fig:mnist_epistemic}, we show that BaCOUn's uncertainty estimates for OOD data sampled from EMNIST, CIFAR and USPS align well with human intuition: BaCOUn gives high epistemic uncertainty  for CIFAR samples (which do not contain digit-like images), whereas EMNIST samples (which has some data overlap with MNIST) are given the highest aleatoric uncertainty. In comparison, baselines either provide epistemic uncertainties that do not align with intuition or aleatoric uncertainties that are hard to interpret.

In a qualitative experiment, we track changes in BaCOUn's resulting uncertainties as we rotate an image (Figure \ref{fig:rotated-9}). We see that, as expected, epistemic uncertainty is low when the  rotated image looks like a digit and is high when it does not. Aleatoric uncertainty is high when a rotated image can be ambiguously classified and low when it cannot. In Appendix \ref{appendix:qualitative}, we describe additional similar experiments.

\section{Conclusion}
In this paper, we show that BNNs, NLMs and MC-Dropout underestimate OOD uncertainty in classification tasks. We propose a novel Bayesian classifier, BaCOUn, which explicitly encourages posterior to capture OOD uncertainty. On synthetic and real datasets with complex patterns and sources of uncertainty, we demonstrate the ability of epistemic uncertainty produced by BaCOUn to distinguish in-distribution and OOD points. We further show that the uncertainty decomposition provided by BaCOUn aligns with human intuition on image datasets.

\newpage
\section{Acknowledgments}
DV, TG, and WP are supported by the Harvard Institute of
Applied Computational Sciences. YY acknowledges support from NIH 5T32LM012411-04 and from IBM Research.


\nocite{langley00}
\bibliography{ref}
\bibliographystyle{icml2020}

\clearpage
\appendix

\section{Data}
\label{appendix:data}
\subsection{Synthetic Datasets}
\paragraph{Gaussian Mixture}
We generate points from a Gaussian Mixture Model with 3 components corresponding to 3 different classes. The three classes are distributed with means $(0, 2)$, $(-\sqrt{3}, -1)$ and $(\sqrt{3}, -1)$ respectively, and identical isotropic covariance matrices $\sigma \cdot \mathbb{I}_{\mathbb{R}^2}$. We set $\sigma = 3 $ in order to obtain class overlap. We generate 500 points per cluster, for a total of 1500 training points.

We generate 2000 OOD points for BaCOUn by sampling from a circle of radius 3 bounding the data distribution in latent space, and then mapping the points into input space using BaCOUn's Normalizing Flow. See \ref{appendix:flows} for more details about the generation of OOD points.

Uncertainties of BaCOUn and baseline methods  are presented in \ref{appendix:quantitative}.

\paragraph{Moon-shaped data}
We use \texttt{sklearn} \cite{scikit-learn} and its built-in ``two-moons" data generator to generate $K=2$ classes, with 2000 points per class (for a total of 4000 data points), using a noise parameter of 0.05. For this dataset, we generate 3000 OOD points. Uncertainties for various models are presented in \ref{appendix:quantitative} and \ref{appendix:qualitative}. 

\subsection{Real Datasets}
\label{sec:realdatasets}
\paragraph{UCI Wine Quality}
This dataset \cite{CorCer09} contains instances of wines with ratings ranging from 3 to 9. All features are standardized to have zero mean and unit standard deviation.

We use class 5, 7 during training and we test the model's predictions and corresponding uncertainties on held-out instances of class 5, 7, as well as on instances of held-out classes 3,4,8,9 and artificially created OOD points.

OOD points are created by inflating the first dimension with Gaussian noise centered at 10 with unit variance (i.e. the first dimension is larger by 10 standard deviations on average compared to in-distribution points). Boundary points used to train BaCOUn were generated by mapping shells of radii $5$ and $5.1$ in latent space into input space (see \ref{appendix:flows} for details). 

Model uncertainties are presented in \ref{appendix:quantitative}. 

\paragraph{Images}
We use instances from MNIST \cite{lecun-mnisthandwrittendigit-2010} as in-distribution points and points from USPS \cite{usps}, EMNIST \cite{cohen2017emnist}, CIFAR \cite{cifar} as OOD instances. All datasets are normalized using the training set's statistics, converted to grayscale and the dimensions reduced to $28 \times 28$ when necessary.

USPS consists of digits, similar to MNIST, however the images present patterns different from MNIST. Therefore, we intuitively expect that epistemic and aleatoric uncertainties should be relatively high for USPS instances compared to in-distribution MNIST digits. EMNIST consists of some digit images which are also part of MNIST, but also includes other categories such as hand-written characters. On instances from EMNIST, we similarly expect both epistemic and aleatoric uncertainties to be relatively high. 

CIFAR on the other hand is made of images which are significantly different from the training data. Intuitively, on CIFAR instances, a well-calibrated model would output higher epistemic uncertainty and lower aleatoric uncertainty than for the rest of the image datasets.

Also, a model with well-calibrated uncertainty should assign higher epistemic uncertainty to points on the data boundary than to in-distribution points.  The boundaries used to train BaCOUn are generated by training one normalizing flow for each class in the training data (id est 10 normalizing flows) and mapping a shell of radii $3$ and $3.1$ in the latent space into input space for each flow (see \ref{appendix:flows} for details). In total, we generate 10,000 OOD samples.

Model uncertainties are presented in \ref{appendix:quantitative} and \ref{appendix:qualitative}. 

\paragraph{Selecting the number of OOD points}
The number of OOD points is selected with a simple heuristic: the OOD class should be (approximately) balanced with the K initial classes in order to avoid issues related to class imbalance. Depending on the number of classes in the data, we may want to generate more OOD samples than the number of samples in each class (small K) in order to properly bound the points in that class. As the number of classes K grows, we ensure that the OOD samples do not over-dominate the dataset. In addition, potentially negative outcomes caused by the inclusion of a new class in the dataset (e.g degraded training accuracy) is mitigated when we fit a Bayesian classifier that accounts for the uncertainty over the position of the generated boundaries.

\section{Implementation details}
\label{appendix:implementation}
We implement deterministic Neural Networks using PyTorch \cite{pytorch}. Bayesian layers are implemented using Pyro \cite{bingham2018pyro}.

\paragraph{GMM, Moons}
We use similar architectures for experiments on both synthetic datasets. \textbf{BaCOUn} and \textbf{NLM} are based on a feed-forward network with 4 hidden layers of dimensions $64, 64, 64, 1024$. We use a batch size of $256$, a L2 penalty term of $1$ and a dropout rate \cite{JMLR:v15:srivastava14a} of $0.1$.

The deterministic network is trained for 500 epochs, using Adam optimizer \cite{kingma2014adam} and a learning rate of $10^{-3}$. 

Inference for the Bayesian logistic regression built on top of the features of the deterministic network is performed using Hamiltonian Monte-Carlo (HMC) \cite{neal2011mcmc}, in particular, we use the No-U-Turn Sampler (NUTS) \cite{hoffman2011nouturn}. We run NUTS for 1000 iterations, with 100 burn-in steps. 

\textbf{MC-Dropout} and the \textbf{BNN} baselines are based on a network of 4 hidden layers with dimension $512$. A dropout rate of 0.2 is used for MC-Dropout, with batch sizes, number of epochs, and learning rates  the same as for the deterministic network. Inference for BNN is performed using BBVI \cite{ranganath2014black}, with 10,000 total iterations, and a learning rate of $10^{-2}$. 
LeakyRelu activation was chosen (for all networks). In all cases, 200 samples from the posterior distribution were used.

We run each baseline with the following architectures: 
\begin{itemize}
    \item 2 hidden layers of size $64$
    \item 4 hidden layers of size $64$
    \item 4 hidden layers with sizes $64, 64, 64, 1024$
    \item 4 hidden layers with size $512$
\end{itemize}
The best architecture is selected for testing.

The Real-NVP normalizing flow used to generate data boundaries consists of $5$ blocks with $64$ hidden neurons each. Batch Normalization \cite{ioffe2015batch} is used to stabilize the training. We use a batch size of $256$, L2 penalty term of $4$ (GMM) and $5$ (Moons) and $200$ (GMM) or $100$ (Moons) epochs. The learning rate for training the flows is set to $10^{-3}$. Hyperparameters are selected using grid search.

We note that for a fair comparison we allow a larger parameters budgets for \textbf{MC-Dropout} and \textbf{BNN} trained with BBVI, since the speed of the inference for these models allows for larger architectures. In realistic settings, HMC sampling for BNNs is not efficient. On the other hand, for \textbf{BaCOUn} and \textbf{NLM}, exact inference is feasible using HMC.

It is worth noting that the expressiveness of BaCOUn framework depends on the width of the last layer of the base neural network. This is expected since the final layer of the neural network forms an information bottleneck making it harder to encode the boundaries of K+1 classes with a small number of nodes.

\paragraph{UCI Wine Quality} All models were trained using 1238 points of class 5 and 1238 of class 7. For BaCOUn training, we additionally generated 1400 boundary OOD points. 

The base for all models is a neural network with 4 hidden layers of 64 nodes each. The deterministic network is trained for 500 epochs with Adam optimizer and a learning rate of $10^{-4}$ and L2 penalty equal to 1. For \textbf{MC-Dropout}, we use a dropout rate of 0.4 and calculated posterior predictive mean over 500 samples. 

We train \textbf{BNN} with BBVI \cite{ranganath2014black} for 2000 iterations and calculated posterior predictive mean over 500 samples. The inference for Bayesian layer in \textbf{NLM} and \textbf{BaCOUn} is performed with NUTS for 1000 steps with a burn-in of 100 steps. The posterior predictive mean is computed over 500 samples from the Markov chain. 

To generate boundary points for BaCOUn framework, we use one Real-NVP of 5 blocks with 64 neurons per class, trained for 100 epochs, with batch size equal to 256, learning rate equal to $10^{-3}$ and L2 penalty equal to 1.

\paragraph{Image Datasets}
All methods except for BNN, use a neural network with 2 Convolutional layers (kernels of size 5), Max Pooling followed by either one or two hidden layers of size $50, 1024$ (\textbf{BaCOUn}) or $50$ (\textbf{NLM, MC-Dropout}). As a matter of fact, an additional layer does not yield better results for NLM and MC-Dropout.  

We use a batch size of $64$, L2 penalty term of $1$ and a dropout rate of $0.5$. The deterministic network is trained for 15 epochs, using Adam and a learning rate of $10^{-3}$. Inference for the Bayesian layer in \textbf{NLM} and \textbf{BaCOUn} is performed using BBVI (for all methods), with a learning rate of $10^{-2}$ and for 30,000 iterations.

One Normalizing Flow per class in the training data is used to generate boundary points. Each flow had the same architecture, that is 5 blocks with 64 hidden neurons each, trained for $250$ epochs, with a batch size of $512$, and a L2 penalty of $2$.


\section{Uncertainty measures}
\label{appendix:measures}
The output probabilities of a $K$-classes classifier can be considered as the parameters of a Categorical distribution. Thus, entropy of the predicted Categorical distribution is a measure of the total uncertainty in predictions of a traditional non-Bayesian classifier. However, given an ensemble of models (such as the models obtained by sampling from the posterior distribution over the parameters of a BNN), the entropy of the posterior predicted Categorical distribution is a measure of the total predictive uncertainty. 

The total uncertainty can be further decomposed into: aleatoric uncertainty, arising due to noise, and epistemic uncertainty, arising due to lack of knowledge when no data are observed on a given region of the feature space. We can calculate the expected entropy at a given point by averaging the entropies of the Categorical distributions predicted by each member of the ensemble. This is a measure of aleatoric uncertainty in predicting a single point. Finally, the Mutual Information (MI) between the categorical label $y$ and the parameters of the model $W $(i.e. in deep learning models these are the weights of the network), calculated by subtracting the expected entropy from total entropy, is a measure of the spread of the ensemble, and in turn a measure of epistemic uncertainty \cite{depeweg2017decomposition}:
\begin{align}
    \begin{split}
        \underbrace{I(y,W|x^*, \mathcal{D})}_{\text{Epistemic Uncertainty}} = &\underbrace{\mathbb{H}[\mathbb{E}_{p(W|\mathcal{D})}[p(y|x^*,W)]]}_{\text{Total Uncertainty}} \\ 
        &- \underbrace{\mathbb{E}_{p(W|\mathcal{D})}[\mathbb{H}[p(y|x^*,W)]]}_{\text{Expected Aleatoric Uncertainty}}
    \end{split}
    \label{appendix:uncertainty-eq}
\end{align}

\section{Generating boundary points with Normalizing Flows}
\label{appendix:flows}
Normalizing Flows \cite{rezende,realnvp,huang2018neural} are generative models that use the change of variables theorem to transform a simple distribution $p(z)$ into a complex distribution $p(x)$ using a function $f^{-1}_\phi(z)$, composed of a sequence of simple invertible transforms with cheaply computable Jacobians (for tractability).
Since directly sampling from the ``boundary" of the data manifold in the original data space is a complicated, we propose to train a normalizing flow to maximize the log likelihood of the $x$'s, and then use the boundary in the latent space of the flow to generate data lying on the boundary of the data-space.
Specifically, since we assume a simple distribution over the latent space, $f(X) \approx \mathcal{N}(0, \sigma)$ in $\mathcal{Z}$, we propose to sample $\tilde{z} \in Bound_{\mathcal{Z}}$, the latent boundary and use our flow to map it into the input space, obtaining $\tilde{x} = f^{-1}(\tilde{z})$. In practice, we choose $Bound_{\mathcal{Z}}$ to be a shell of given inner and outer radii, that is, $\{\tilde{z}| \alpha < d(0,\tilde{z} ) < \beta \}$. For example, in the case of a uni-dimensional standard Gaussian distribution, since approximately $99\%$ of the probability mass is  in the interval $[-3;3]$, we may want to invert points that are sampled at the boundary of this interval and we consider $ Bound_{\mathcal{Z}} = ]-3 - \eta; -3[ \cup ]3; 3+\eta [$, where $\eta$ is a smoothing term. 

\FloatBarrier
\section{Pathologies of the Neural Linear Model}
\label{appendix:pathologies}
In this section we show that in an NLM classifier, one generally does not learn features capable of expressing OOD uncertainty. In NLM training, one learns features $\phi_\theta$ by maximizing the log likelihood of the observed data, as shown in Equation \ref{eq:learned-basis}.
While the learned $\phi^*$ is good for classification, it is not capable of expressing the boundary of the high-mass region of the data. 
Specifically, the learned basis can express functions that separate the classes, but not functions that surround the data,
and as a result cannot model a distribution over decision boundaries that surrounnd the data (corresponding to OOD uncertainty).

\begin{figure*}[!h]
\centering
\begin{subfigure}[t]{0.49\linewidth}
\includegraphics[width=1.0\linewidth]{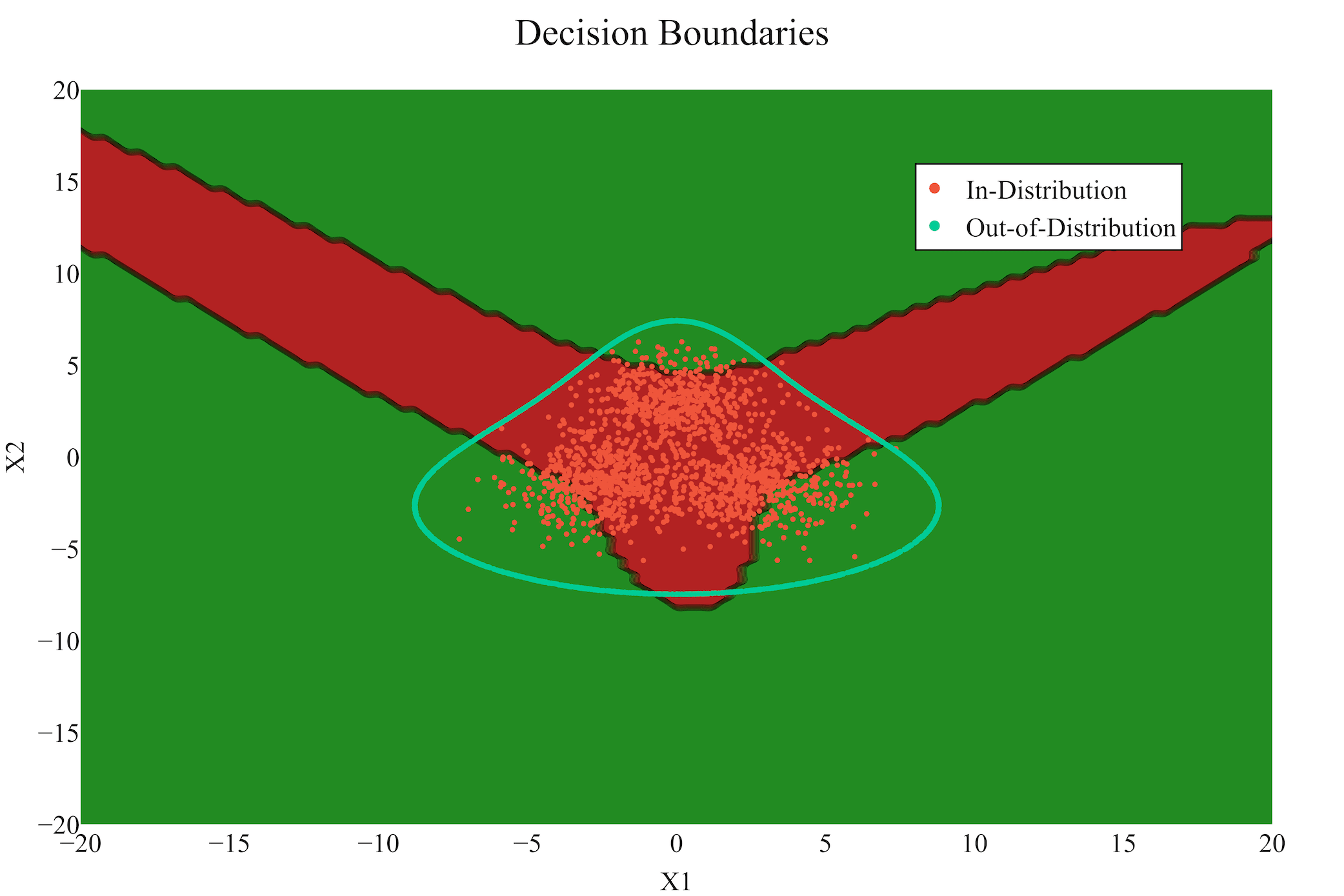}
\end{subfigure}%
\begin{subfigure}[t]{0.49\linewidth}
    \includegraphics[width=1.0\linewidth]{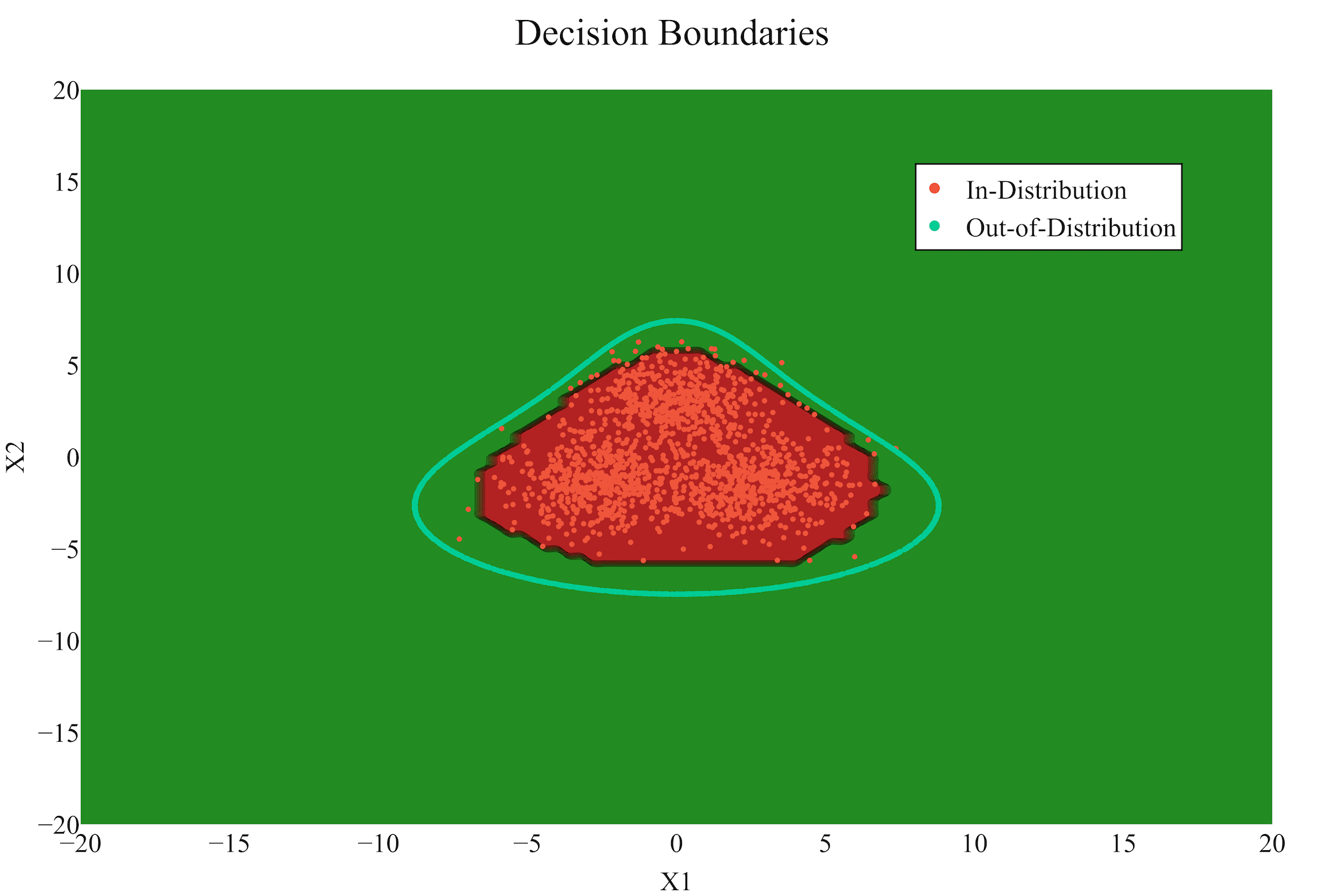}
\end{subfigure}
\caption{
Comparison of features used to classify OOD points vs. non-OOD points.
\textbf{Left}: We train the features to maximize the log likelihood of the data (i.e. to classify the points into their respective mixture components correctly).  We see that a classifier given these features fails to classify points as OOD vs. non-OOD fails. Specifically, far from the data the learned classification boundaries generalize along the boundaries between the original $K$ classes as opposed to between the in-distribution data and the OOD data. 
\textbf{Right}:  We train the features using BaCOUn's basis functions -- i.e. we train on the original $K$ classes (corresponding to the mixture components) and additionally a class representing the boundary of the data. We see that using those features, a classifier trained to classify points as OOD or non-OOD performs well. 
}
\label{app_fig:nlm_pathology}
\end{figure*}

We verify this hypothesis as follows: consider a synthetic dataset $\{ (x_n, y_n, b_n) \}_{n=1}^N$, where $x_n$ is the input, $y_n$ is the label, and $b_n$ is a binary indicator signaling whether the data-point is OOD or not. 
We train a classifier to predict $y | x$ (as in Equation \ref{eq:learned-basis}).
We then fix the features given by the last hidden layer of this classifier, $\phi(x; \theta^*)$, and train a logistic regressor to predict $b | x$. In other words, we train a regressor to map $\phi(x) \rightarrow b$.
As demonstrated in Figure \ref{app_fig:nlm_pathology}, this learned classifier is incapable of learning $b | x$ given features useful for predicting $y | x$.
Specifically, far from the data, the learned classification boundaries incorrectly generalize along the boundaries between the original $K$ classes, as opposed to between the in-distribution data and the OOD data. 
As a result, a Bayesian logistic regression model trained using this learned basis functions will not be able to express functions that surround the data, and thus will not model OOD uncertainty. 
In contrast, we train a classifier to predict $y,b|x$ and then use the learned features to predict $b | x$ (i.e $\phi(x) \rightarrow b$). In this case, since the basis is explicitly trained to predict the OOD points, the resultant classifier predicts $b | x$ accurately. 

We emphasize that these findings show a pathology of the training objective.
Thus, increasing the dimensionality and expressivity of the basis will not help.
In fact, we show later that BaCOUn is able to model OOD uncertainty while remaining predictive of $y$ for the same capacity networks.

Based on this evidence, we further hypothesize that with finite-sized BNNs, the uncertainty over the weights does not add enough diversity of features in order to model OOD uncertainty, in contrast to a GP which has an infinite dimensional basis.

\FloatBarrier
\section{Quantitative Results}
\label{appendix:quantitative}
\subsection{BaCOUn models OOD uncertainty}

\paragraph{GMM}
In Table \ref{table:gmm}, we present the aleatoric and epistemic uncertainties obtained by BaCOUn on the GMM dataset. We use a grid of the input space $[15, 17.5] \times [15, 17.5]$, ``far" from the observed data manifold (and consider it``OOD"). We also take points from high data density with no class overlap (``In") as well as points from a region with high class overlap (``Middle"). 

We take 100 points from each region, and report averages of the uncertainties obtained by the different methods, as well as the classification accuracy for each method. The models used are: the BaCOUn framework, a BNN fitted with BBVI, MC-Dropout (MCD) and a Gaussian Process (GP). All models provide relatively similar accuracies and aleatoric uncertainties in all regions. However, only the Gaussian Process (that we consider more or less as the gold standard) and BaCOUn provides reliable epistemic uncertainty in the OOD region. BNN, NLM and MC-Dropout dramatically fail by providing excessively low epistemic uncertainty for OOD points.  

\begin{table*}[htbp]
 \centering
 \begin{tabular}{*8c}
 \toprule Model & Accuracy & \multicolumn{3}{c}{Aleatoric}  & \multicolumn{3}{c}{Epistemic} \\
 \midrule
 {} & In-Dist & In & Out & Middle & In & Out & Middle \\
 \textbf{BaCOUn} & 0.94  &  0.11 & 0.15  & \textbf{0.30} & 0.03  & \textbf{0.74} & 0.08 \\
 \textbf{BNN}  & 0.93 & 0.10 & 8.7e-4 &  0.32
  &  0.07 & 0.04 & 0.11 \\
 \textbf{MCD}  & 0.94 & 0.123 & 5.8e-05 & 0.40 & 0.01 & 8.4e-06 & 0.03\\
 \textbf{GP} & 0.95 & 0.42  & 1.1 & 0.54 & 0.42  & 1.1 & 0.54\\
\bottomrule
\label{table:cfa2}
\end{tabular}
\caption{Results on synthetic GMM dataset. The uncertainty estimates are averaged over three types of regions: ``In'', which corresponds to in-distribution points in regions $p(x)$ is high and the classes do not overlap, ``Out'', which corresponds to OOD points far from the observed data, and ``Middle'', which correspond to in-distribution points in regions of high class overlap.}
\label{table:gmm}
\end{table*}

\subsection{BaCOUn yields interpretable aleatoric and epistemic uncertainty}
In both our experiments on the \textit{Wine Quality} dataset and Images datasets, BaCOUn is the only approach that provides interpretable and accurate epistemic and aleatoric uncertainty. Here, we define interpretable as aligned with human intuition about what is considered OOD.
Specifically, on both datasets we construct OOD examples using either held-out classes or other datasets, which a human would agree are considered OOD with respect to the training data.
We then check whether models distinguish the original training data (as in-distribution) from the held-out classes / different datasets (as OOD data).
In all cases, baselines systematically fail to decompose uncertainty accuracy, producing over-confident predictions,
while BaCOUn is able to correctly distinguish between OOD and in-distribution.

\begin{figure*}[h]
\centering
\begin{subfigure}{1.0\linewidth}
\includegraphics[width=\linewidth]{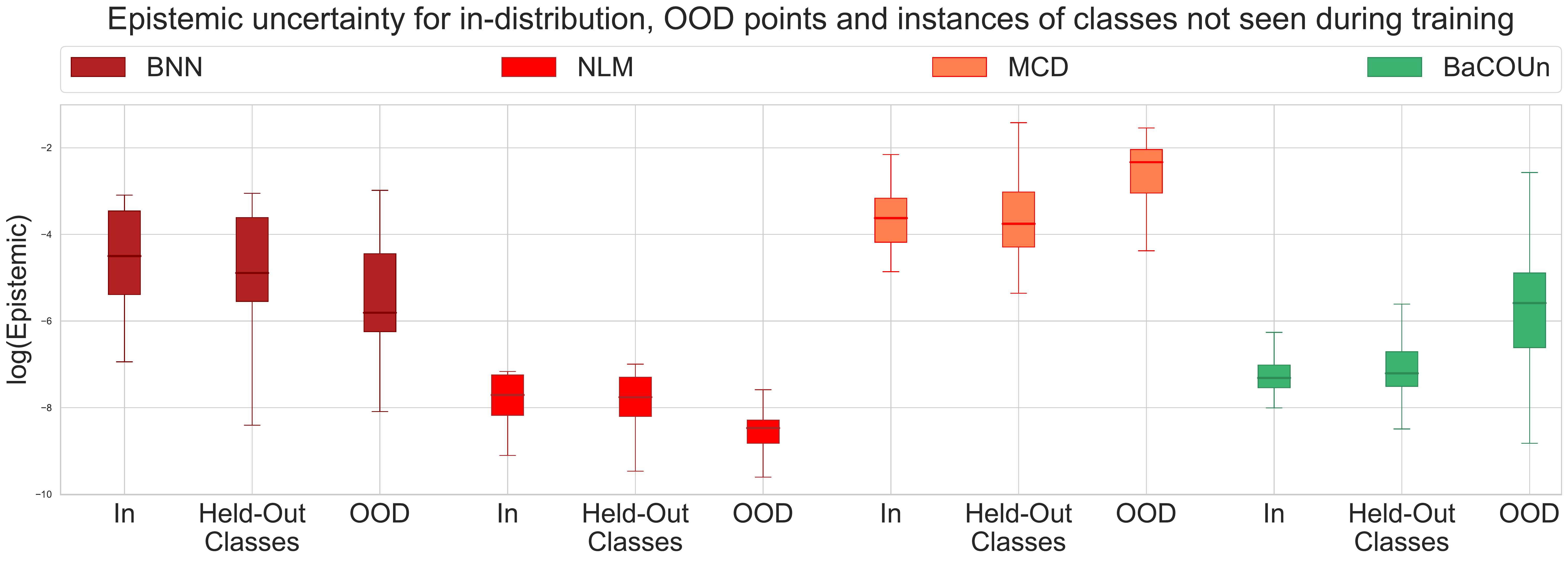}
\end{subfigure}
\begin{subfigure}{1.0\linewidth}
    \includegraphics[width=\linewidth]{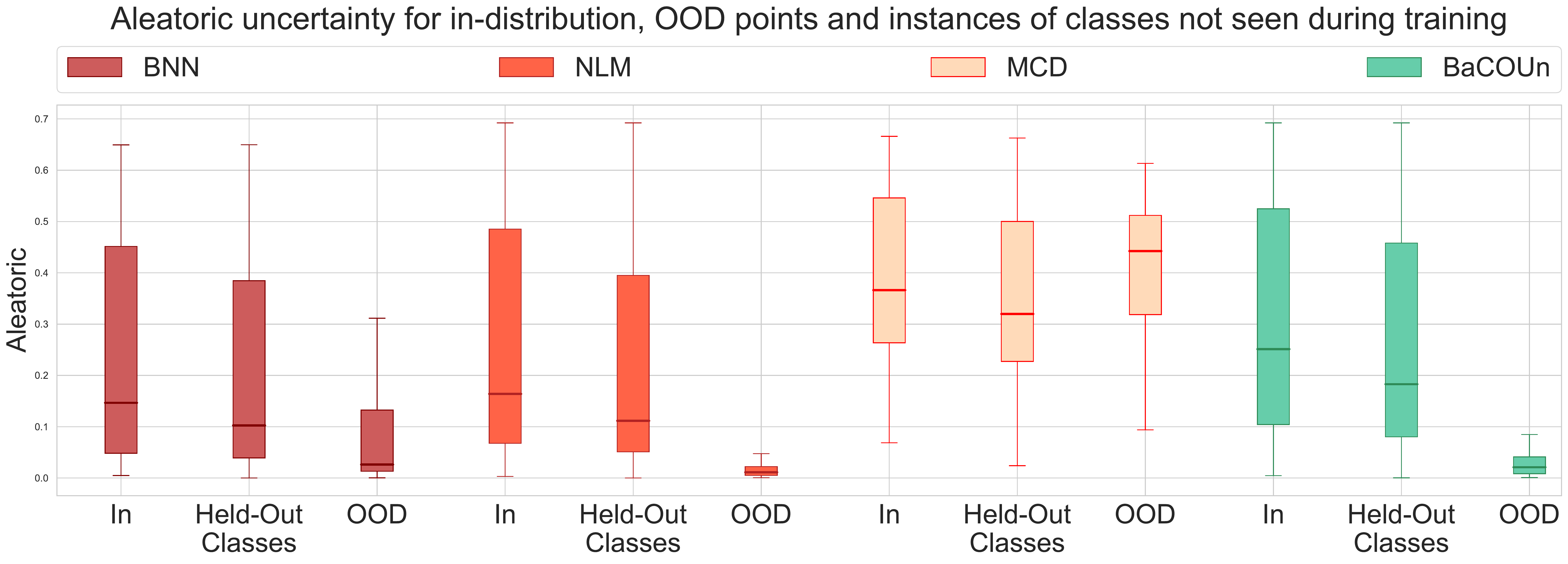}
\end{subfigure}
\caption{Epistemic and aleatoric uncertainty of BNN, NLM, MC-Dropout and BaCOUn for in-distribution points (i.e. held-out instances of class 5 and 7), instances of held-out classes (i.e. classes 3,4,8,9) and OOD points. BaCOUn provides reliable OOD uncertainty while maintaining accurate aleatoric uncertainty. Specifically, epistemic uncertainty is maximized for OOD points while aleatoric uncertainty is minimized in OOD regions where data have not been observed.}
\label{app_fig:wine}
\end{figure*}

\paragraph{Wine quality dataset}
In Table \ref{table:wine_ep} and  \ref{table:wine_al}, we show the accuracy (as a goodness of fit indicator) and decomposition of uncertainties obtained by BaCOUn and the baselines. In Figure \ref{app_fig:wine}, we additionally present boxplots that summarize the entire distribution of the uncertainties.
\begin{table*}[htbp]
 \centering
 \begin{tabular}{*9c}
 \toprule Model & AUC & \multicolumn{3}{c}{Epistemic ($\times 10^{-2}$)} & \multicolumn{2}{c}{Epistemic Pct Change From In}  \\
 \midrule
 {} & In & In &  Held-Out Classes & OOD & Held-Out Classes & OOD  \\
\textbf{BaCOUn} & \textbf{0.862} & 0.089 & 0.121  & 0.667  &\textbf{0.357}  & \textbf{6.479}  \\
\textbf{BNN}    & 0.854    & 1.719    & 1.492  & 0.96  & -0.132  & -0.441 \\
\textbf{MCD}    & 0.852    & 3.225 & 3.622 &  9.19   & 0.123    & 1.85    \\
\textbf{NLM}    & 0.856       & 0.049      & 0.046  & 0.02   & -0.047   & -0.581\\     
\bottomrule
\end{tabular}
\caption{Epistemic uncertainty obtained by BaCOUn, BNN, MCD and NLM on the \textit{Wine Quality} dataset. BaCOUn provides higher epistemic uncertainty for points of held-out classes and much higher epistemic uncertainty for OOD points compared to in-distribution points.}
\label{table:wine_ep}
\end{table*}

\begin{table*}[htbp]
 \centering
 \begin{tabular}{*9c}
 \toprule Model & \multicolumn{4}{c}{Aleatoric} & \multicolumn{3}{c}{Aleatoric Pct Change From In}  \\
 \midrule
 {} & In &  Held-Out (3,4,8,9) & Held-Out (6) & OOD & Held-Out (3,4,8,9) & Held-Out (6) & OOD \\
\textbf{BaCOUn} &  0.308  & 0.269  & 0.392  & 0.031  & -0.127  & \textbf{0.273} & \textbf{-0.899}  \\
\textbf{BNN}    & 0.245                 & 0.212           & 0.320   & 0.112                & -0.136         & 0.308            & -0.543 \\
\textbf{MCD}    & 0.392                 & 0.354           & 0.45   & 0.413                & -0.095         & 0.149            & 0.054    \\
\textbf{NLM}    &  0.266                 & 0.224           & 0.343  & 0.023                & -0.159         & 0.291            & -0.914   \\  
\bottomrule
\label{table:cfa2Wine}
\end{tabular}
\caption{Aleatoric uncertainty obtained by BaCOUn, BNN, MCD and NLM on the \textit{Wine Quality} dataset. BaCOUn provides aleatoric uncertainty which is higher for class 6 which contains many instances which are similar to both points of class 5 and class 7, while aleatoric uncertainty is much lower for OOD points far from where classes overlap.}
\label{table:wine_al}
\end{table*}

\begin{figure*}[!htb]
\centering
\includegraphics[width=\linewidth]{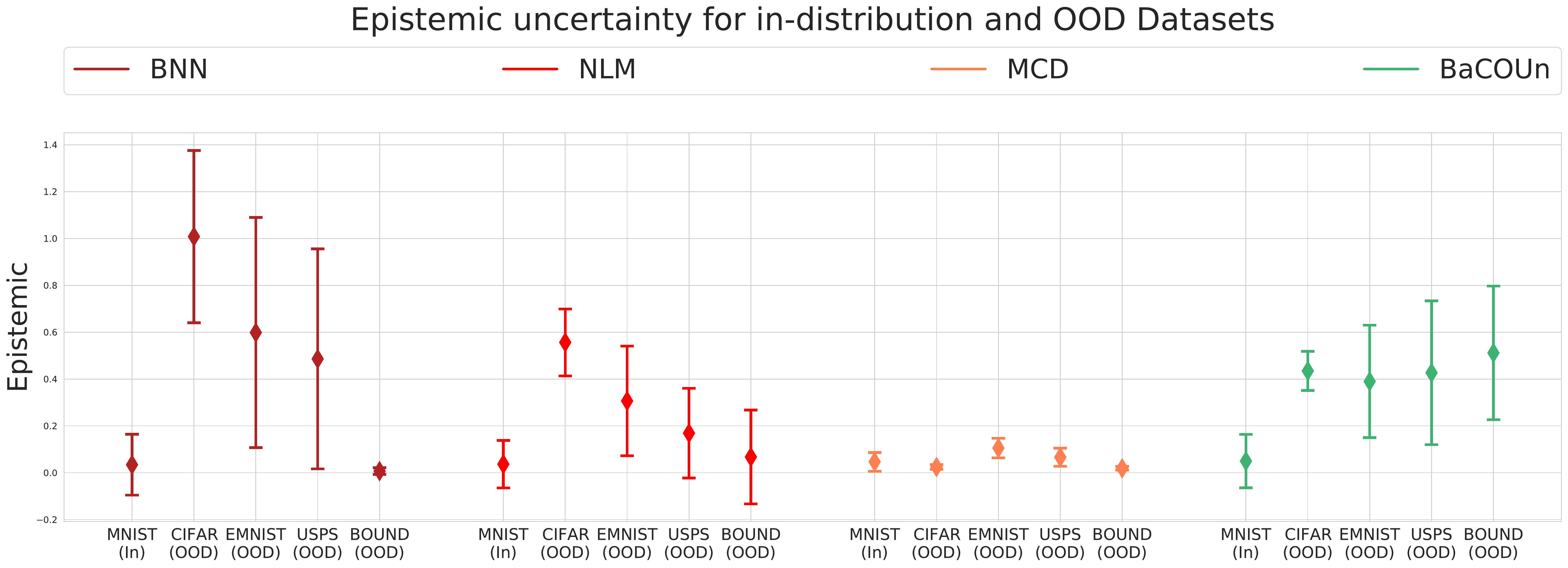}\par
\caption{Mean  $\pm$ standard deviation of the epistemic uncertainty obtained on the different datasets by BaCOUn and its competitors. Despite the apparent good intuitive results obtained by the NLM and a BNN run with BBVI, those results are to be contrasted with the inaccurate aleatoric uncertainty obtained below.}
\label{app_fig:mnist_epistemic}
\end{figure*}

\begin{figure*}[!h]
\centering
\includegraphics[width=\linewidth]{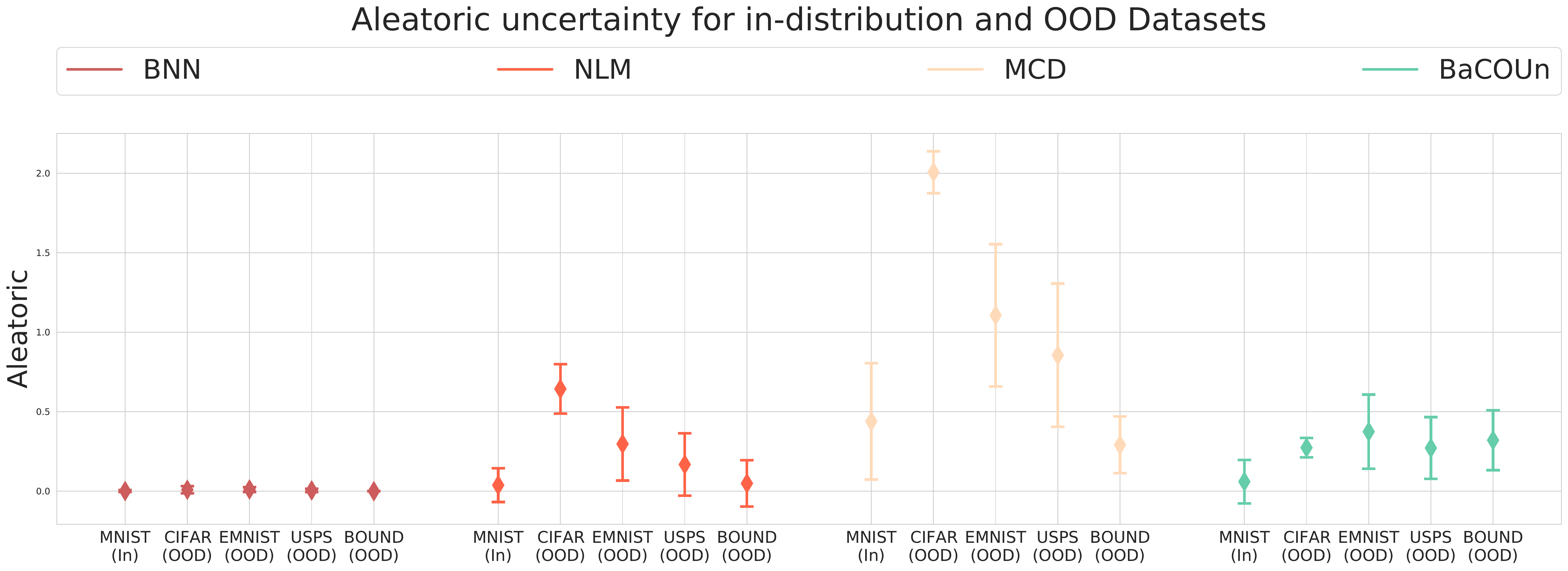}\par
\caption{Mean $\pm$ standard deviation of the aleatoric uncertainties obtained by different models on the image datasets. }
\label{app_fig:mnist_aleatoric}
\end{figure*}

\paragraph{Image Datasets}
To check whether BaCOUn's uncertainty decomposition aligns with human intuition, we used MNIST as the training data (and thus the ``in-distribution'' data), and other datasets (CIFAR, USPS, EMNIST) as OOD examples, since a human would regard those as qualitatively different from MNIST.
In Figure \ref{app_fig:mnist_epistemic} and \ref{app_fig:mnist_aleatoric}, we present the mean $\pm$ standard deviation of the epistemic and aleatoric uncertainties obtained on different image datasets. 
As shown in the figures, uncertainties obtained using MC-Dropout or a BNN (with BBVI) do not seem to align with human intuition, with either the epistemic or aleatoric uncertainty being predominant and the other uncertainty being negligible. The Neural Linear Model appears to have uncertainties that align better with human intuition; however, for NLM, the uncertainties on OOD points close to the data manifold (i.e. on the boundary) extremely low, implying overconfident predictions. We hypothesize that for robustness evaluations and downstream tasks, the Neural Linear Model could be easily attacked with adversarial samples, generated ``close enough" to the data manifold. Lastly, as desired, BaCOUn learns models with low epistemic uncertainty on the in-distribution points and high epistemic uncertainty on the OOD points (both close to and far from the data).

\FloatBarrier
\section{Qualitative Results}
\label{appendix:qualitative}
\subsection{BaCOUn provides GP-like behavior}

\begin{figure*}[!h]
\centering
\begin{subfigure}[t]{0.33\linewidth}
    \includegraphics[width=1.0\linewidth]{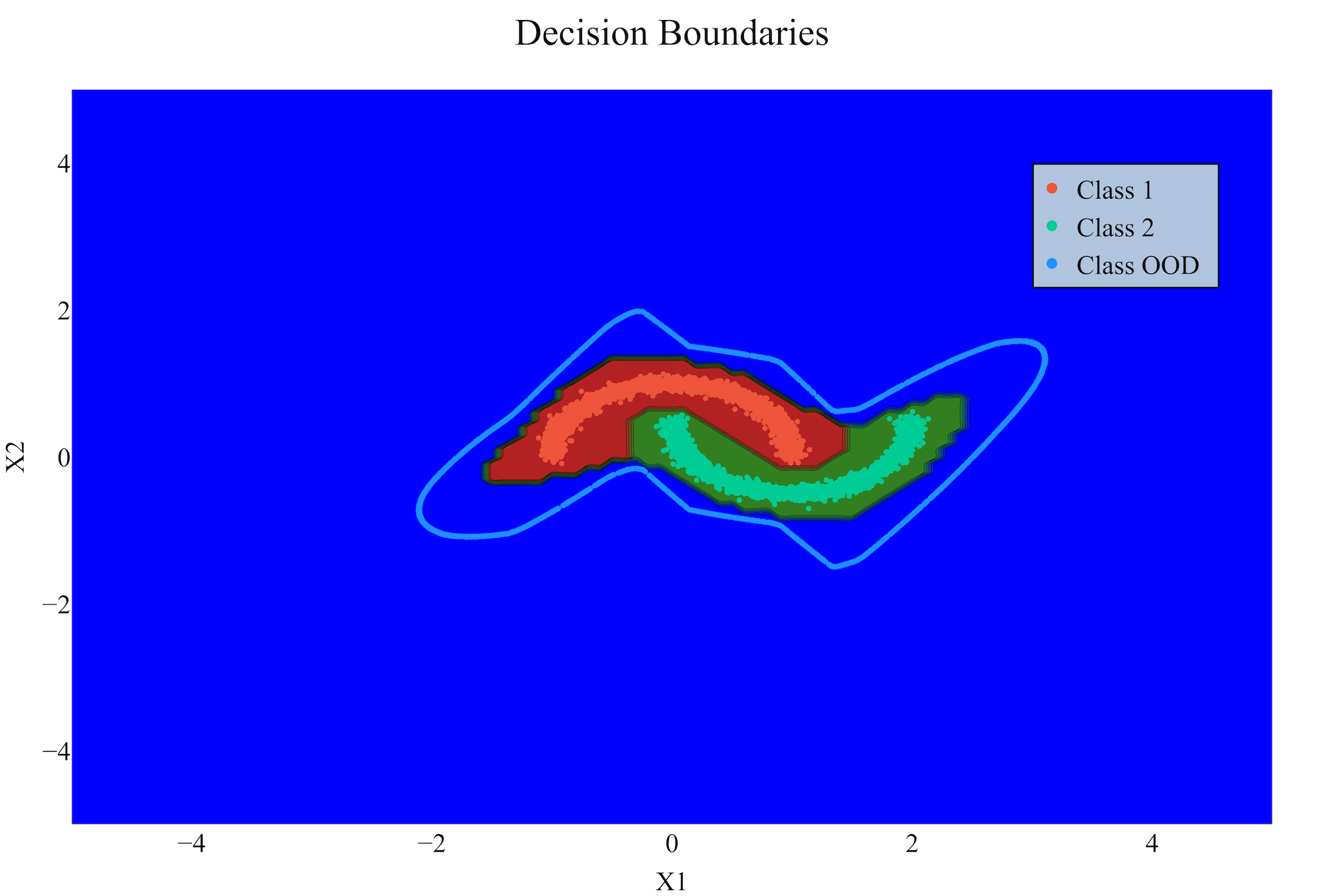}
    \caption{Decision Boundary}
\end{subfigure}
\begin{subfigure}[t]{0.33\linewidth}
    \includegraphics[width=1.0\linewidth]{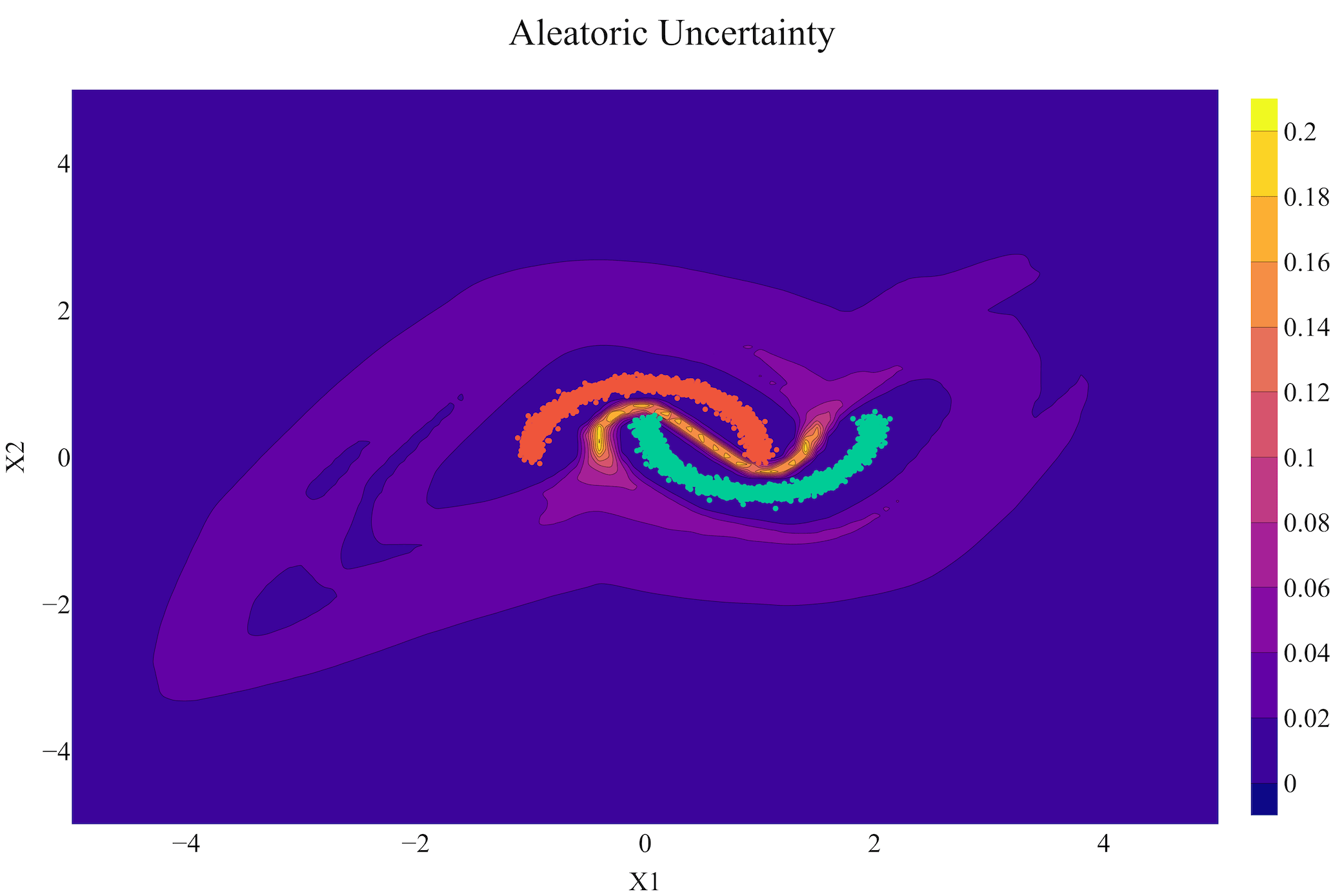}
    \caption{Aleatoric Uncertainty}
\end{subfigure}
\begin{subfigure}[t]{0.33\linewidth}
    \includegraphics[width=1.0\linewidth]{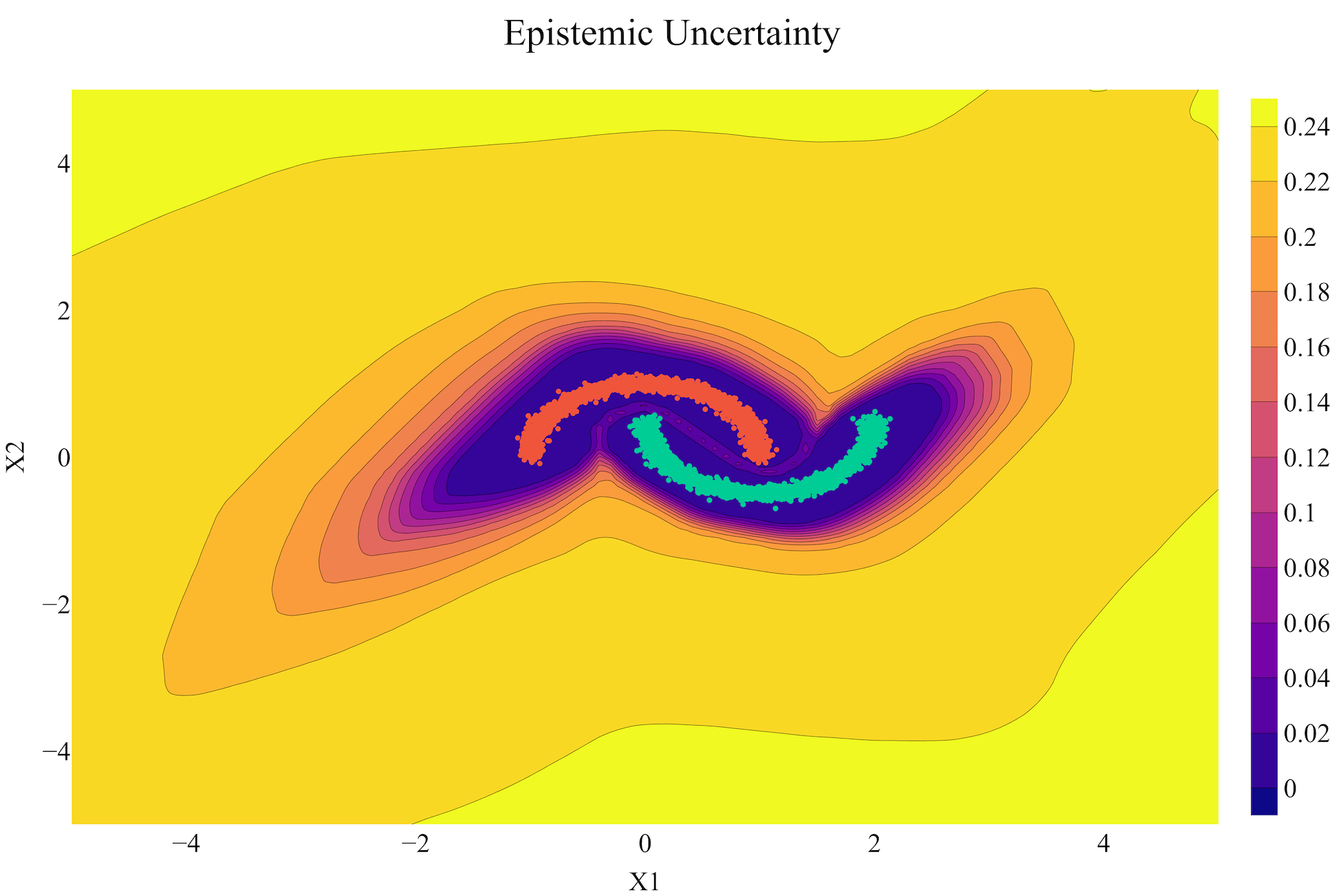}
    \caption{Epistemic Uncertainty}
\end{subfigure}
\caption{\textbf{BaCOUn Uncertainty Decomposition.} BaCOUn provides reliable OOD uncertainty while maintaining accurate aleatoric uncertainty in regions with high class overlap.}
\label{app_fig:moons_bacoun}
\end{figure*}

\begin{figure*}[!h]
\centering
 \begin{subfigure}[t]{0.33\linewidth}
 \includegraphics[width=1.0\linewidth]{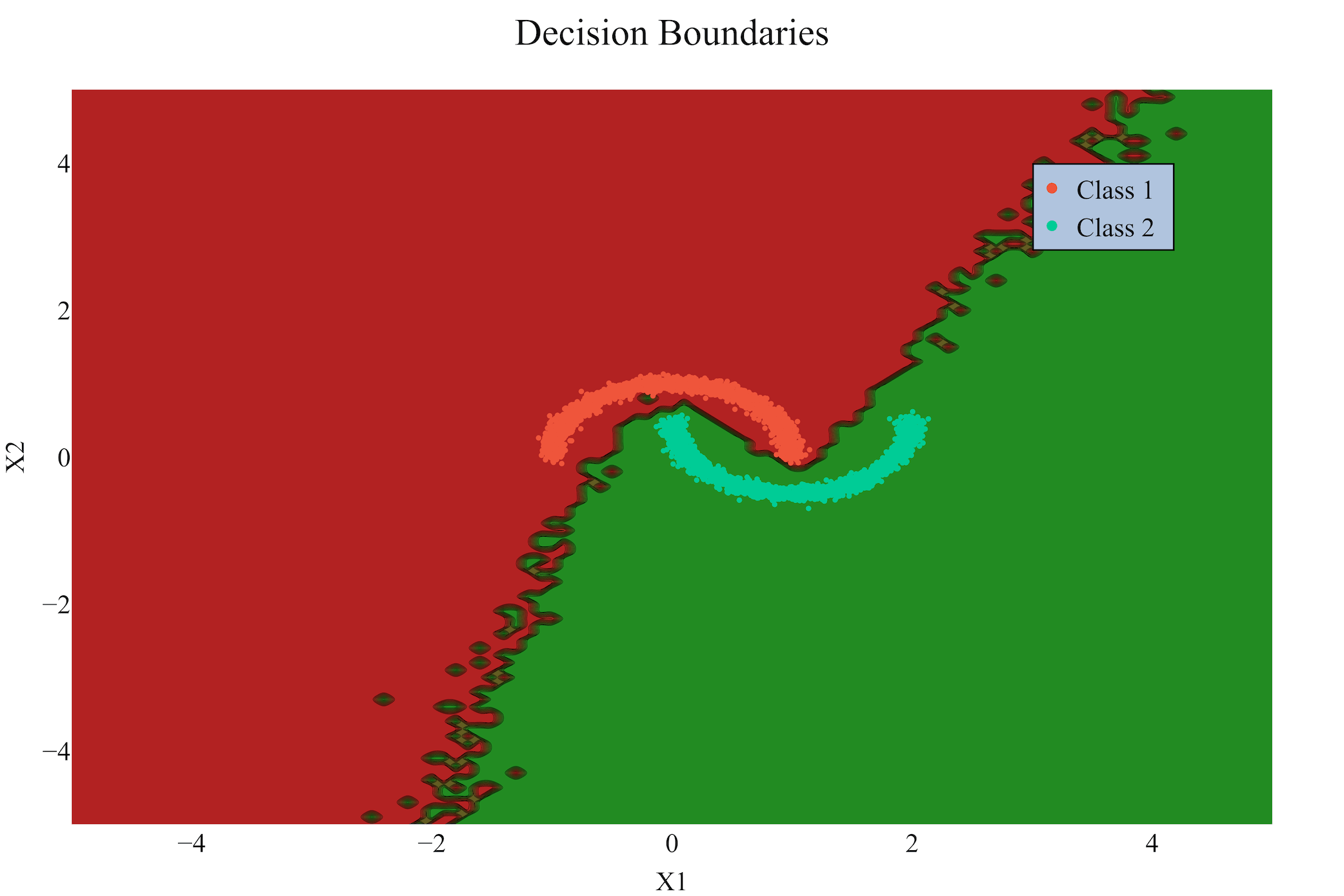}
 \caption{Decision Boundary}
\end{subfigure}
\begin{subfigure}[t]{0.33\linewidth}
    \includegraphics[width=1.0\linewidth]{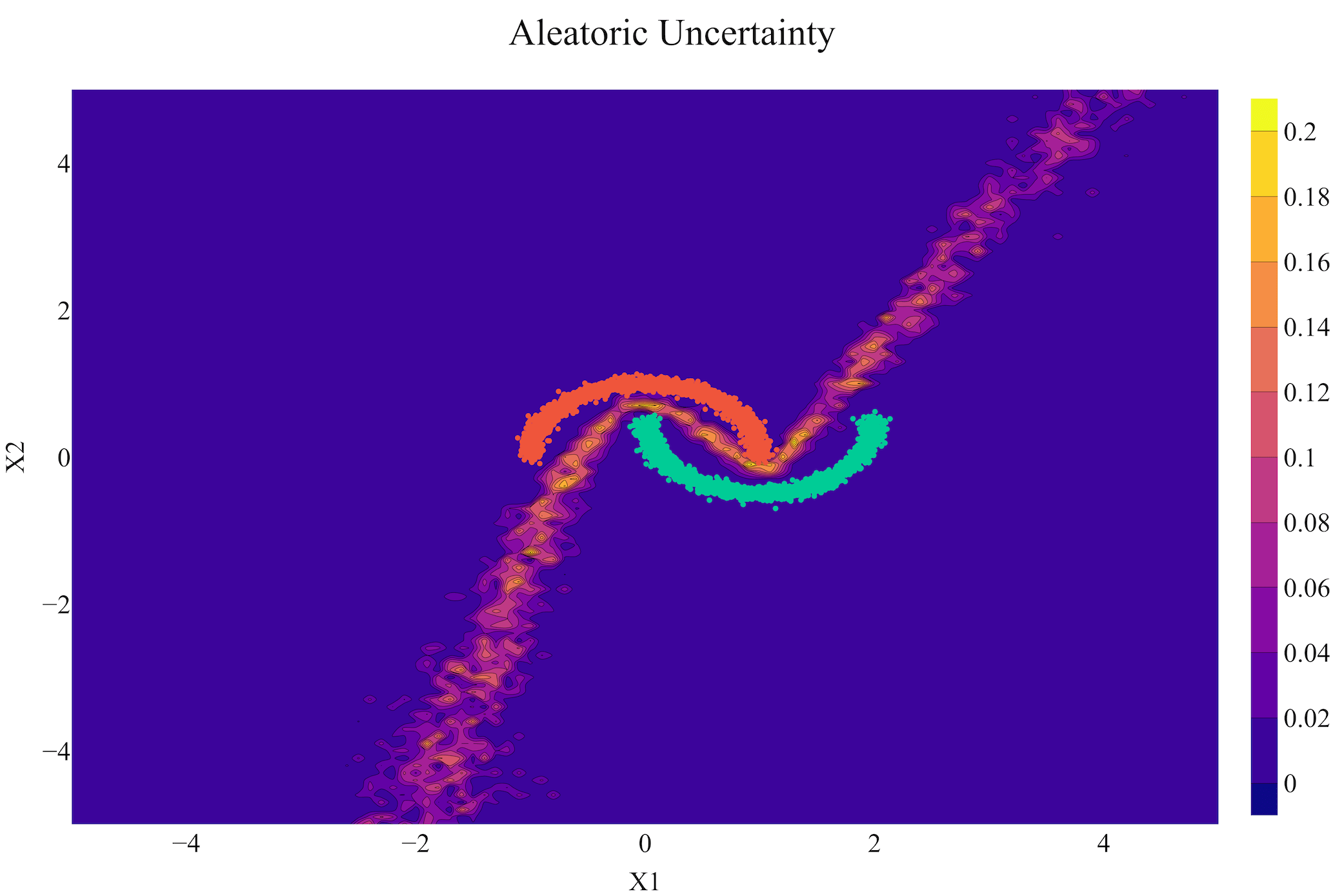}
    \caption{Aleatoric Uncertainty}
\end{subfigure}
\begin{subfigure}[t]{0.33\linewidth}
    \includegraphics[width=1.0\linewidth]{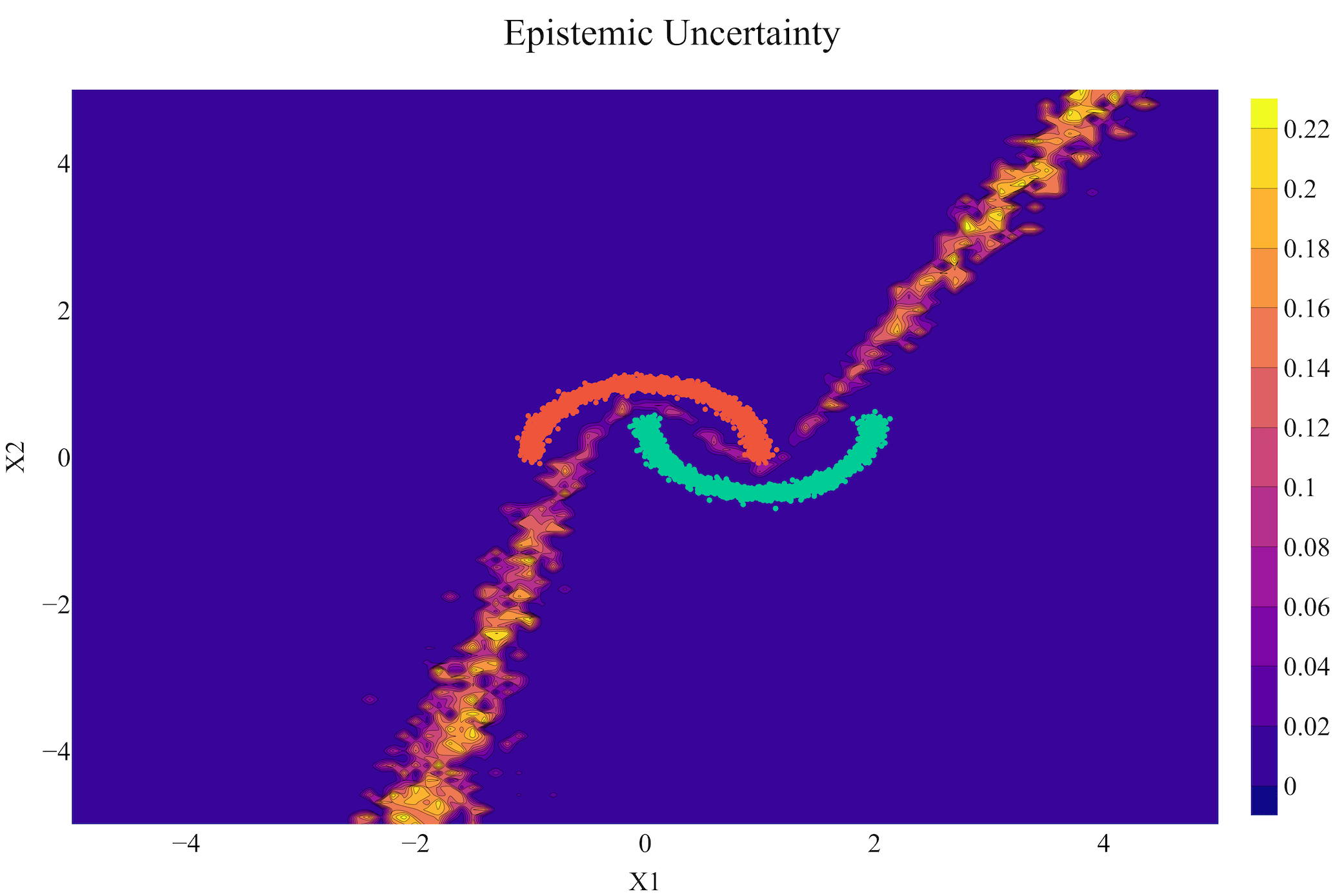}
    \caption{Epistemic Uncertainty}
\end{subfigure}
\caption{\textbf{MC-Dropout Uncertainty Decomposition.} The model overgeneralizes and produces overconfident predictions in regions of low-data density, failing to provide well-calibrated OOD uncertainty.}
\label{app_fig:moons_mcd}
\end{figure*}

\begin{figure*}[!h]
\centering
\begin{subfigure}[t]{0.33\linewidth}
\includegraphics[width=1.0\linewidth]{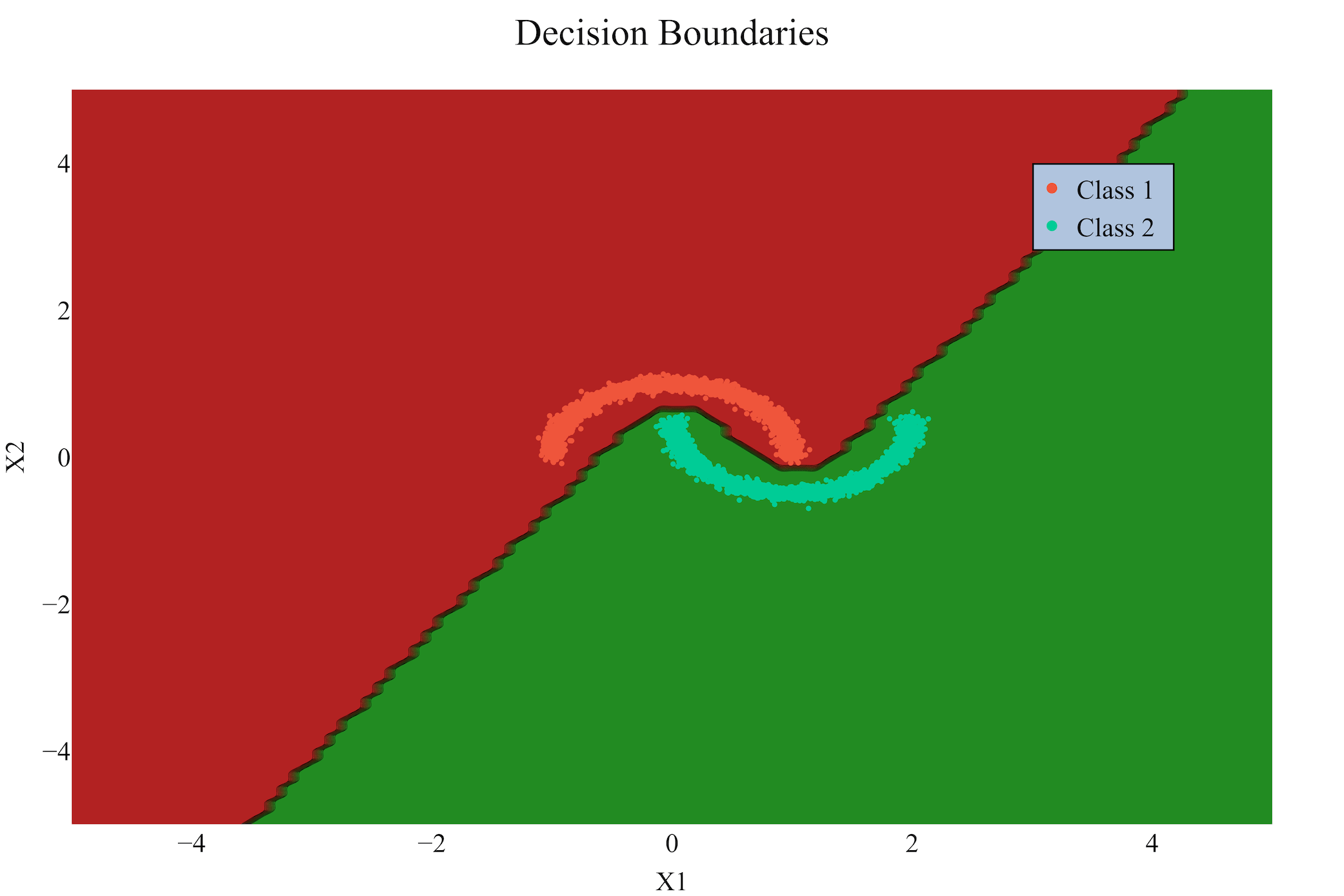}
\caption{Decision Boundary}
\end{subfigure}
\begin{subfigure}[t]{0.33\linewidth}
    \includegraphics[width=1.0\linewidth]{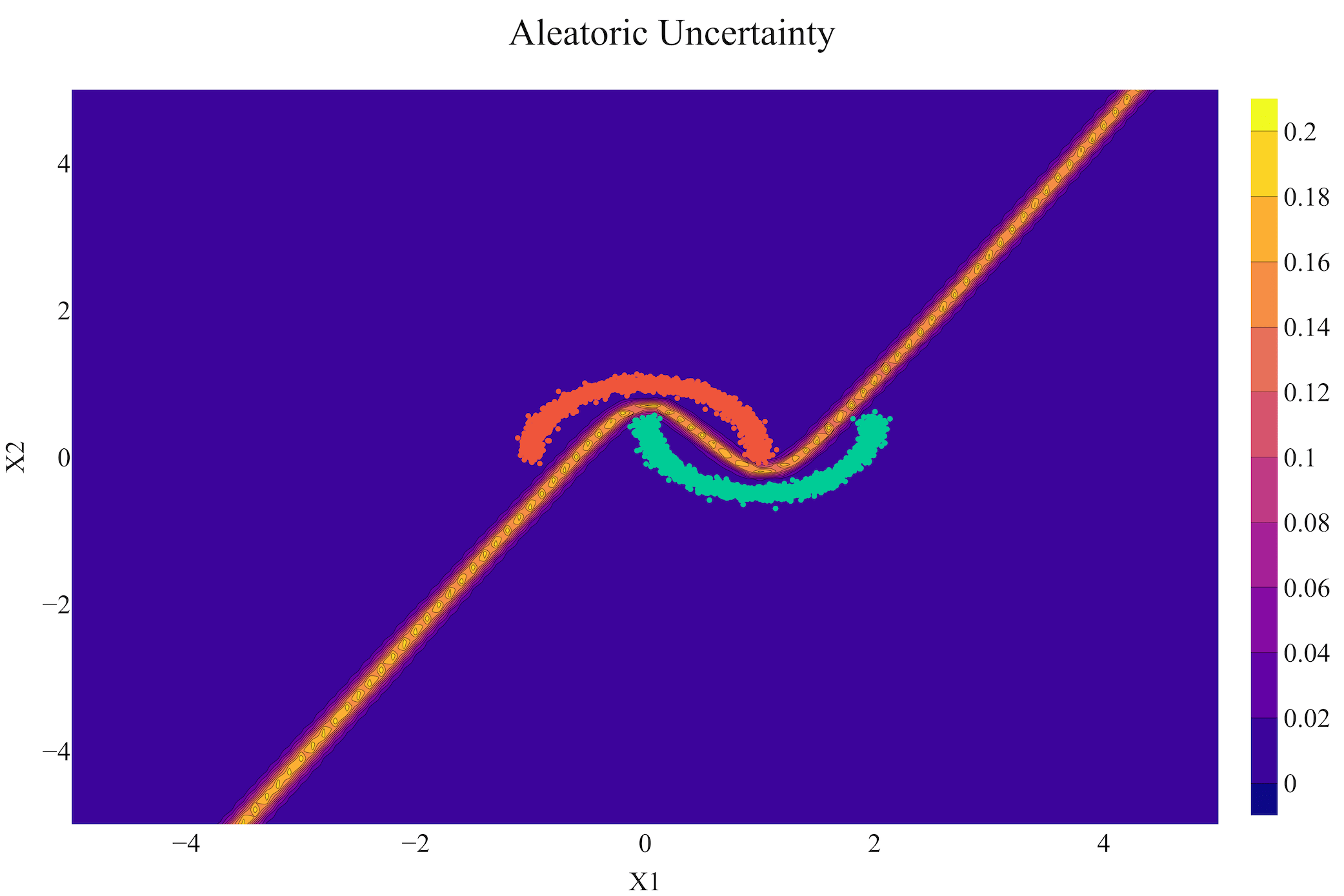}
    \caption{Aleatoric Uncertainty}
\end{subfigure}
\begin{subfigure}[t]{0.33\linewidth}
    \includegraphics[width=1.0\linewidth]{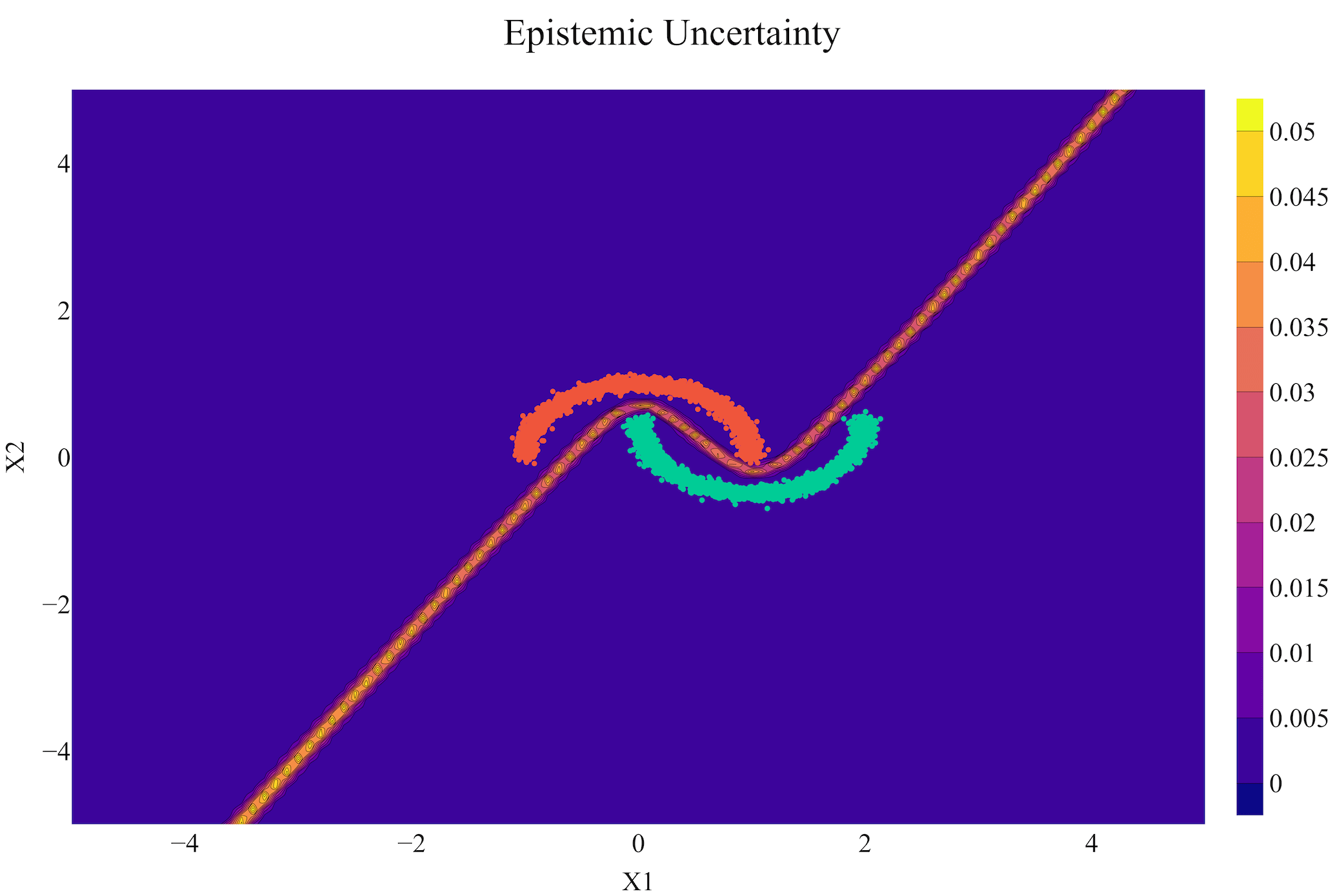}
    \caption{Epistemic Uncertainty}
\end{subfigure}
\caption{\textbf{NLM Uncertainty Decomposition.} Similarly to MC-Dropout, the model overgeneralizes and produces overconfident predictions in regions of low-data density, failing to provide well-calibrated OOD uncertainty.}
\label{app_fig:moons_nlm}
\end{figure*}

\begin{figure*}[!h]
\centering
\begin{subfigure}[t]{0.33\linewidth}
    \includegraphics[width=1.0\linewidth]{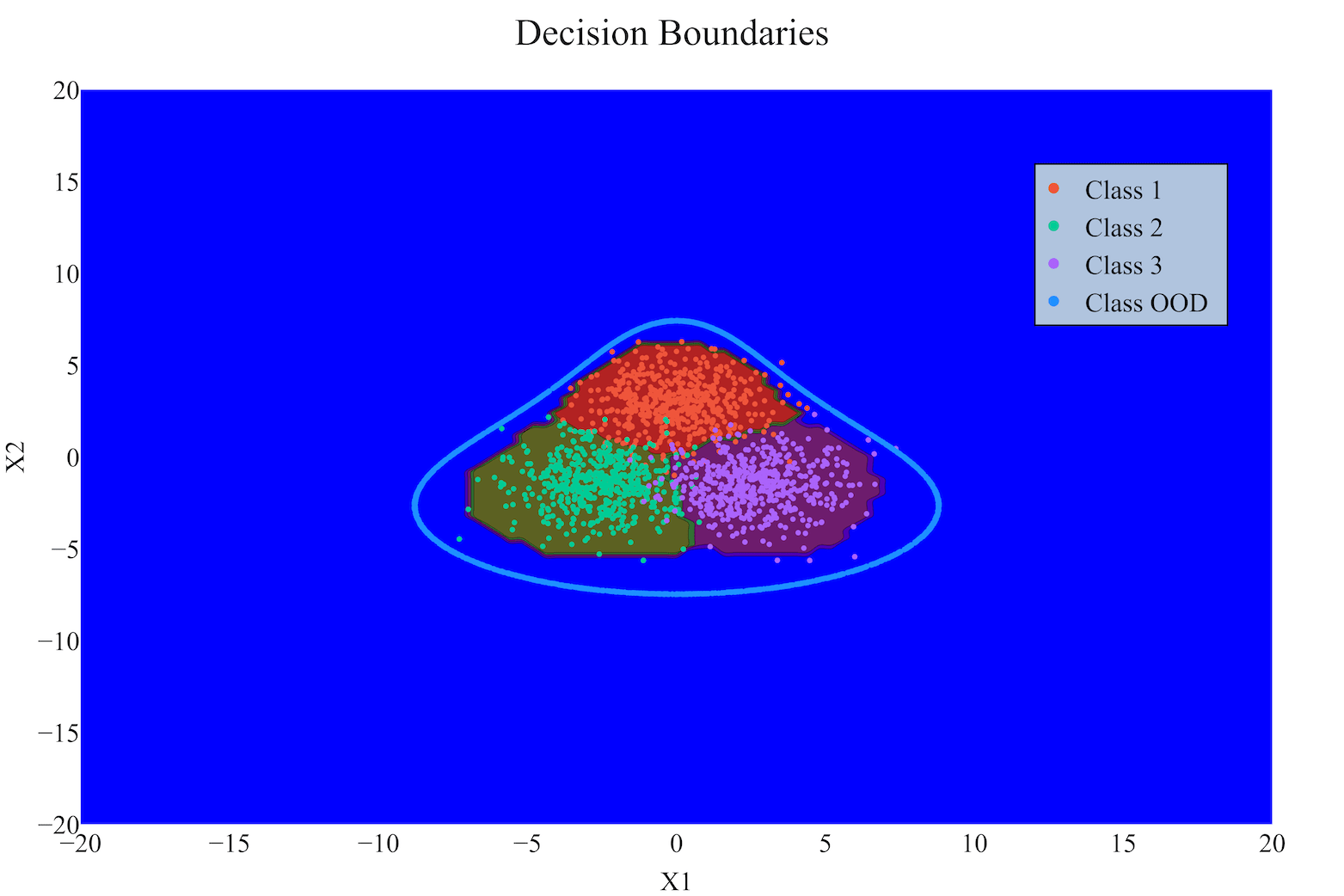}
    \caption{Decision Boundary}
\end{subfigure}
\begin{subfigure}[t]{0.33\linewidth}
    \includegraphics[width=1.0\linewidth]{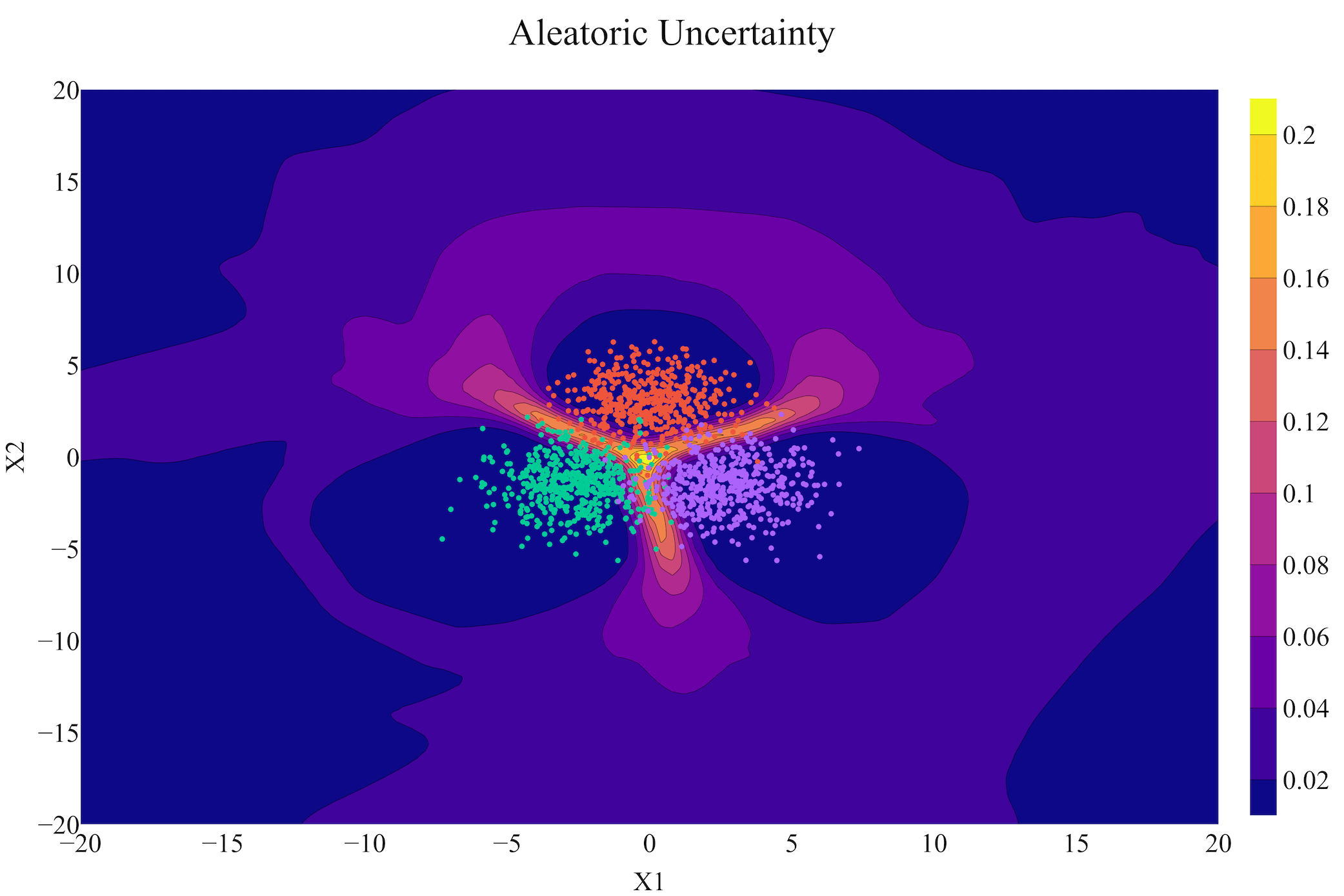}
    \caption{Aleatoric Uncertainty}
\end{subfigure}
\begin{subfigure}[t]{0.33\linewidth}
    \includegraphics[width=1.0\linewidth]{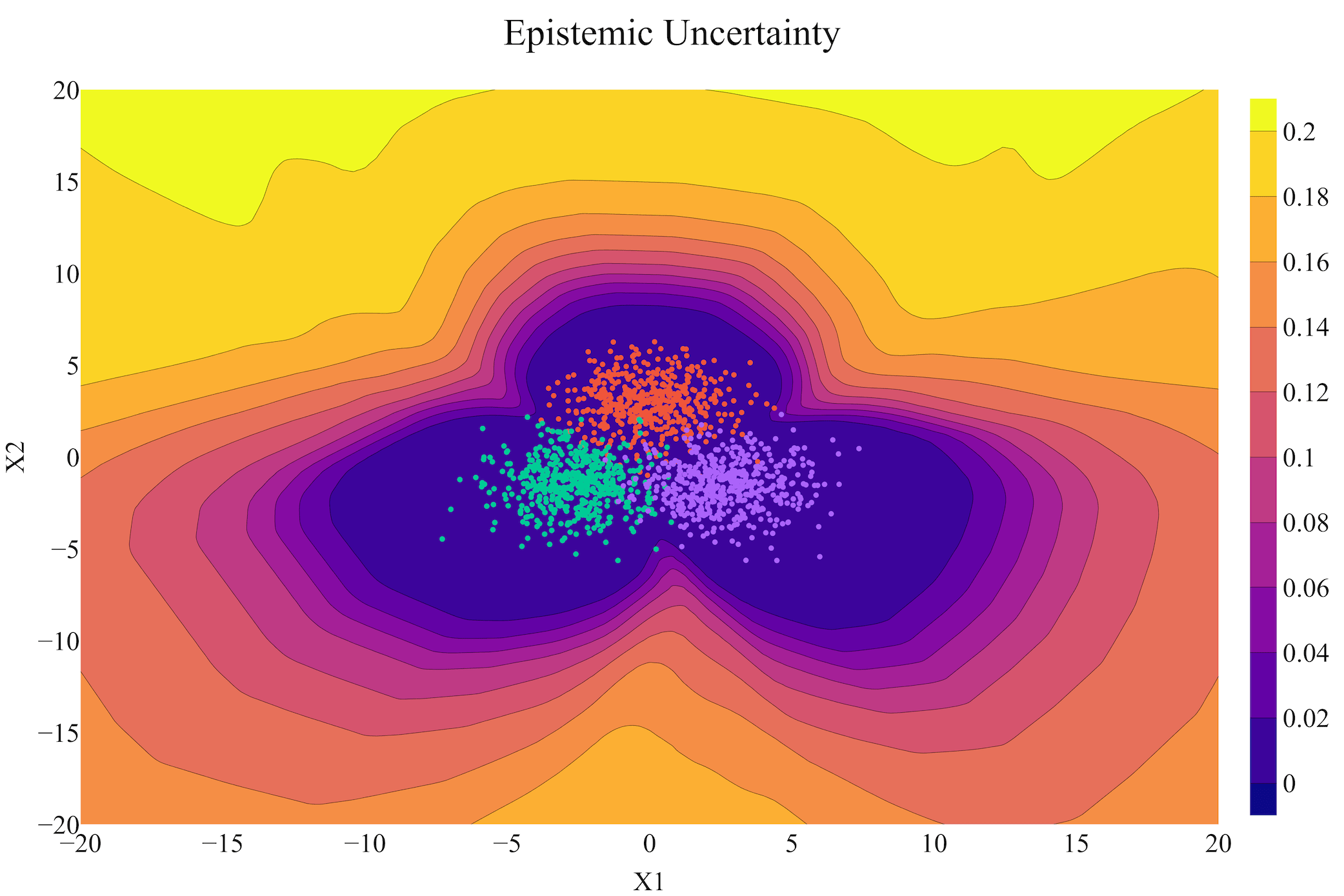}
    \caption{Epistemic Uncertainty}
\end{subfigure}
\caption{\textbf{BaCOUn Uncertainty Decomposition.} BaCOUn provides reliable OOD uncertainty while maintaining accurate aleatoric uncertainty in regions with high class overlap.}
\label{app_fig:gmm_bacoun}
\end{figure*}

\begin{figure*}[!h]
\centering
 \begin{subfigure}[t]{0.33\linewidth}
 \includegraphics[width=1.0\linewidth]{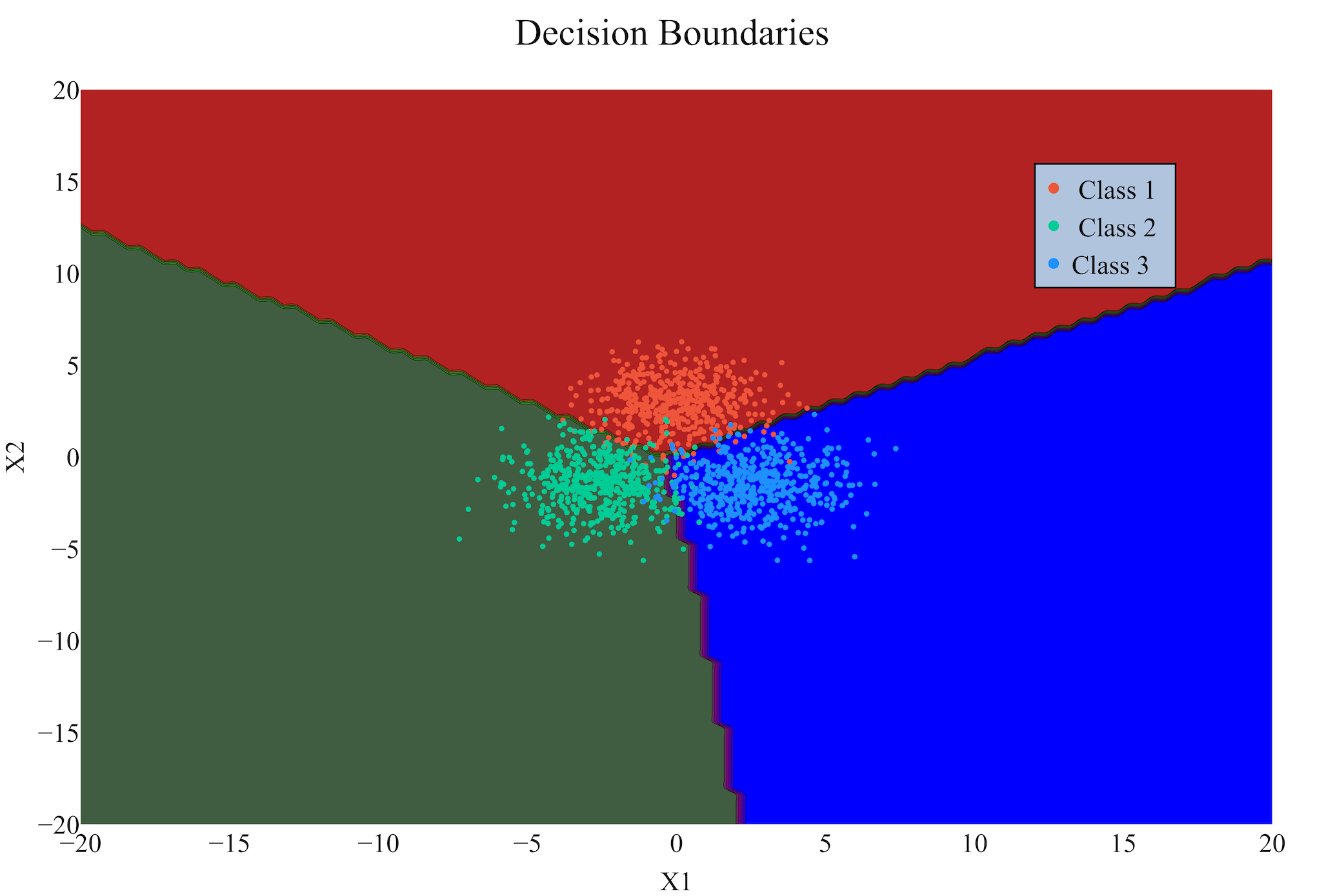}
 \caption{Decision Boundary}
\end{subfigure}
\begin{subfigure}[t]{0.33\linewidth}
    \includegraphics[width=1.0\linewidth]{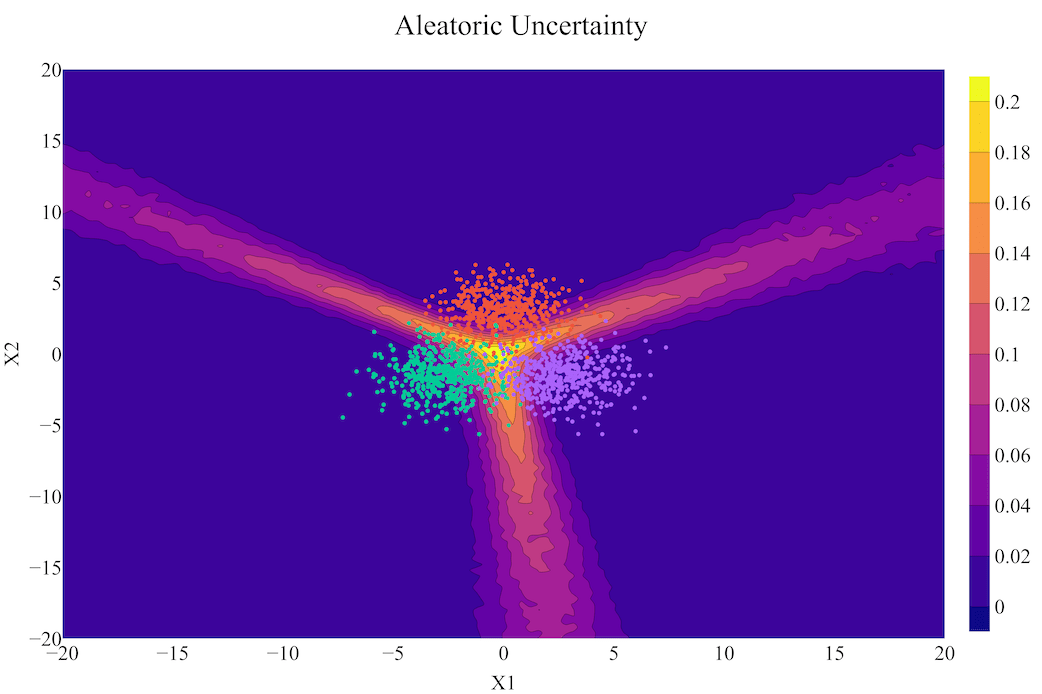}
    \caption{Aleatoric Uncertainty}
\end{subfigure}
\begin{subfigure}[t]{0.33\linewidth}
    \includegraphics[width=1.0\linewidth]{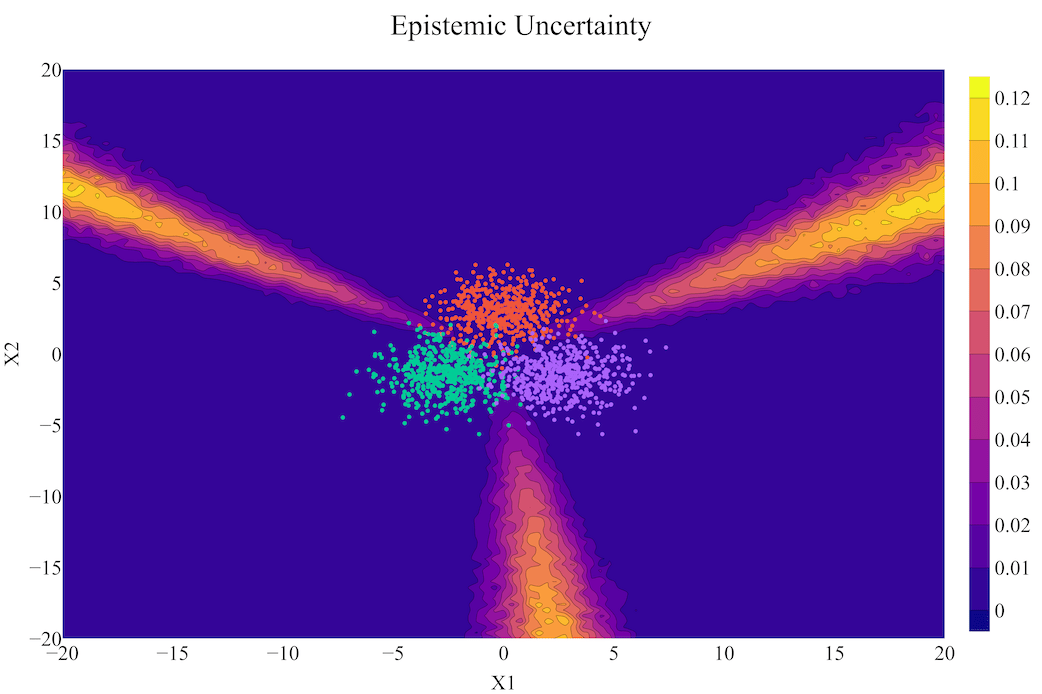}
    \caption{Epistemic Uncertainty}
\end{subfigure}
\caption{\textbf{MC-Dropout Uncertainty Decomposition.} The model overgeneralizes and produces overconfident predictions in regions of low-data density, failing to provide well-calibrated OOD uncertainty.}
\label{app_fig:gmm_mcd}
\end{figure*}

\begin{figure*}[!h]
\centering
\begin{subfigure}[t]{0.33\linewidth}
\includegraphics[width=1.0\linewidth]{figures_gmm/baselines/nlm/decision_boundary_copy.png}
\caption{Decision Boundary}
\end{subfigure}
\begin{subfigure}[t]{0.33\linewidth}
    \includegraphics[width=1.0\linewidth]{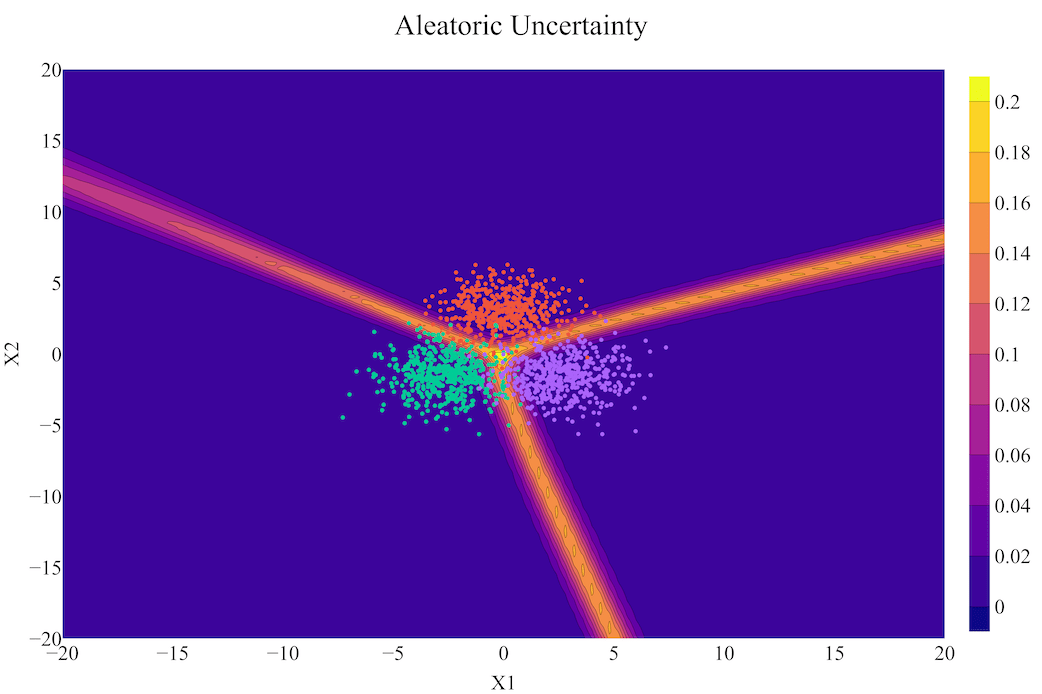}
    \caption{Aleatoric Uncertainty}
\end{subfigure}
\begin{subfigure}[t]{0.33\linewidth}
    \includegraphics[width=1.0\linewidth]{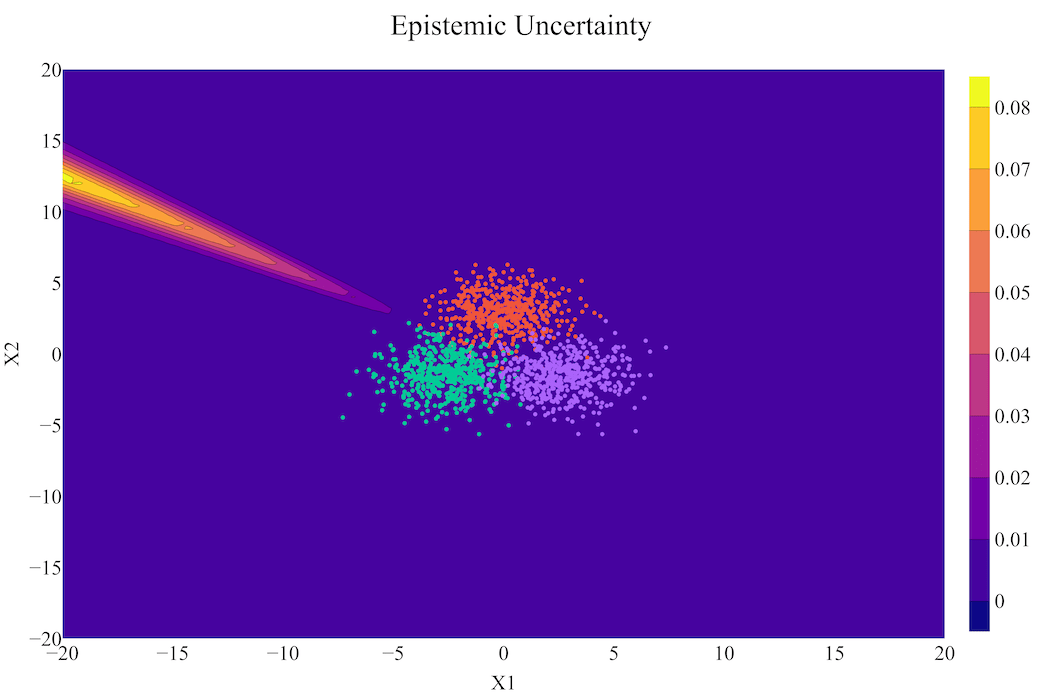}
    \caption{Epistemic Uncertainty}
\end{subfigure}
\caption{\textbf{NLM Uncertainty Decomposition.} Similarly to MC-Dropout, the model overgeneralizes and produces overconfident predictions in regions of low-data density, failing to provide well-calibrated OOD uncertainty.}
\label{app_fig:gmm_nlm}
\end{figure*}

\begin{figure*}[!h]
\centering
\begin{subfigure}[t]{0.33\linewidth}
\includegraphics[width=1.0\linewidth]{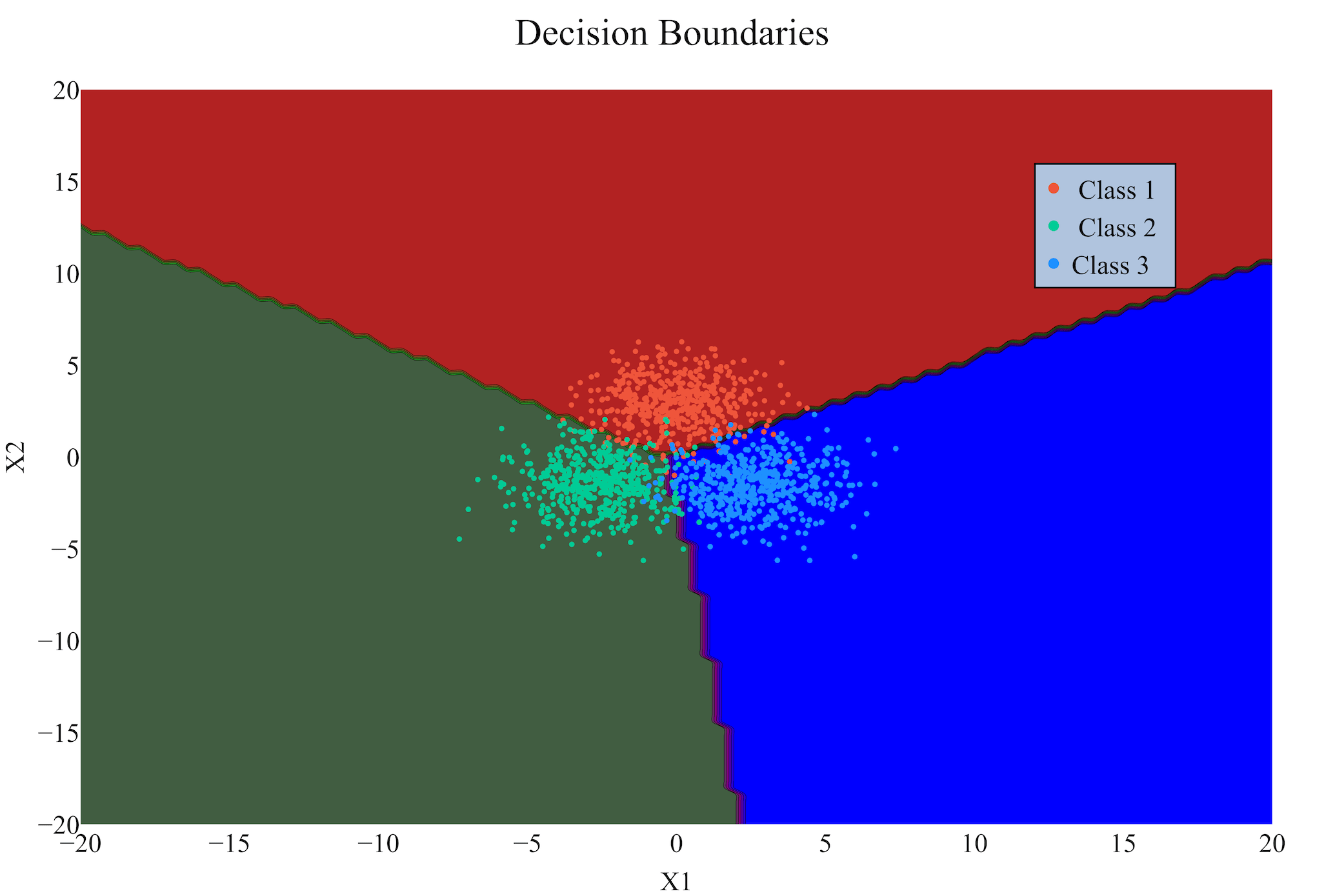}
\caption{Decision Boundary}
\end{subfigure}
\begin{subfigure}[t]{0.33\linewidth}
    \includegraphics[width=1.0\linewidth]{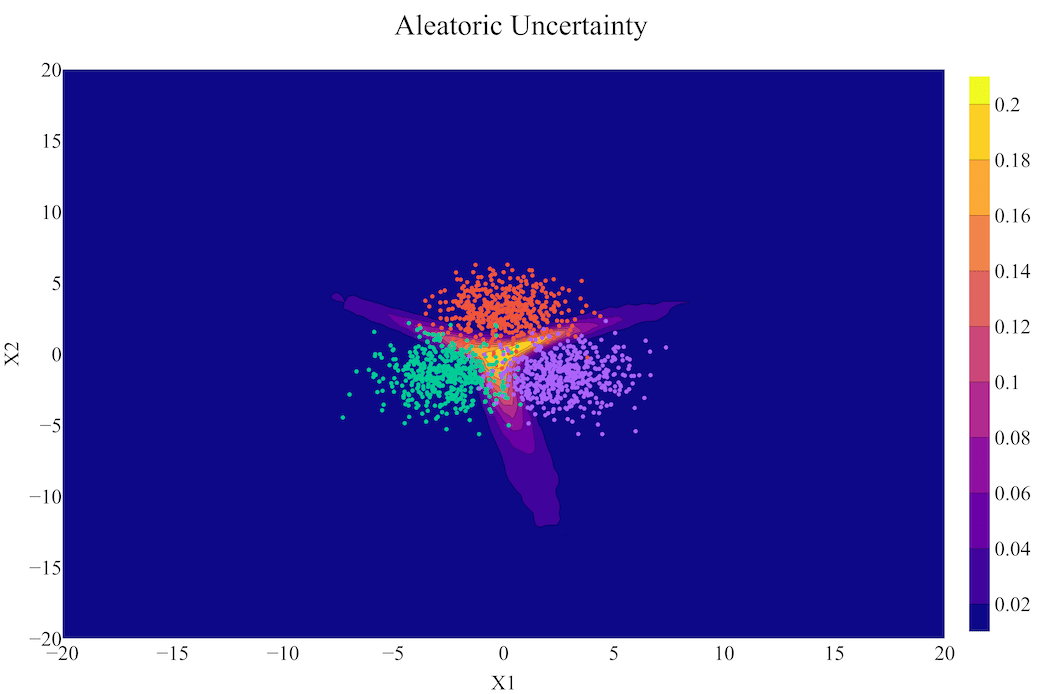}
    \caption{Aleatoric Uncertainty}
\end{subfigure}
\begin{subfigure}[t]{0.33\linewidth}
    \includegraphics[width=1.0\linewidth]{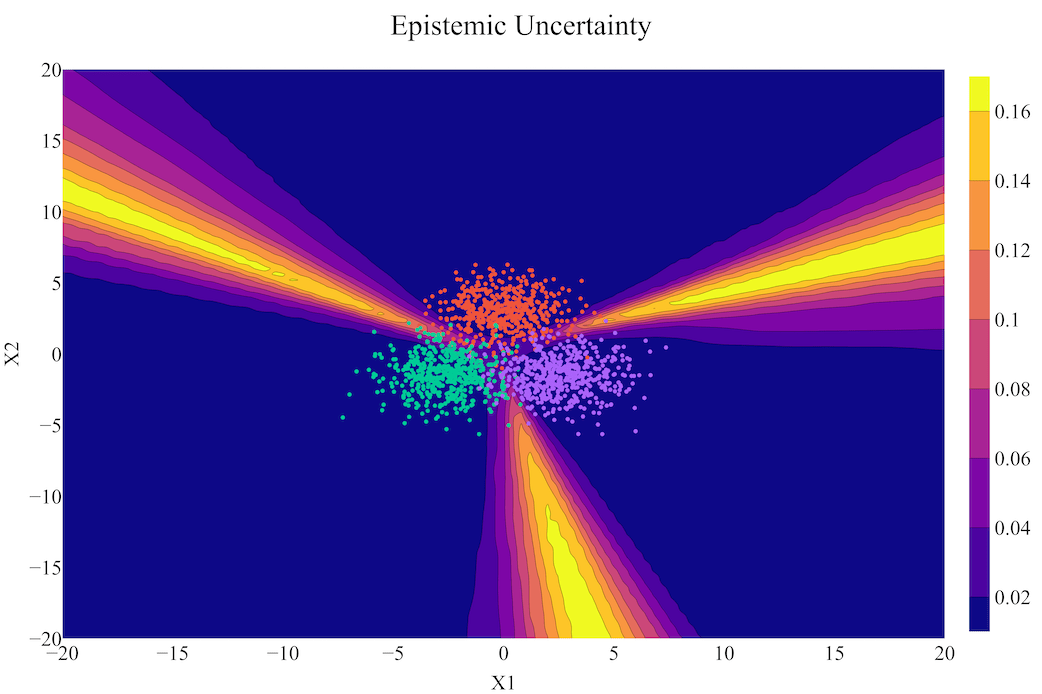}
    \caption{Epistemic Uncertainty}
\end{subfigure}
\caption{\textbf{BNN Uncertainty Decomposition.} A BNN provides the same pattern, as MC-Dropout and the NLM: the OOD uncertainty is underestimated in some directions.}
\label{app_fig:gmm_bnn}
\end{figure*}

Figure \ref{app_fig:total_unce_moons} shows that, on the moon-shaped clusters dataset (see Appendix \ref{appendix:data}),
BaCOUn learns GP-like uncertainty whereas MC-Dropout and NLM do not.
Figures \ref{app_fig:moons_bacoun}, \ref{app_fig:moons_mcd}, \ref{app_fig:moons_nlm} show the uncertainty decomposition of BaCOUn, MC-Dropout and NLM on the moons dataset, respectively.
BaCOUn is the only model that has increasing epistemic uncertainty as distance from the data increases; the remaining models make overconfident predictions nearly everywhere in the data-space. Moreover, MC-Dropout and NLM have high aleatoric uncertainty in data-poor regions where the classifier should only have high epistemic uncertainty. Similarly, we present the uncertainty decomposition obtained by the aforementioned methods on the GMM data (see Appendix \ref{appendix:data}) in Figures \ref{app_fig:total_unce_gmm}, \ref{app_fig:gmm_bacoun}, \ref{app_fig:gmm_mcd}, \ref{app_fig:gmm_nlm} and \ref{app_fig:gmm_bnn}.

\begin{figure*}[!h]
\centering
 \begin{subfigure}[t]{0.24\linewidth}
 \includegraphics[width=1.0\linewidth]{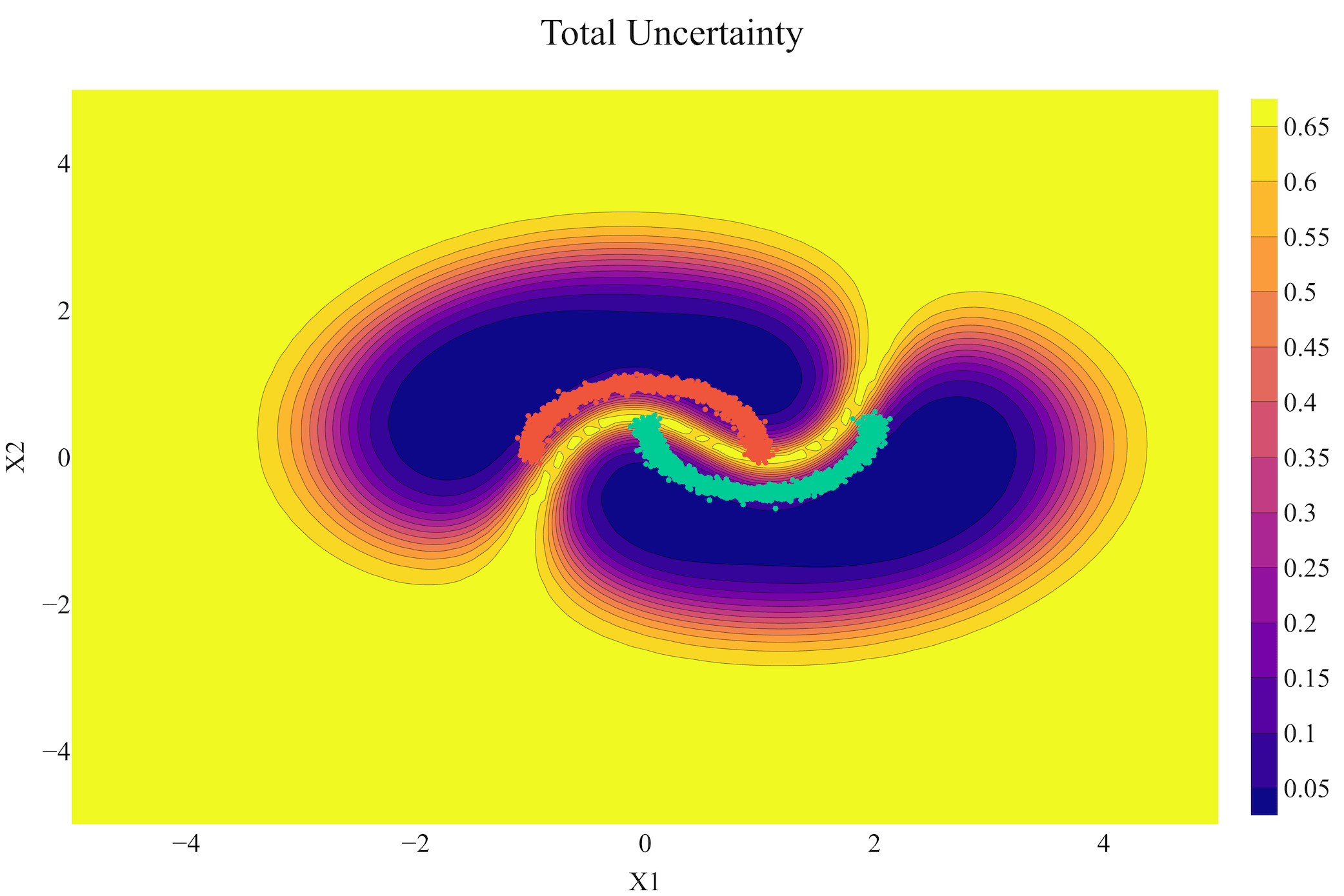}
 \caption{Gaussian Process}
 \end{subfigure}
\begin{subfigure}[t]{0.24\linewidth}
     \includegraphics[width=1.0\linewidth]{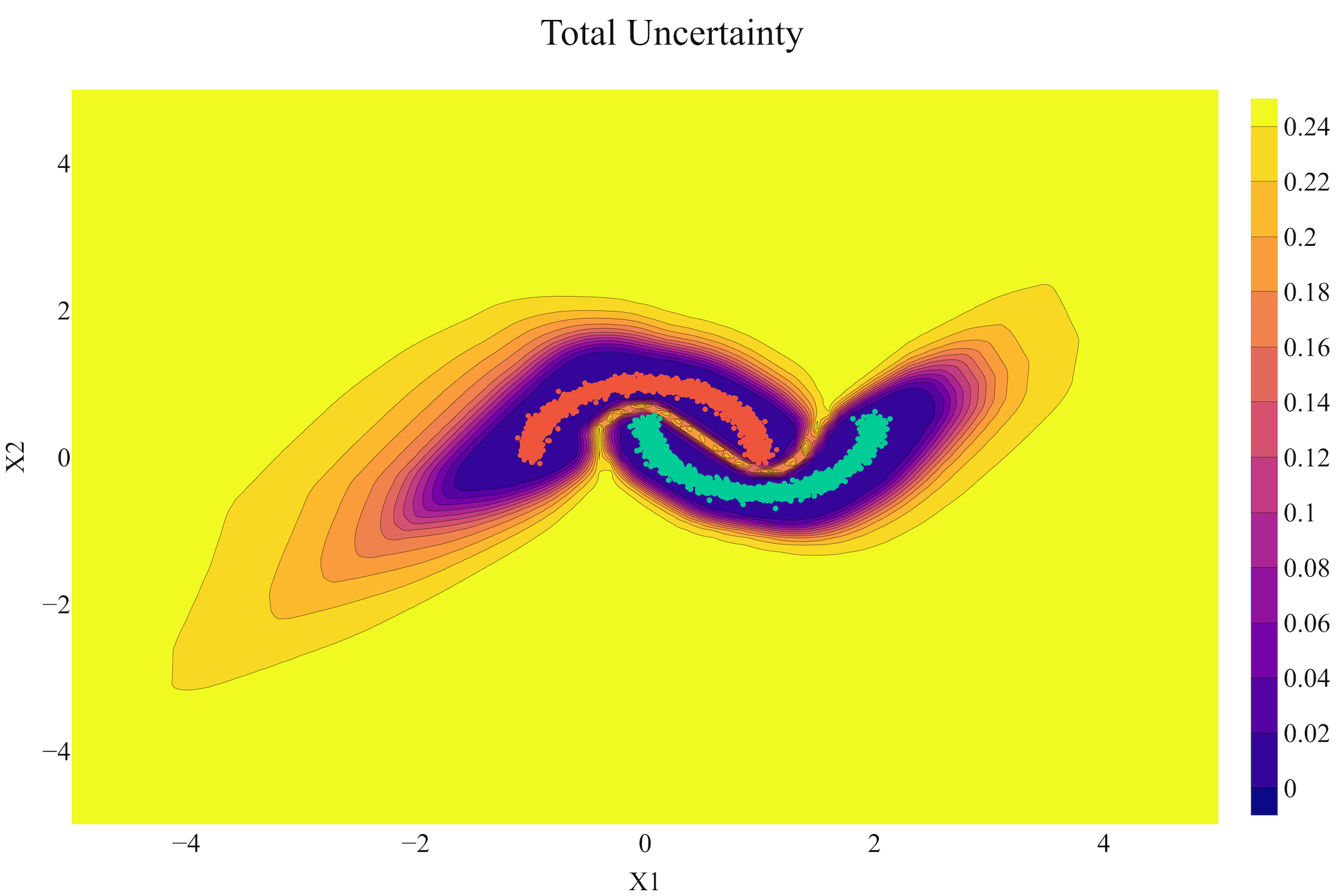}
      \caption{BaCOUn}
 \end{subfigure}
\begin{subfigure}[t]{0.24\linewidth}
    \includegraphics[width=1.0\linewidth]{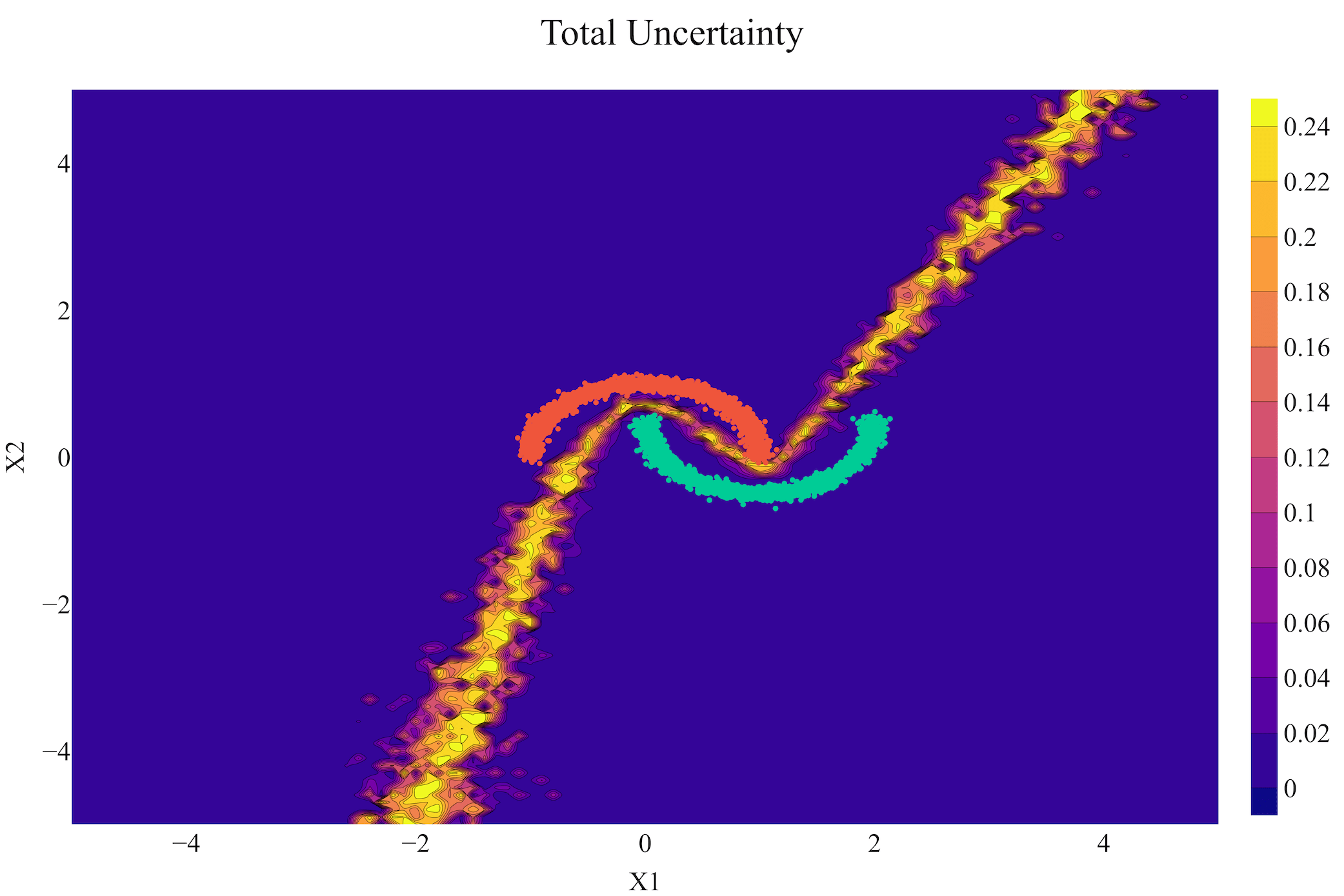}
    \caption{MC-Dropout}
\end{subfigure}
 \begin{subfigure}[t]{0.24\linewidth}
     \includegraphics[width=1.0\linewidth]{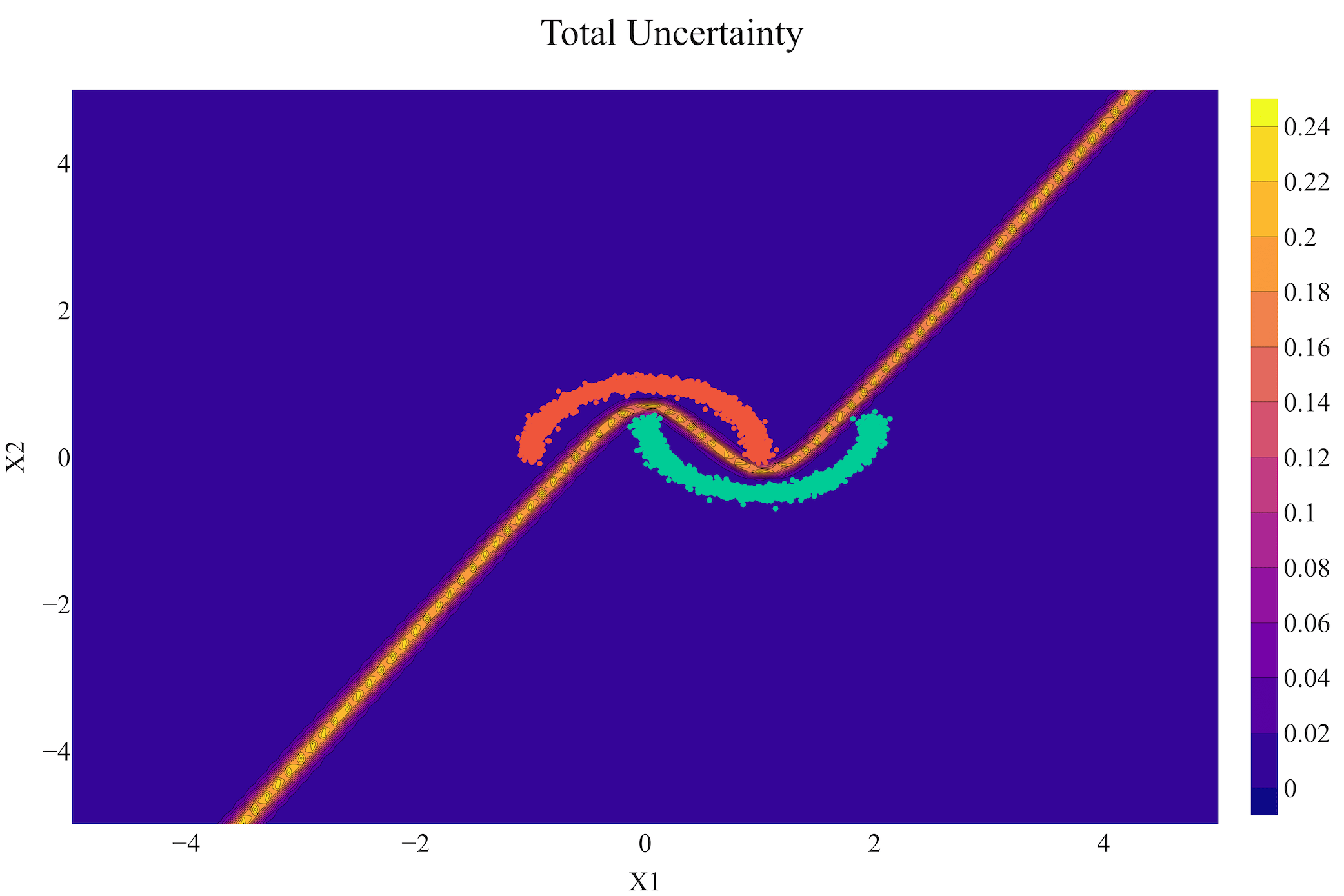}
     \caption{NLM}
\end{subfigure}
\caption{\textbf{Comparison of Total Uncertainty.} BaCOUn is the only model which obtains GP-like uncertainty, with high uncertainty far from the observed data as well as in regions of high class overlap.}
\label{app_fig:total_unce_moons}
\end{figure*}

\begin{figure*}[!h]
\centering
\begin{subfigure}[t]{0.33\linewidth}
\includegraphics[width=1.0\linewidth]{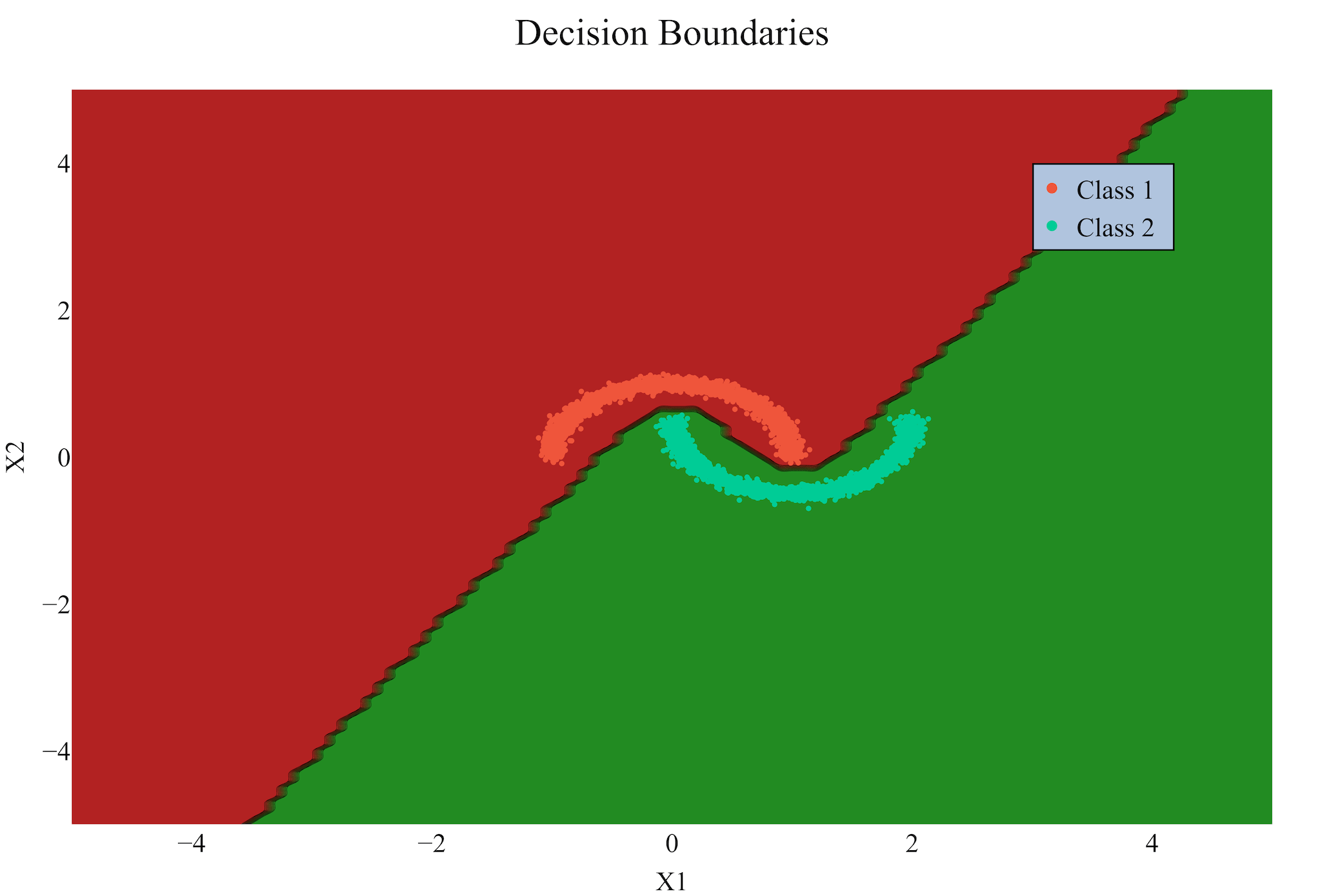}
\caption{Decision Boundary}
\end{subfigure}
\begin{subfigure}[t]{0.33\linewidth}
    \includegraphics[width=1.0\linewidth]{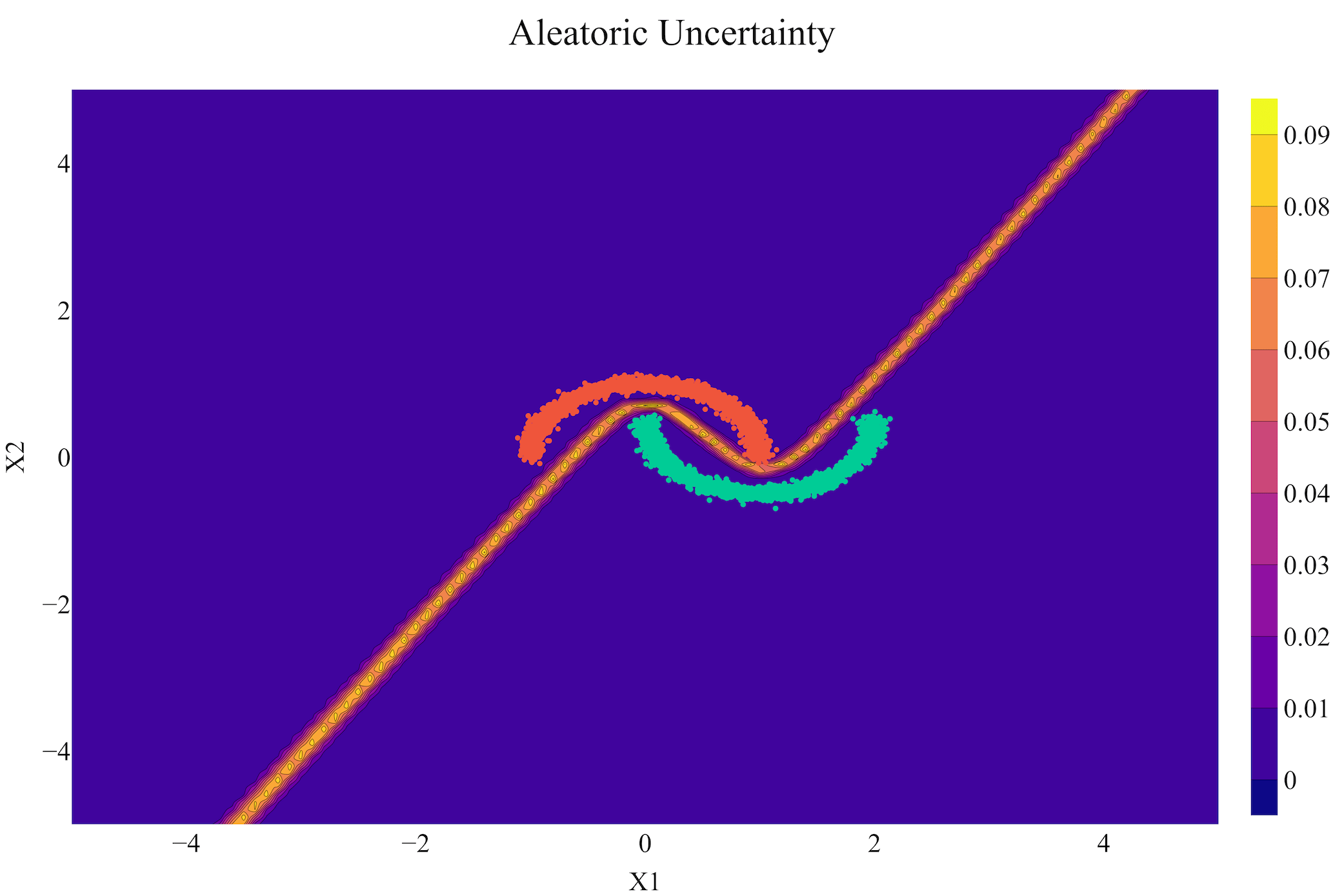}
    \caption{Aleatoric Uncertainty}
\end{subfigure}
\begin{subfigure}[t]{0.33\linewidth}
    \includegraphics[width=1.0\linewidth]{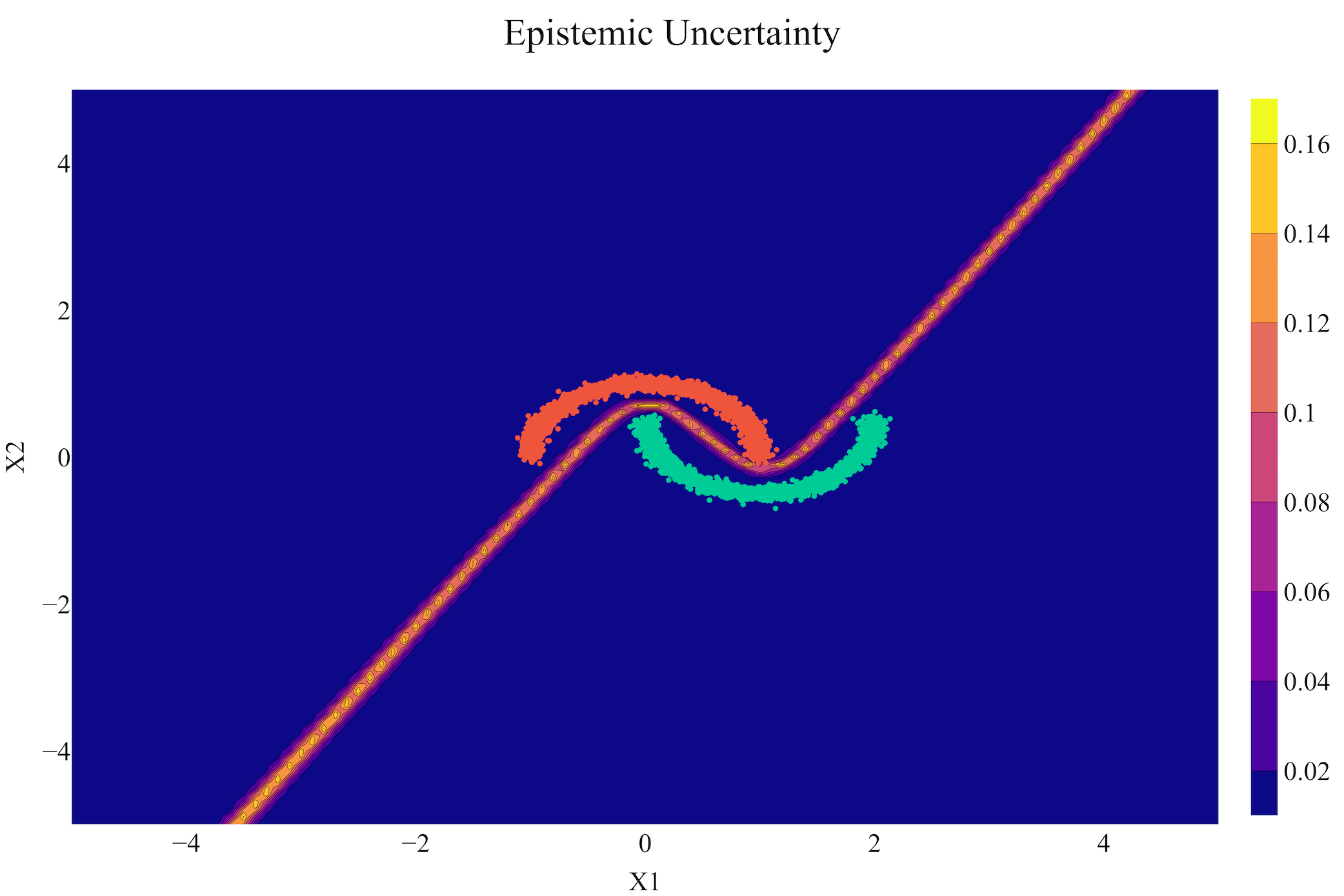}
    \caption{Epistemic Uncertainty}
\end{subfigure}
\caption{\textbf{BNN Uncertainty Decomposition.} A BNN provides the same pattern, as MC-Dropout and the NLM: the OOD uncertainty is underestimated in some directions.}
\label{app_fig:moons_bnn}
\end{figure*}

\begin{figure*}[!h]
\centering
 \begin{subfigure}[t]{0.24\linewidth}
 \includegraphics[width=1.0\linewidth]{figures_gmm/baselines/gp/entropy_copy.png}
 \caption{Gaussian Process}
 \end{subfigure}
\begin{subfigure}[t]{0.24\linewidth}
     \includegraphics[width=1.0\linewidth]{figures_gmm/bacoun/entropy_of_expected_copy.png}
      \caption{BaCOUn}
 \end{subfigure}
\begin{subfigure}[t]{0.24\linewidth}
    \includegraphics[width=1.0\linewidth]{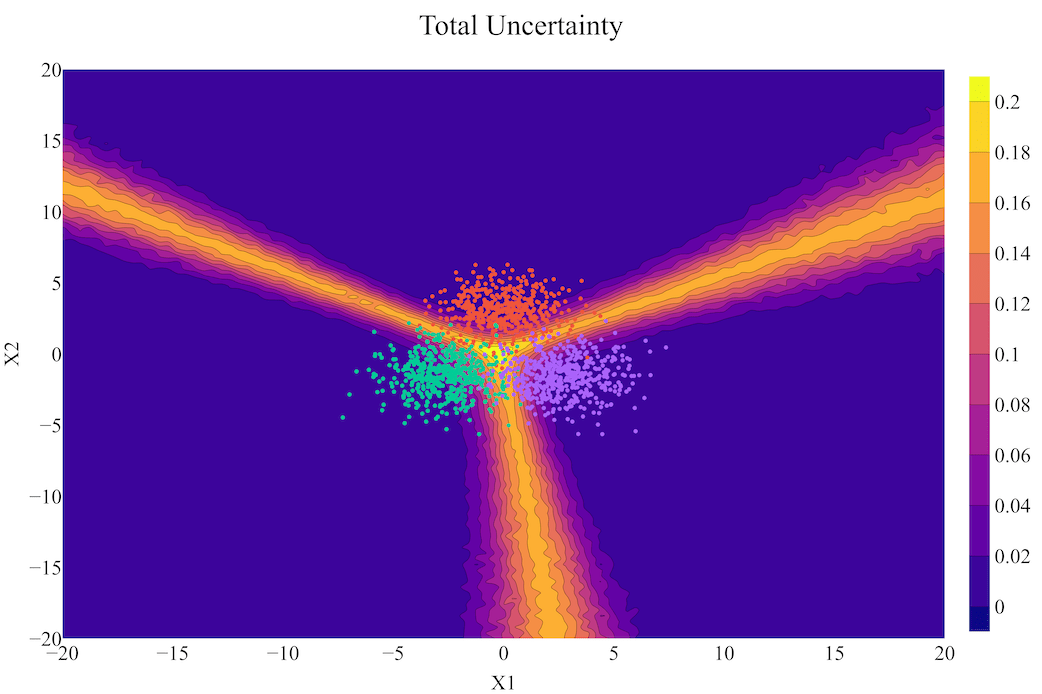}
    \caption{MC-Dropout}
\end{subfigure}
 \begin{subfigure}[t]{0.24\linewidth}
     \includegraphics[width=1.0\linewidth]{figures_gmm/baselines/nlm/total_uncertainty_copy.png}
     \caption{NLM}
\end{subfigure}
\caption{\textbf{Comparison of Total Uncertainty.} BaCOUn is the only model which obtains GP-like uncertainty, with high uncertainty far from the observed data as well as in regions of high class overlap.}
\label{app_fig:total_unce_gmm}
\end{figure*}

\subsection{BaCOUn provides interpretable uncertainty}

\paragraph{Individual Examples for Models Trained on MNIST}

\begin{figure*}[h]
\centering
\includegraphics[width=1.0\linewidth]{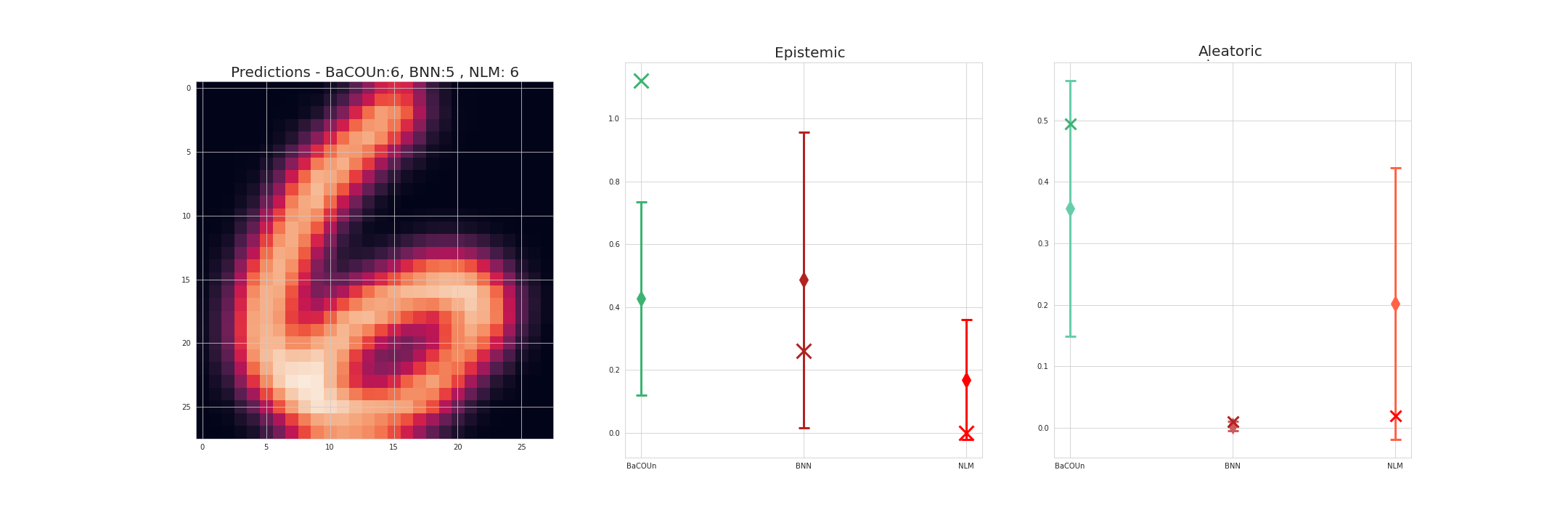}
\caption{\textbf{Uncertainty decomposition on MNIST example not in MNIST training set.} Given an image not in the training-set,
the BNN makes an overconfident and wrong prediction, the NLM correctly predicts the digit but does so with low uncertainty, whereas BaCOUn provides higher epistemic uncertainty.}
\label{fig:mnist_ood_6}
\end{figure*}

\begin{figure*}[h]
\centering
\includegraphics[width=1.0\linewidth]{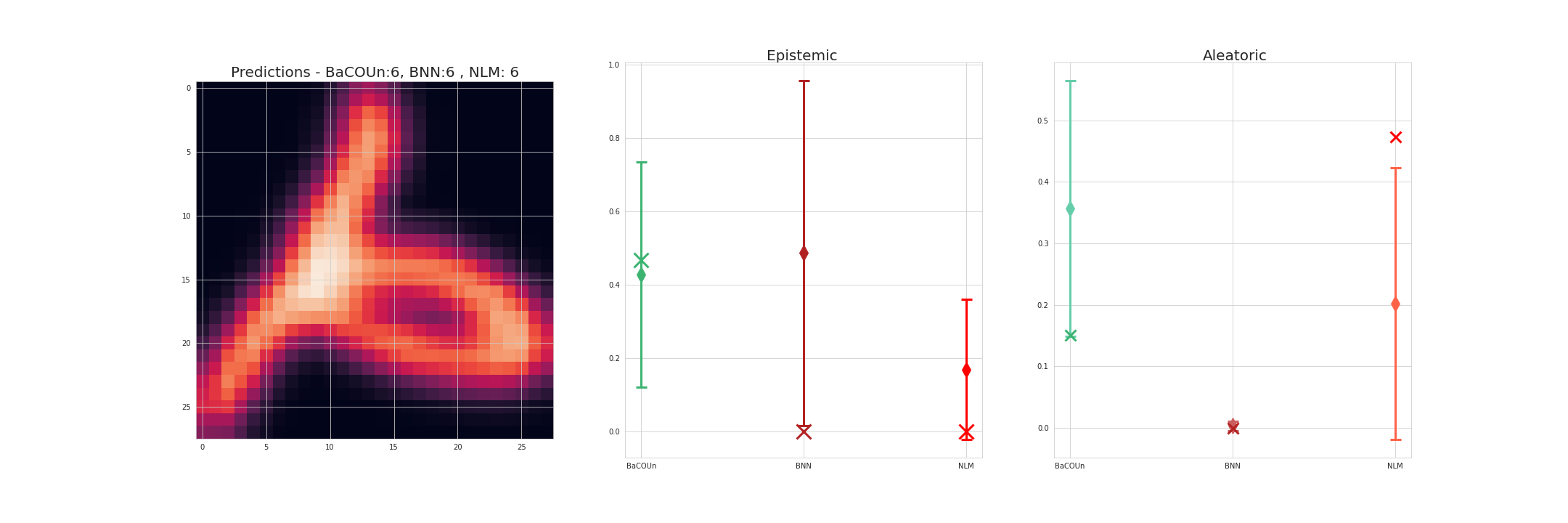}
\caption{\textbf{Uncertainty decomposition given USPS example for model trained on MNIST.} Consider a digit that, according to a human might or might not be a ``6'' (i.e. a human would have high uncertainty).
On this example, BaCOUn is more uncertain (like a human) whereas baseline methods are overconfident.}
\label{fig:mnist_usps1}
\end{figure*}

\begin{figure*}[h]
\centering
\includegraphics[width=1.0\linewidth]{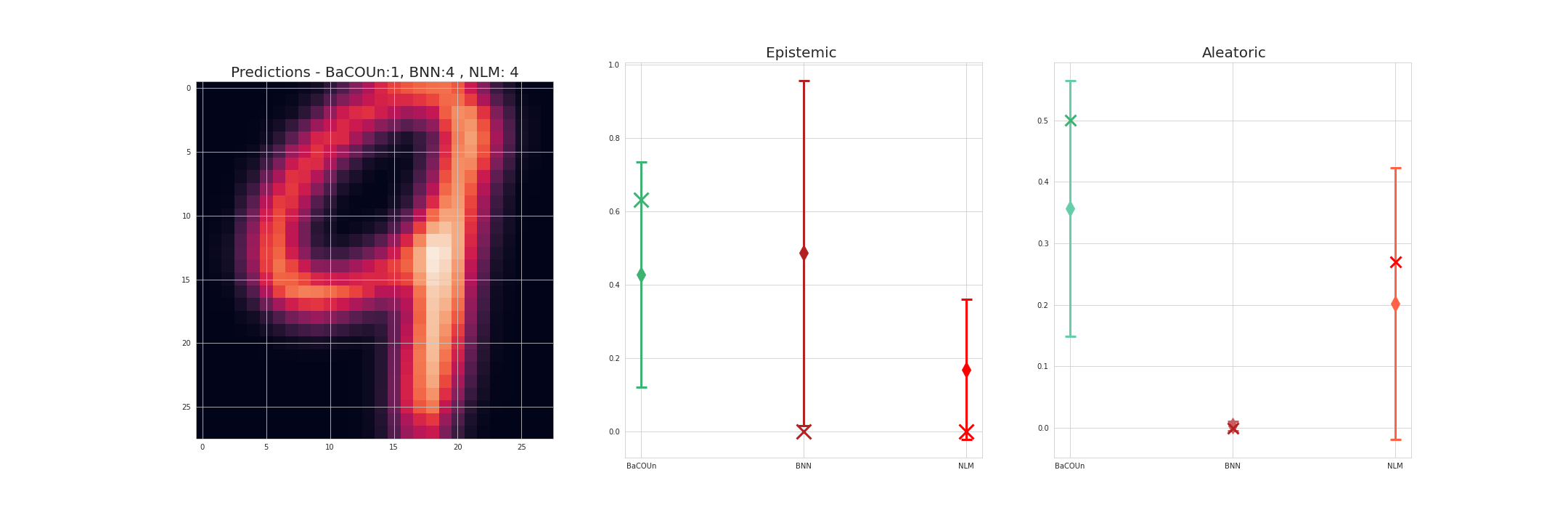}
\caption{\textbf{Uncertainty decomposition given USPS example for model trained on MNIST.} We present an example from USPS on which all methods predict incorrectly. BaCOUn provides high epistemic uncertainty while BNN and NLM are overconfident in their wrong prediction. }
\label{fig:mnist_usps2}
\end{figure*}

\begin{figure*}[h]
\centering
\begin{subfigure}{1.0\linewidth}
\includegraphics[width=1.0\linewidth]{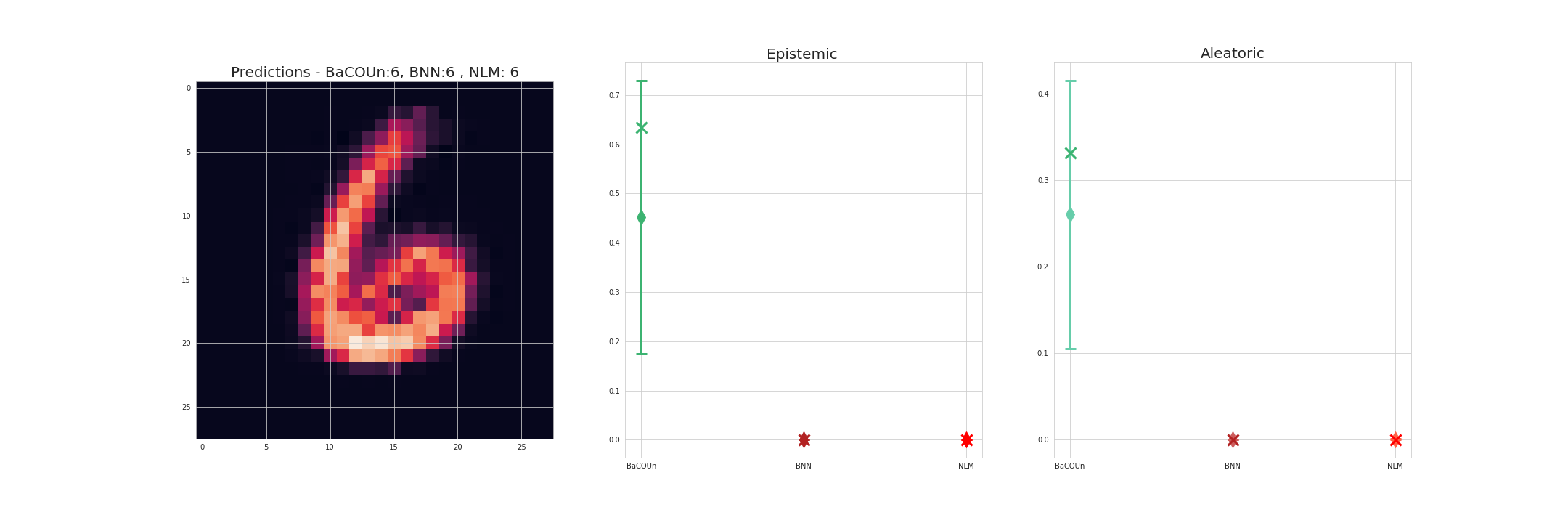}
\end{subfigure}
\begin{subfigure}{1.0\linewidth}
    \includegraphics[width=1.0\linewidth]{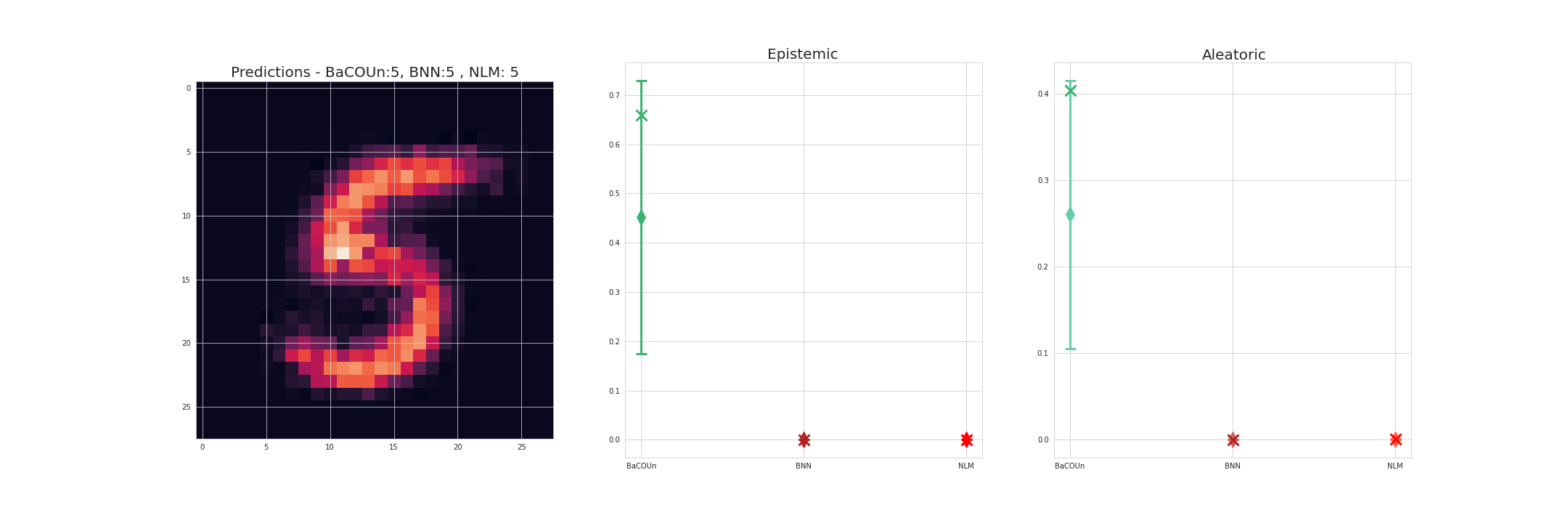}
\end{subfigure}
\caption{\textbf{Uncertainty decomposition on artificially generated OOD MNIST images.} We present the predictions of our methods on points chosen from the ``generated ood" dataset. We observe that the generated examples are very blurry images and thus should be considered OOD. NLM and BNN are excessively confident in their predictions, whereas BaCOUn is appropriately more uncertain.}
\label{fig:mnist_boundaries}
\end{figure*}

On a set of specific inputs, we demonstrate that in comparison to baselines, BaCOUn is able to obtain uncertainty decomposition that align with human intuition.
That is, BaCOUn gives higher epistemic uncertainty on OOD data points that are
\begin{itemize}
\item in the MNIST dataset but not in the training data (Figure \ref{fig:mnist_ood_6})
\item in the USPS dataset (Figure \ref{fig:mnist_usps1} and \ref{fig:mnist_usps2})
\item artificially generated to lie on the boundary of the MNIST data (Figure \ref{fig:mnist_boundaries})
\end{itemize}
In all of these case, baselines consistently make over-confident predictions (even when wrong), whereas BaCOUn remains conservatively uncertain. 
We note that in Figures \ref{fig:mnist_ood_6} and  \ref{fig:mnist_usps2}, baseline models make overconfident and \emph{incorrect} predictions.

\begin{figure*}[h]
\centering
\begin{subfigure}{0.33\linewidth}
\includegraphics[width=1.0\linewidth]{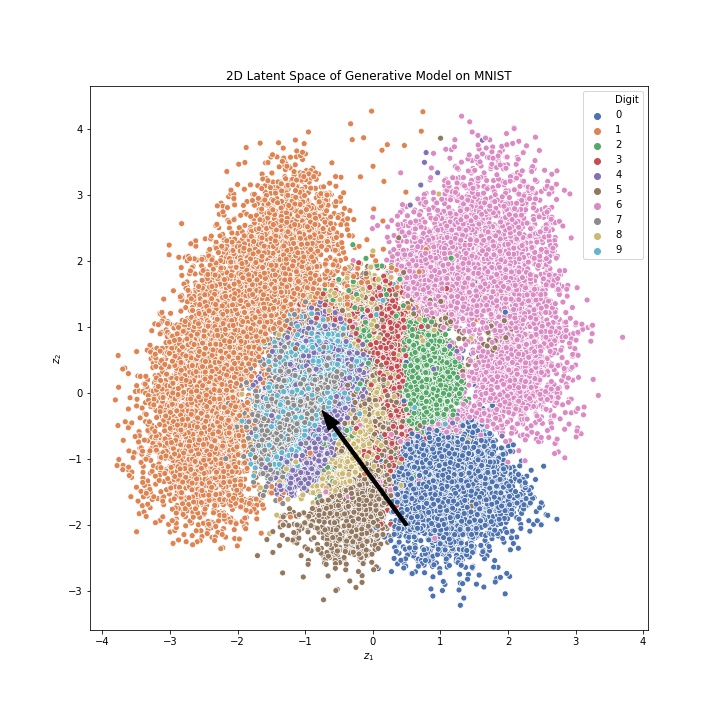}
\caption{Linear interpolation in 2D latent space of VAE trained on MNIST}
\end{subfigure}
\begin{subfigure}{0.66\linewidth}
\includegraphics[width=\linewidth]{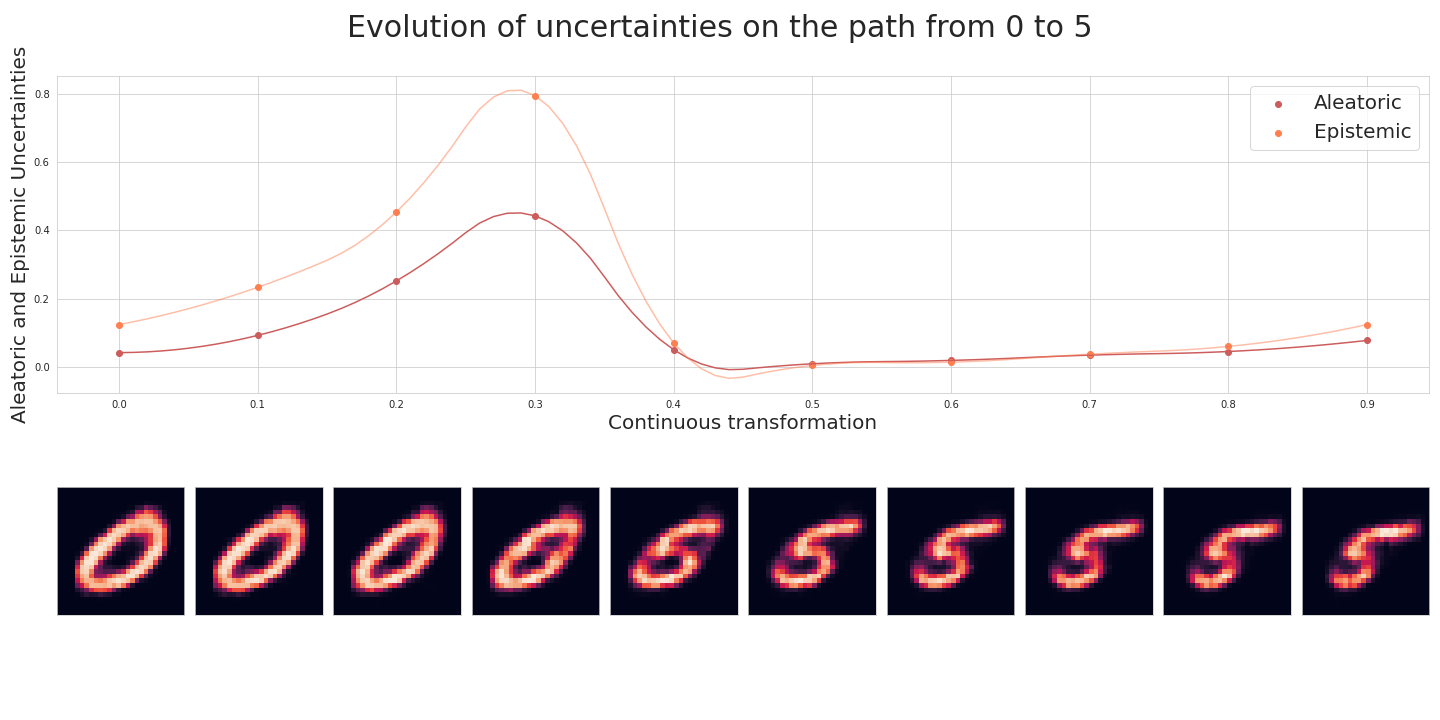}
\caption{Decomposition of uncertainty for digits along latent space interpolation.}
\end{subfigure}
\caption{\textbf{BaCOUn's decomposition of uncertainty across class boundaries: MNIST's 0 and 5 classes.} 
(a) We use a VAE to morph a 0 into a 5 by linearly interpolating in the VAE's latent space, (b) we present the uncertainty for the generated images.
At the beginning (and end) of the trajectory, we see that BaCOUn is confident that the digit is a 0 (and a 5).
In the middle, when the trajectory crosses BaCOUn's decision boundary, we see both high epistemic uncertainty and aleatoric uncertainty.
This is because the classes are not perfectly separable -- some digits look like both a 0 and a 5 -- leading to high aleatoric uncertainty, and because between the classes there's uncertainty over the exact boundary, leading to high epistemic uncertainty.
}
\label{app_fig:vae0to5}
\end{figure*}

\begin{figure*}[h]
\centering
\begin{subfigure}{0.33\linewidth}
\includegraphics[width=1.0\linewidth]{mnist/vae/0_to_5_1.jpg}
\caption{Linear interpolation in 2D latent space of VAE trained on MNIST}
\end{subfigure}
\begin{subfigure}{0.66\linewidth}
\includegraphics[width=\linewidth]{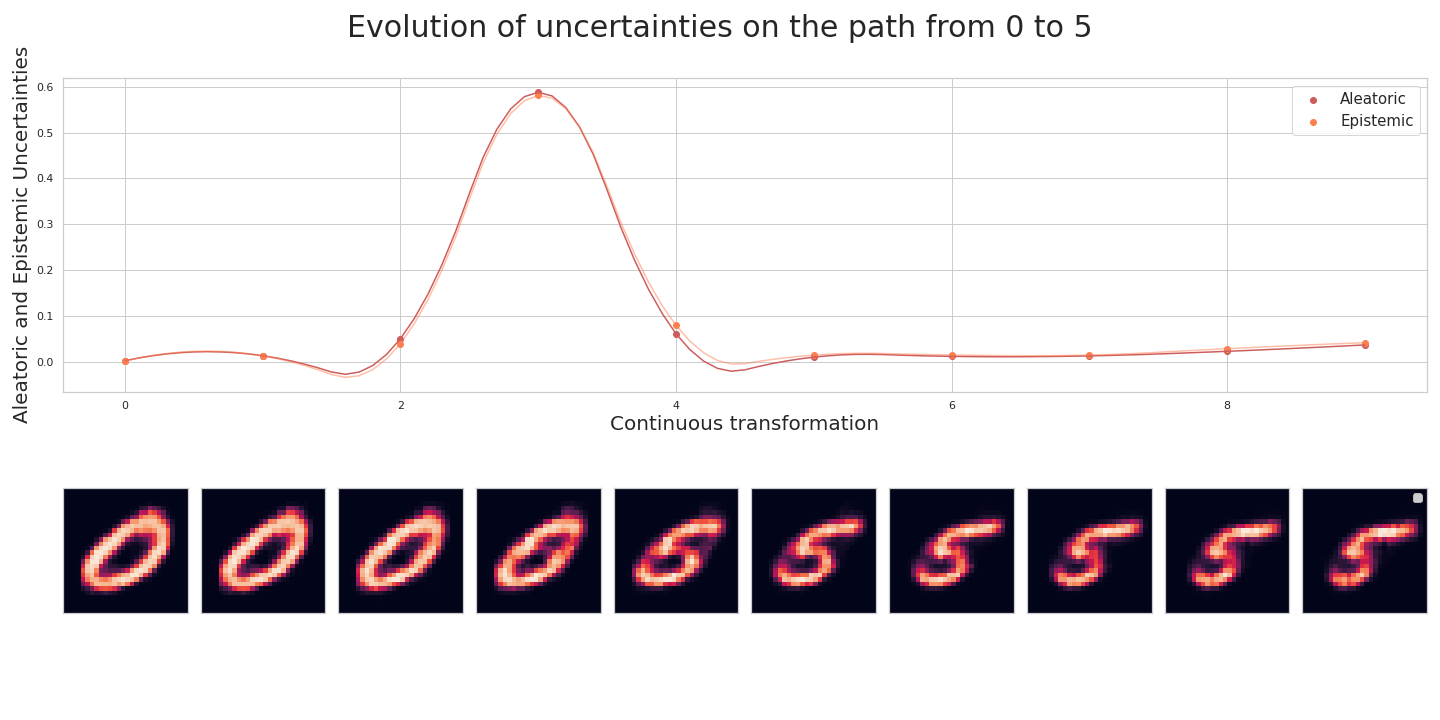}
\caption{Decomposition of uncertainty for digits along latent space interpolation.}
\end{subfigure}
\caption{\textbf{NLM's decomposition of uncertainty across class boundaries: MNIST's 0 and 5 classes.} 
We use a VAE to morph a 0 into a 5 by linearly interpolating in the VAE's latent space. We present the uncertainty for the generated images.
NLM is highly uncertainty on only one of the morphed digits, while it should be uncertain about several in the ``middle" of the trajectory, where the digits look neither like a 0 nor like a 5.
}
\label{app_fig:nlm_0to5}
\end{figure*}

\begin{figure*}[h]
\centering
\begin{subfigure}{0.33\linewidth}
\includegraphics[width=1.0\linewidth]{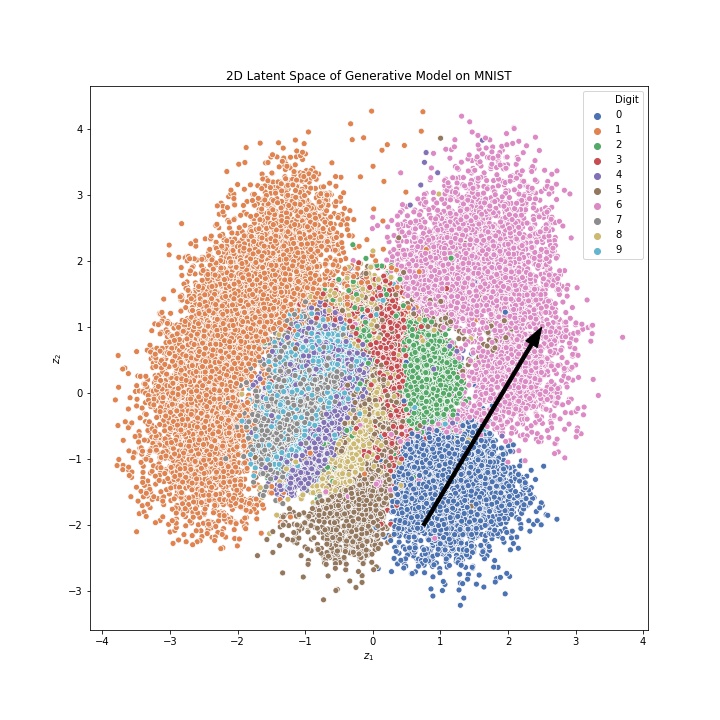}
\caption{Linear interpolation in 2D latent space of VAE trained on MNIST}
\end{subfigure}
\begin{subfigure}{0.66\linewidth}
\includegraphics[width=\linewidth]{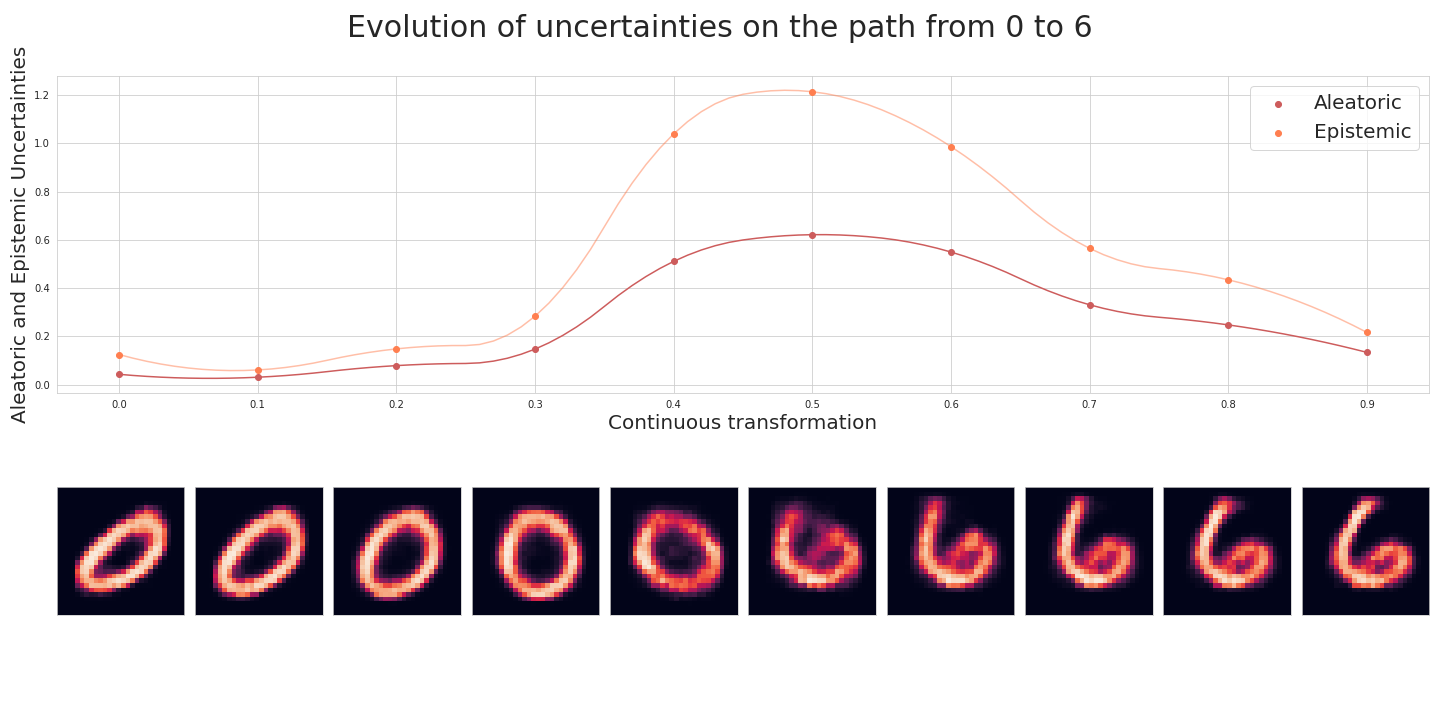}
\caption{Decomposition of uncertainty for digits along latent space interpolation.}
\end{subfigure}
\caption{\textbf{BaCOUn's decomposition of uncertainty across class boundaries: MNIST's 0 and 6 classes.} 
(a) We use a VAE to morph a 0 into a 6 by linearly interpolating in the VAE's latent space, (b) we present the uncertainty for the generated images.
At the beginning (and end) of the trajectory, we see that BaCOUn is confident that the digit is a 0 (and a 6).
In the middle, when the trajectory crosses BaCOUn's decision boundary, we see both high epistemic uncertainty and aleatoric uncertainty.
This is because the classes are not perfectly separable -- some digits look like both a 0 and a 6 -- leading to high aleatoric uncertainty, and because between the classes there's uncertainty over the exact boundary, leading to high epistemic uncertainty.
}
\label{app_fig:vae0to6}
\end{figure*}

\begin{figure*}[h]
\centering
\begin{subfigure}{0.33\linewidth}
\includegraphics[width=1.0\linewidth]{mnist/vae/0_to_6_1.jpg}
\caption{Linear interpolation in 2D latent space of VAE trained on MNIST}
\end{subfigure}
\begin{subfigure}{0.66\linewidth}
\includegraphics[width=\linewidth]{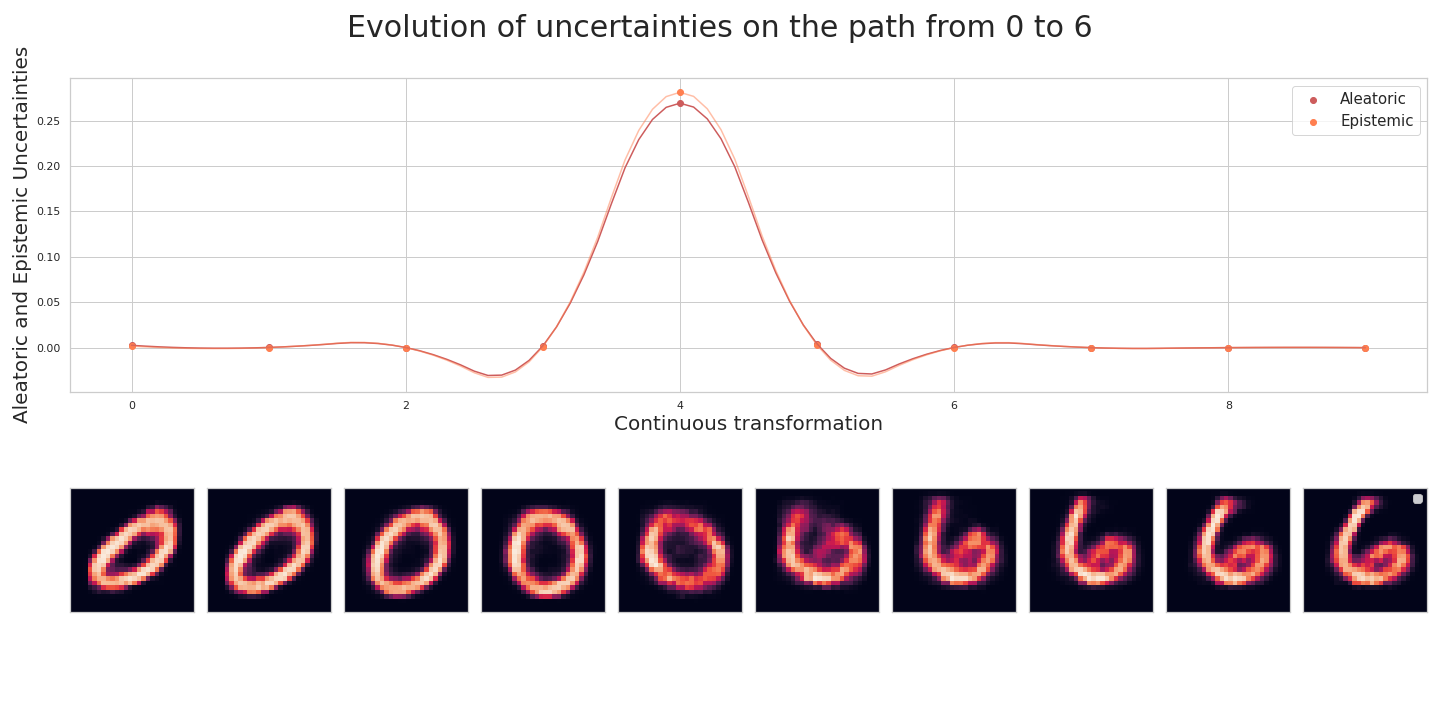}
\caption{Decomposition of uncertainty for digits along latent space interpolation.}
\end{subfigure}

\caption{\textbf{NLM's decomposition of uncertainty across class boundaries: MNIST's 0 and 6 classes.} 
We use a VAE to morph a 0 into a 6 by linearly interpolating in the VAE's latent space. We present the uncertainty for the generated images.
NLM is highly uncertainty on only one of the morphed digits, while it should be uncertain about several in the ``middle" of the trajectory, where the digits look neither like a 0 nor like a 6.
}
\label{app_fig:nlm_0to6}
\end{figure*}

\paragraph{Decomposition of uncertainty when crossing class boundaries}
We use a Variational Autoencoder (VAE) \cite{kingma2013auto}, pre-trained on MNIST~\cite{pretrained-vae}~\footnote{Pre-trained VAE from CreativeAI:
Deep Learning for
Computer Graphics. See \url{https://github.com/smartgeometry-ucl/dl4g}.} 
in order to to encode digits in a continuous 2D latent space. 
We then selected two points in the latent space, one in each of two neighboring classes, linearly interpolated between the two points, and decoded the points in the trajectory. 
Figures \ref{app_fig:vae0to5} and \ref{app_fig:vae0to6} show two latent-space trajectories, as well as the decoded digits and the uncertainty decomposition along these trajectories. 
At the beginning and end of each trajectory, BaCOUn is certain about the digits (as it should be).
In the middle of the trajectories, when the trajectory crosses BaCOUn's decision boundary, we see both high epistemic uncertainty and aleatoric uncertainty.
This is because the classes are not perfectly separable -- digits in the middle of the trajectory look like an uncanny mixture of the two digits -- leading to high aleatoric uncertainty, and because between the classes there's uncertainty over the exact boundary, leading to high epistemic uncertainty.
Figures \ref{app_fig:nlm_0to5} and \ref{app_fig:nlm_0to6} show the corresponding behavior of the NLM. 
They show that the NLM is over-confident about digits in the middle of the trajectory, which do not look realistic. 

\begin{figure*}[h]
\centering
\includegraphics[width=\linewidth]{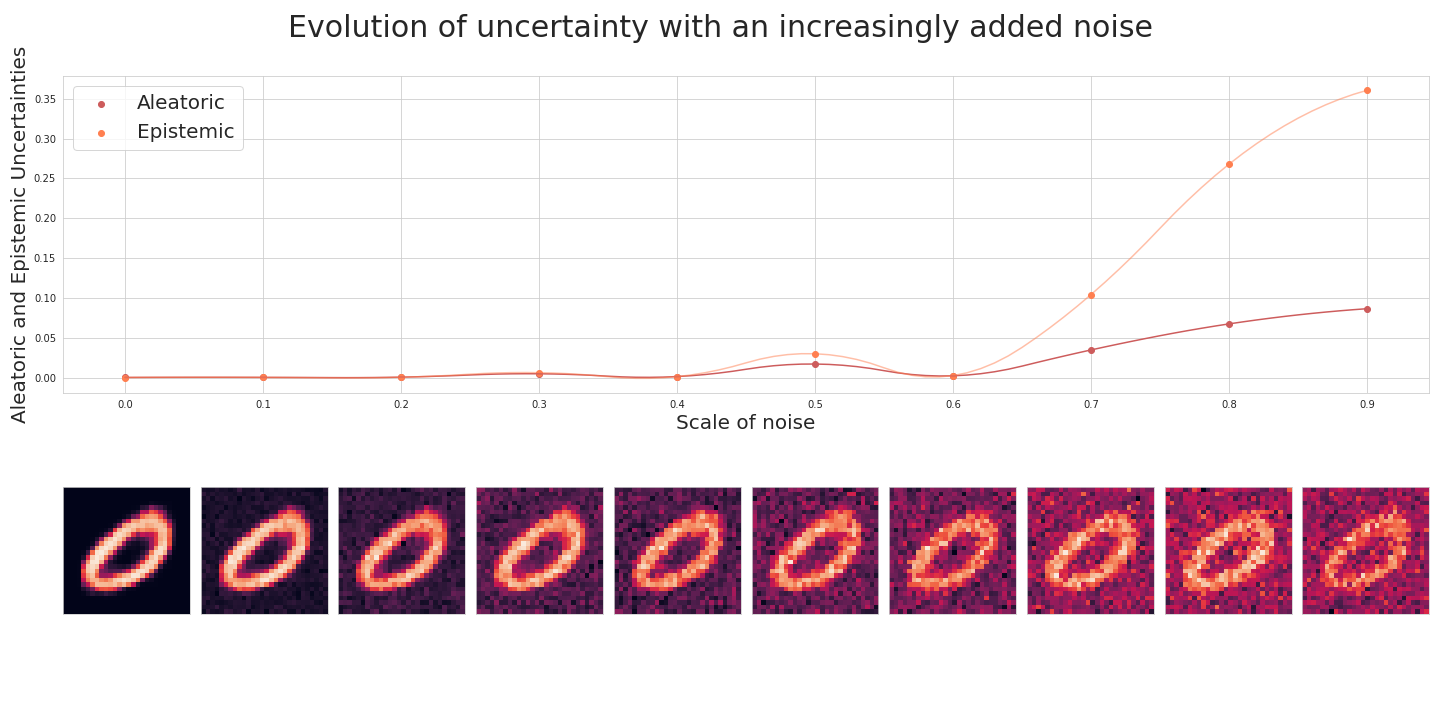}
\caption{\textbf{BaCOUn's decomposition of uncertainty when moving to data-poor regions.}
We take an MNIST image (left) and progressively add more Gaussian noise to it (right) in order to move it farther and farther away from the high-mass region of the data.
We see that as we move farther from the data, epistemic uncertainty increases significantly while aleatoric uncertainty does not.
}
\label{app_fig:blurry0}
\end{figure*}

\begin{figure*}[h]
\centering
\includegraphics[width=\linewidth]{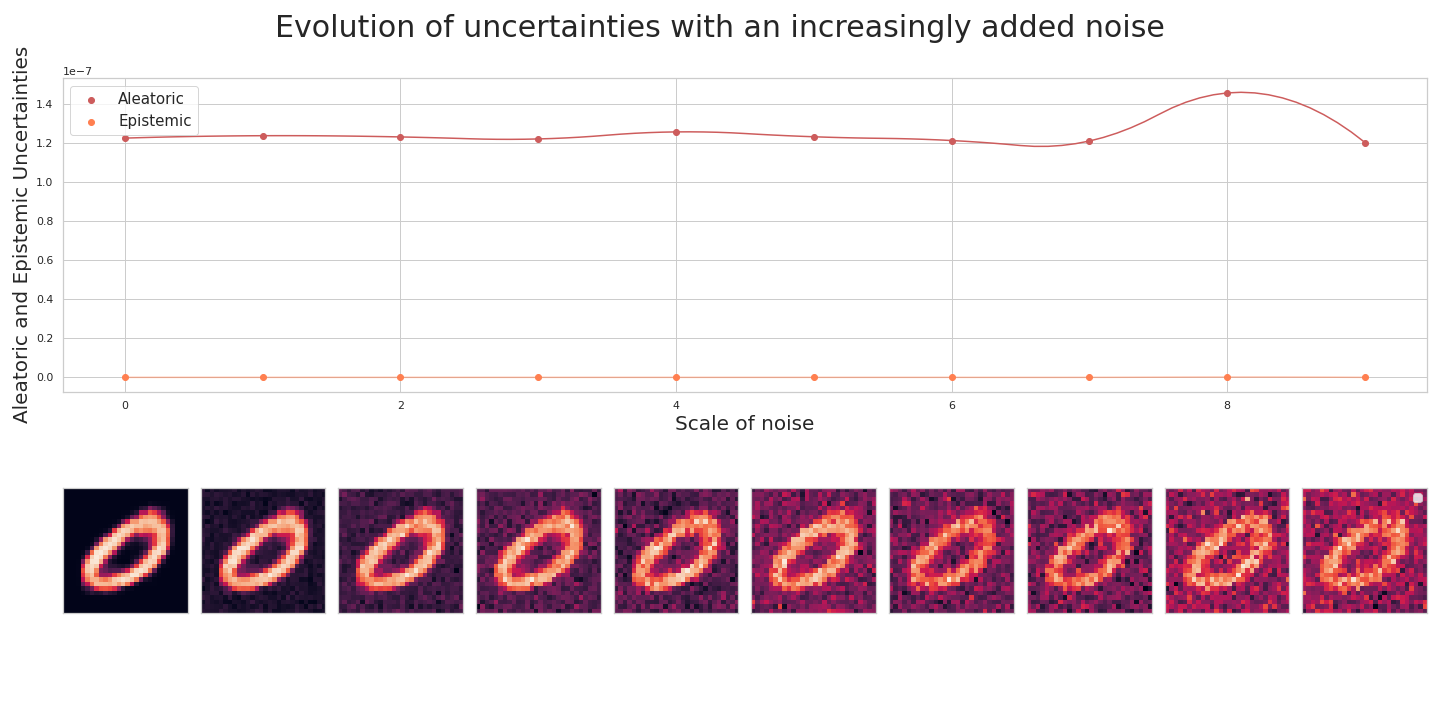}
\caption{\textbf{MLM's decomposition of uncertainty when moving to data-poor regions.}
We take an MNIST image (left) and progressively add more Gaussian noise to it (right) in order to move it farther and farther away from the high-mass region of the data.
We see that as we move farther from the data, epistemic uncertainty and aleatoric remains constant.
As such the learned classifier is unable to distinguish between data-rich and data-poor regions.
}
\label{app_fig:nlm_blurry0}
\end{figure*}

\paragraph{Decomposition of uncertainty when moving away from data-rich regions}
In Figure \ref{app_fig:blurry0}, we demonstrate how BaCOUn's aleatoric and epistemic uncertainty change when moving away from the data-rich regions of the space.
We do this by taking a digit and adding more and more Gaussian noise to it. 
We then examine the uncertainty decomposition as the digit gets more noisy, and observe that epistemic uncertainty increases drastically in comparison to aleatoric uncertainty.
Figure \ref{app_fig:nlm_blurry0} shows the behavior of the NLM model on the same noisy digits.
It shows that NLM is unable to distinguish between data-rich and data-poor regions.

\begin{figure*}[h]
\centering
\includegraphics[width=\linewidth]{mnist/rotations/final_rotations_9_dirty2.png}
\includegraphics[width=\linewidth]{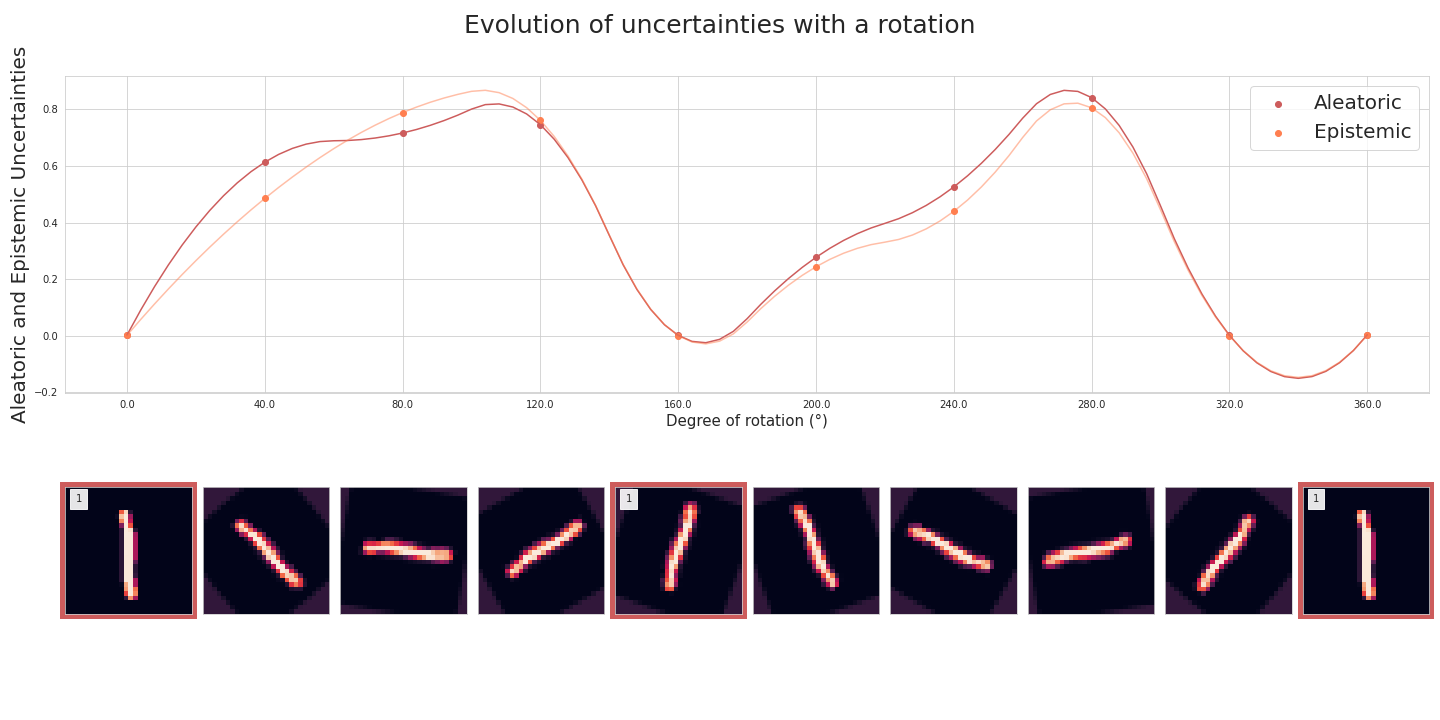}
\caption{\textbf{BaCOUn is certain about images that look like they belong in the training set.} We take MNIST digits and rotate them. We see that BaCOUn's uncertainty is low when the rotated digit resembles another digit. \textbf{Top:} a ``9'' can be flipped upside-down to look like a ``6''. \textbf{Bottom:} a ``1 '' can be flipped upside-down and will still look like a ``1''.}
\label{app_fig:rotation}
\end{figure*}

\paragraph{BaCOUn is more certain about digits that look like they should belong to the data}
In Figure \ref{app_fig:rotation}, we present different images from MNIST's test set along with the uncertainty obtained by BaCOUn as the image gets rotated by different degrees. We observe that uncertainties increase as the digit starts looking like an element that is not digit-like and decreases back when the image looks like a MNIST digit. These uncertainties reflect a human's intuition of uncertainties.

\paragraph{NLM fail to provide intuitive uncertainty decomposition} In Figures, \ref{app_fig:nlm_blurry0}, \ref{app_fig:nlm_0to5} and \ref{app_fig:nlm_0to6} we present the uncertainty metrics obtained by a NLM on the same examples as BaCOUn, namely, when we move away from the data-rich regions of the space, or following a latent trajectory between two classes. We observe that the NLM often provides overconfident predictions, even when the digits do not look like digits from the training data (Figures \ref{app_fig:nlm_0to5} and \ref{app_fig:nlm_0to6}) and fail to provide a useful uncertainty decomposition as we move away from the data manifold (Figure \ref{app_fig:nlm_blurry0}).

\FloatBarrier
\clearpage
\section{Robustness to Outliers}
\label{appendix:outliers}

When dealing with OOD data and uncertainty, the question of outliers arises and becomes an important one. Outliers are often considered as perturbations or anomalies that one should remove from a dataset as a pre-processing step. Even if the border between OOD points and outliers is porous, we present a simple and scalable approach to deal with outliers in a dataset. 


A recent line of work has focused on using Generative models for anomaly detection. For instance, \cite{schlegl2017unsupervised} present AnoGAN and use Generative Adversarial Networks for Marker Discovery, whereas \cite{ryzhikov2019normalizing} propose to use Normalizing Flows for Bayesian inference and deal with class imbalance. Our approach consists of using the density modelled by BaCOUn's Normalizing flow in order to rank and remove the data points which are most likely to be outliers. Despite its simplicity, our method provides reliable anomaly detection in relatively simple settings and is composed of the following steps:

\begin{itemize}
    \item Train the Normalizing flow on the data (including potential outliers).
    \item Compute the log probabilities of the data points under the Normalizing low. 
    \item Remove k data points with lowest probabilities. 
\end{itemize}

\begin{figure*}[h]
\centering
\begin{subfigure}[t]{0.33\linewidth}
\includegraphics[width=1.0\linewidth]{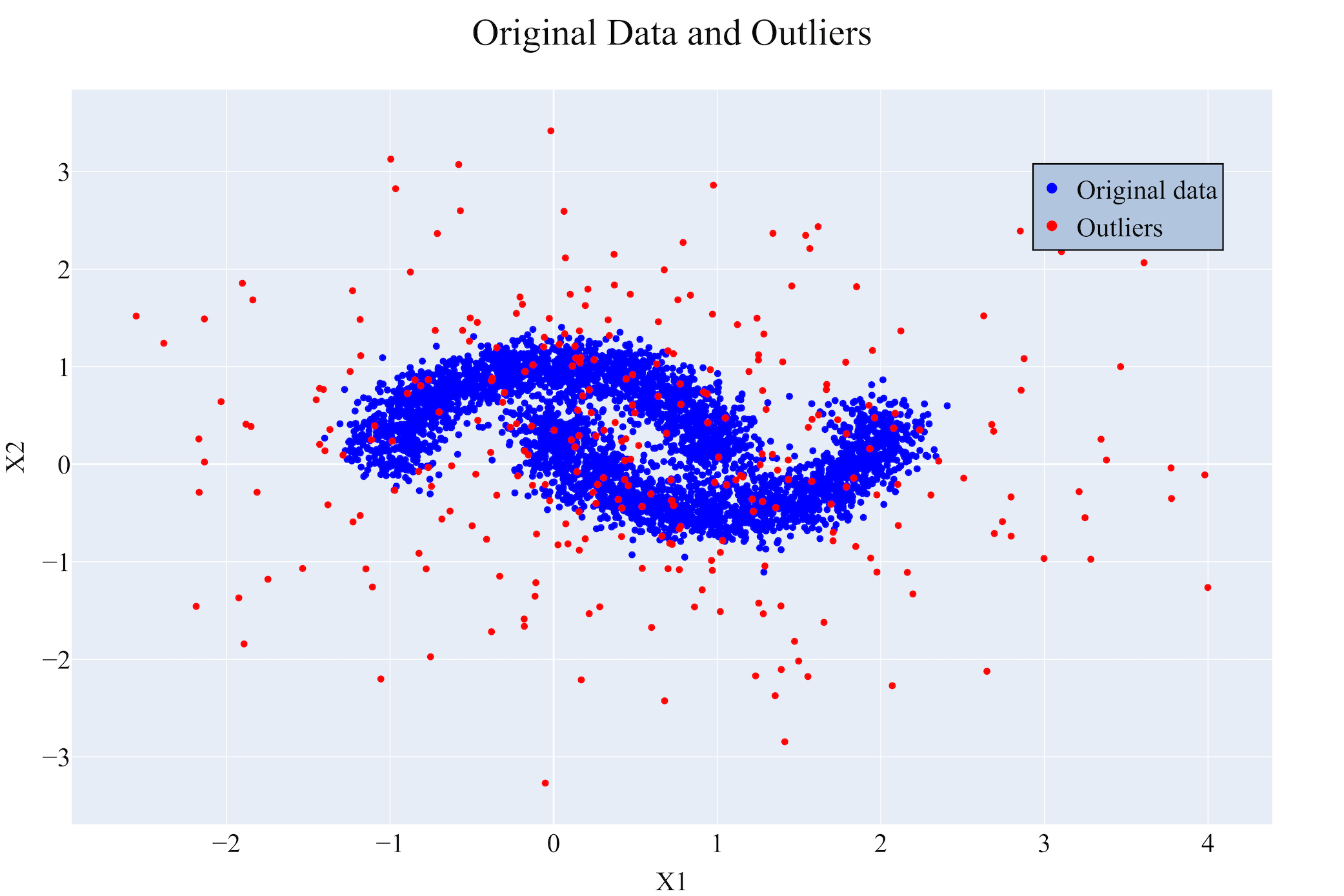}
\end{subfigure}%
\begin{subfigure}[t]{0.33\linewidth}
    \includegraphics[width=1.0\linewidth]{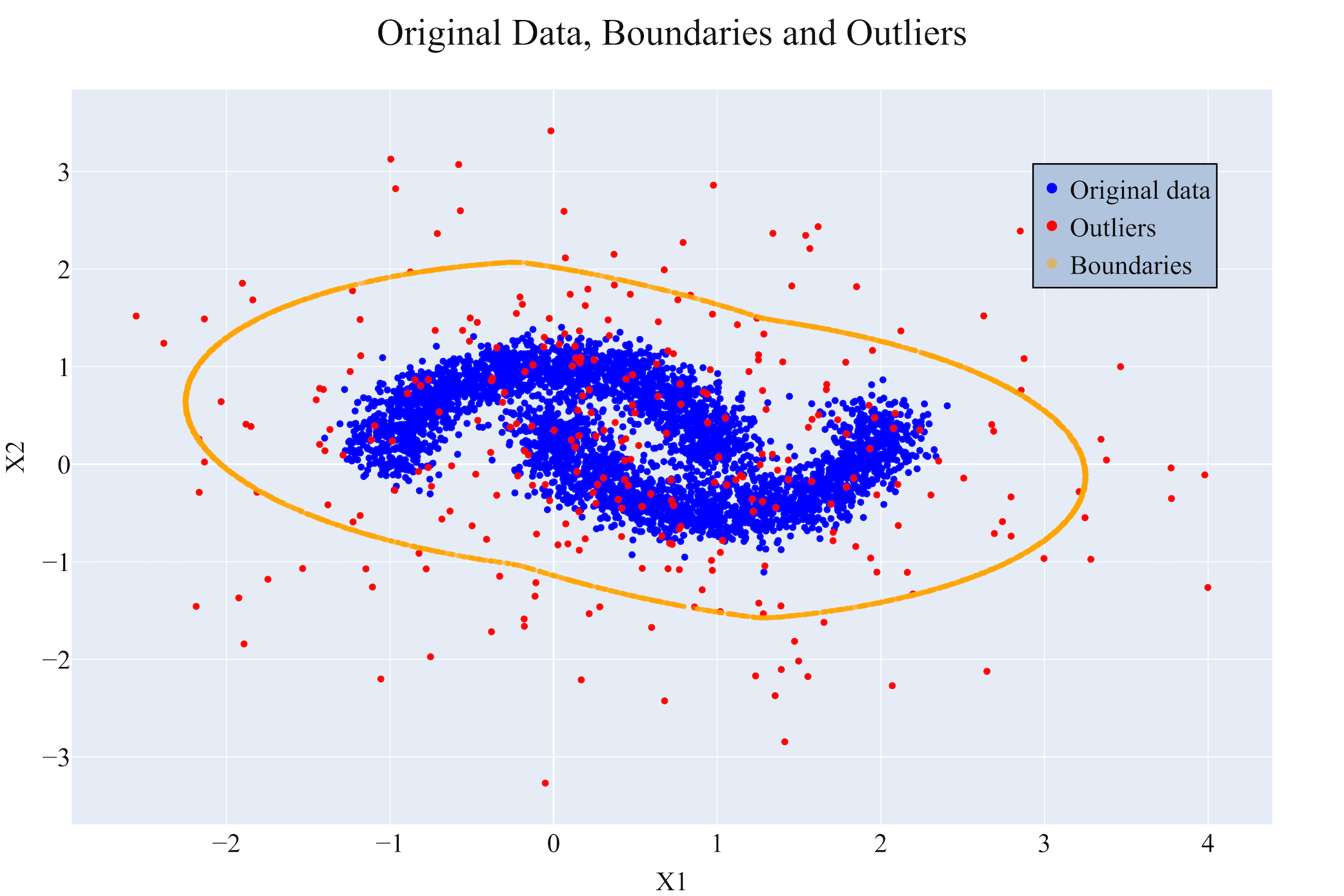}
\end{subfigure}
\begin{subfigure}[t]{0.33\linewidth}
    \includegraphics[width=1.0\linewidth]{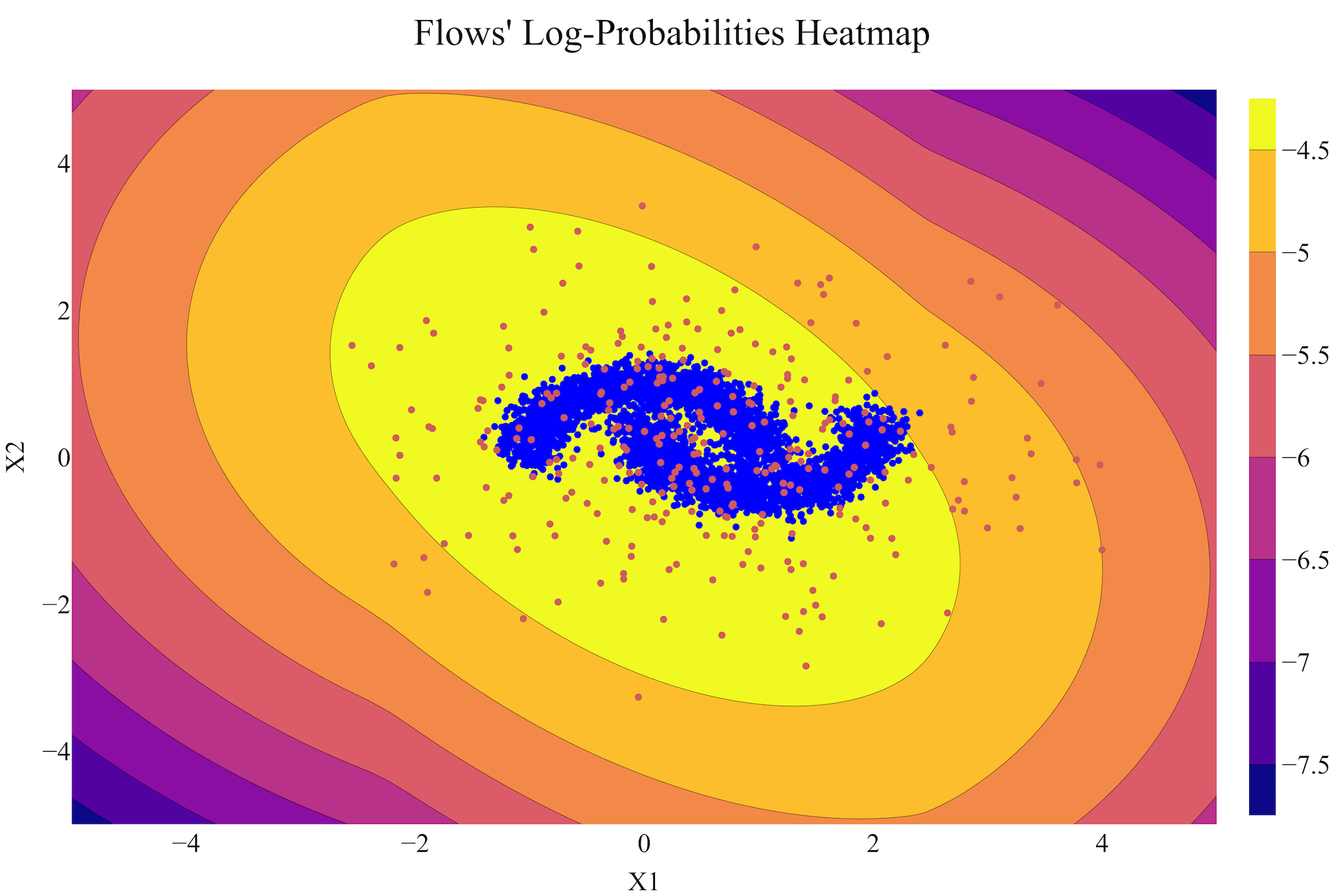}
\end{subfigure}
\caption{Data with outliers (a), Trained Normalizing flow and generated boundaries (b), (c) HeatMap of the Flow's log probabilities. Normalizing Flows are powerful tools to capture the global geometry of the problem without being overly sensitive to outliers, as shown in \textbf{(b)}. The boundary generated doesn't play any role in the selection procedure, but it shows how robust the density estimation is. Furthermore, the heatmap of the Normalizing Flow's probabilities in \textbf{(c)} aligns with our intuition: points far from the training data have lower probability under the flow.}
\label{fig:outliers_removal}
\end{figure*}

\begin{figure*}[h]
\centering
\begin{subfigure}[t]{0.49\linewidth}
\includegraphics[width=1.0\linewidth]{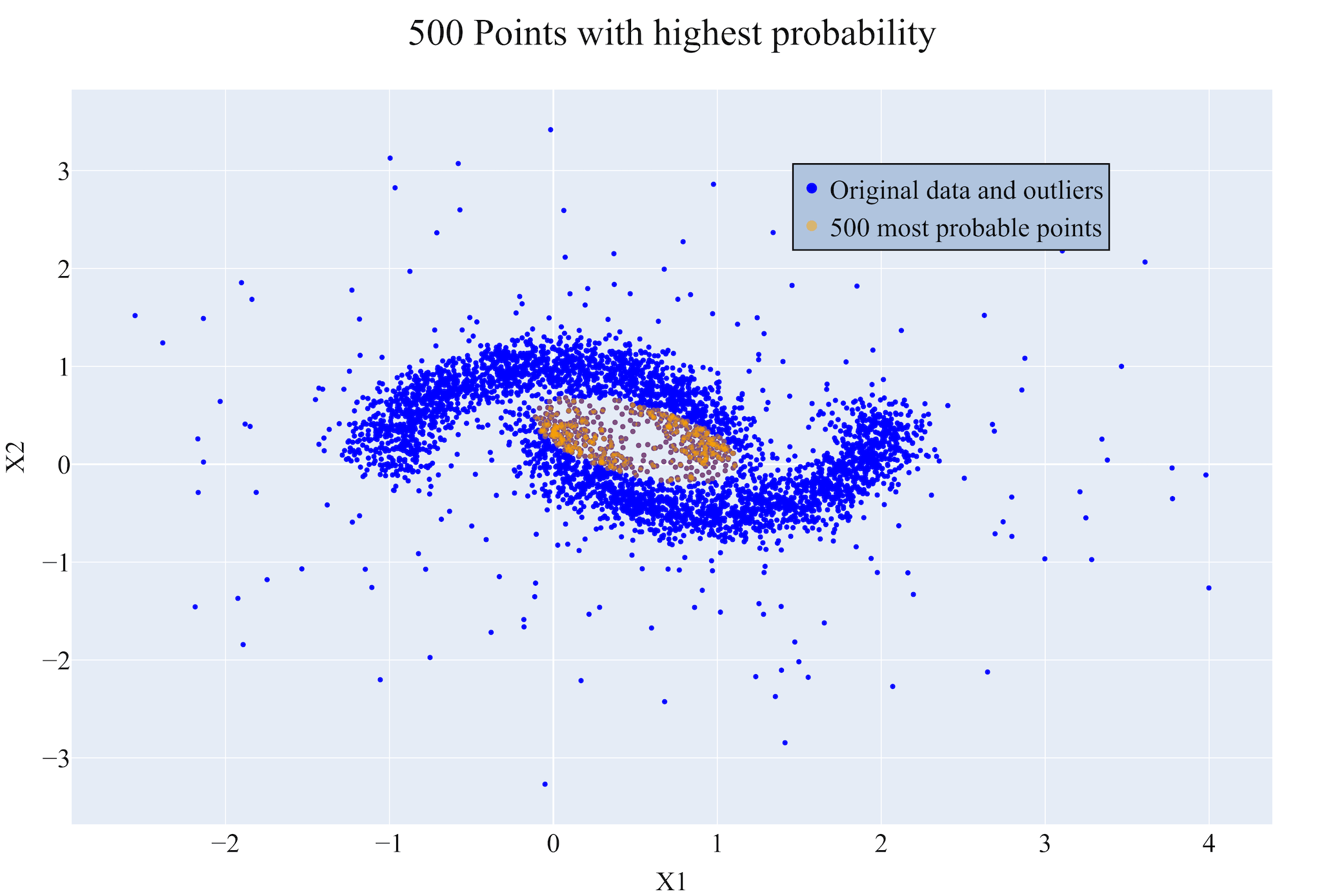}
\end{subfigure}
\begin{subfigure}[t]{0.49\linewidth}
    \includegraphics[width=1.0\linewidth]{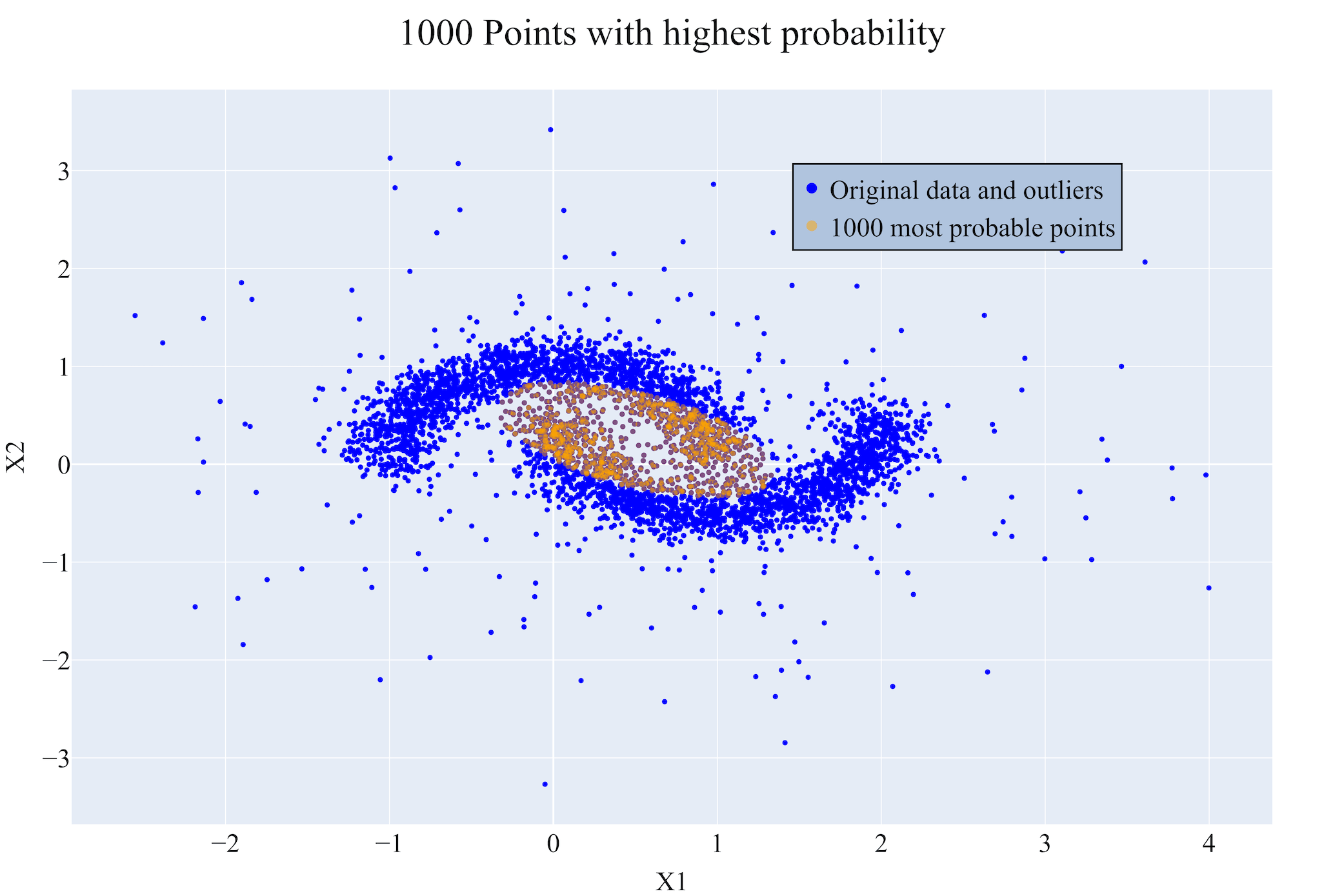}
\end{subfigure}
\begin{subfigure}[b]{0.49\linewidth}
    \includegraphics[width=1.0\linewidth]{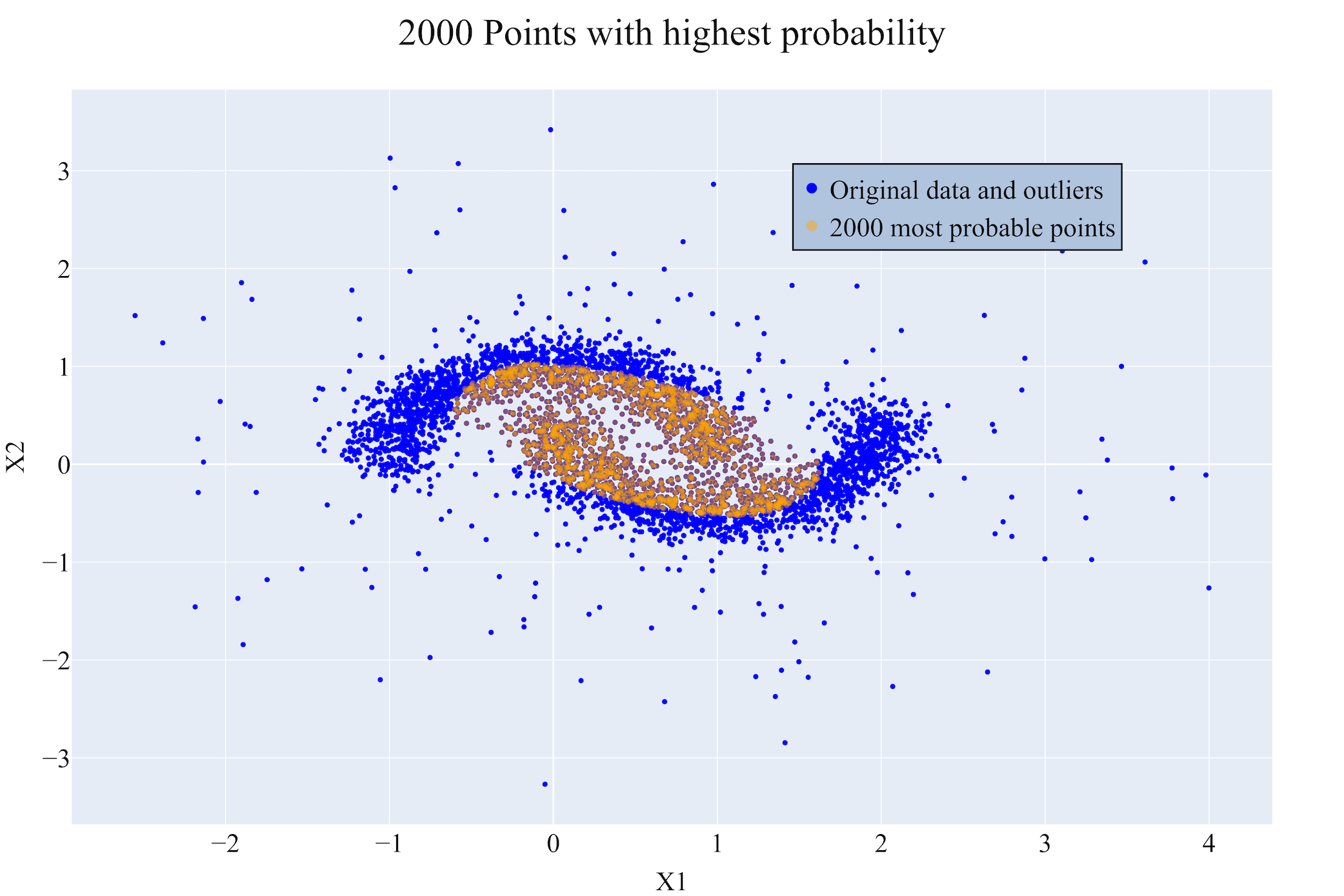}
\end{subfigure}
\begin{subfigure}[b]{0.49\linewidth}
    \includegraphics[width=1.0\linewidth]{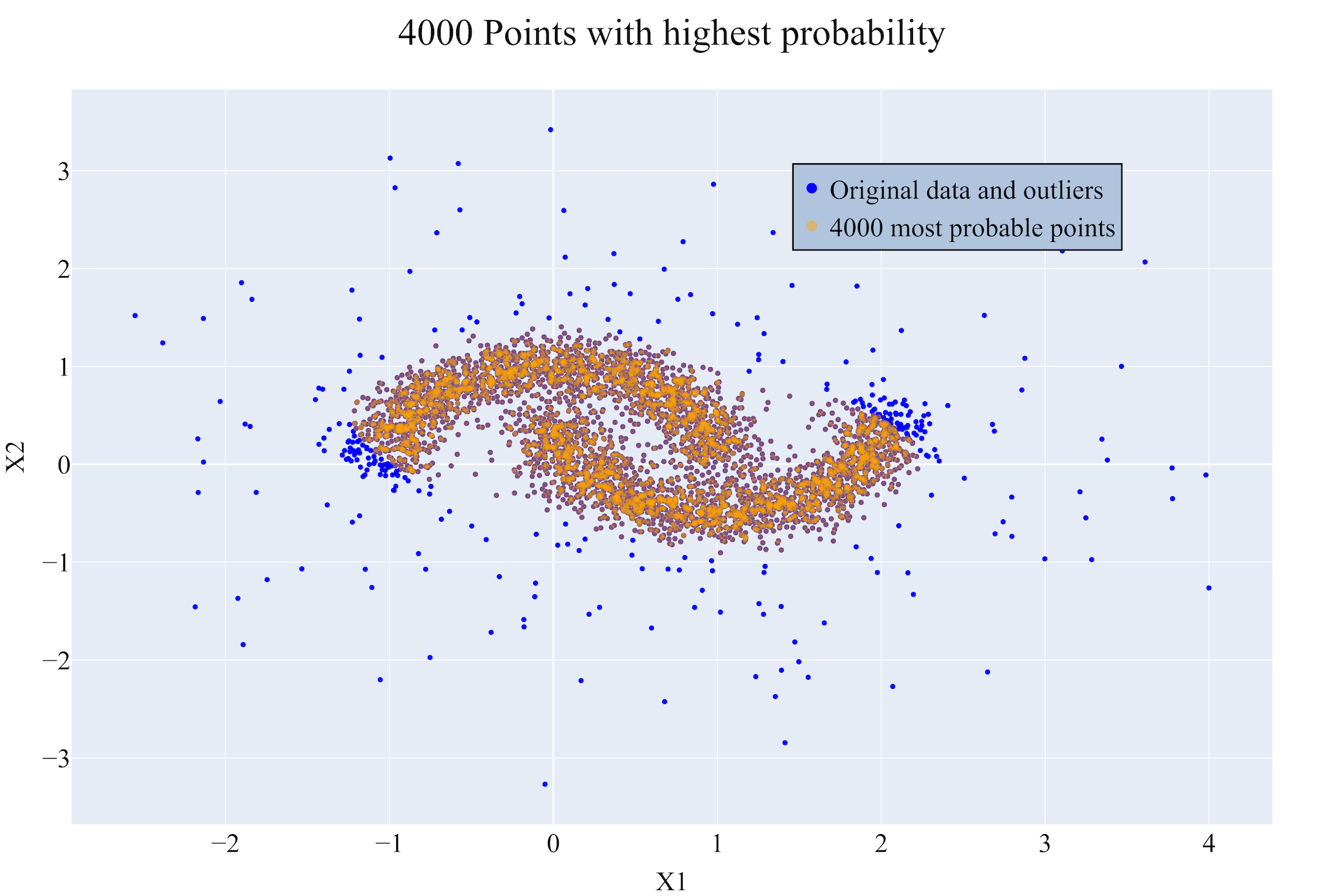}
\end{subfigure}
\caption{k points with highest probability under the normalizing flow, for  \textbf{k=500} (Top left), \textbf{k=2000} (Top right), \textbf{k=3000} (Bottom left) and \textbf{k=4000} (Bottom right). The procedure successfully removes the points farthest away from the real data.}
\label{fig:outliers_removal_ks}
\end{figure*}

To demonstrate our method, we design an experiment on the Moons data (see Appendix \ref{appendix:data}. Outliers are artificially generated as $\tilde{x} = x + \epsilon$, where $\epsilon \sim \mathcal{N}(0, \sigma)$ (see Figure \ref{fig:outliers_removal}). 4000 data points are originally in the dataset, and 200 outliers are added. One Normalizing Flow is then trained on the entire data. Due to its ability to model the data density and the geometry of the space it assigns high probability scores to points the closest to the data manifold and lower probabilities to most outliers that are ``far" from the data manifold.
The k points with highest probability are kept, or equivalently, the $card(X) - k$ are removed. We visualize the remaining data at the end of the procedure, for different values of $k$ in Figure \ref{fig:outliers_removal_ks}.

\section{Future Work}
\label{appendix:futurework}
In future work, we plan to address some the following themes:
    \begin{itemize}
     \item Searching for adequate basis functions presents both theoretical and practical challenges. As we aim to approximate a Gaussian Process, a better understanding of the need for wider or deeper networks is of interest.
     \item In addition, our approach is motivated by the need to obtain accurate uncertainty decomposition in real-world critical applications like medical diagnosis. Using our framework in such context would be an end goal.
     \item Ideally, a better theoretical understanding of the method and the phenomenon at stake would be performed, maybe in order to understand an implied relationship with Gaussian processes. We leave that aspect to a future work. 
 \end{itemize}

\end{document}